\documentclass{article}

\usepackage{microtype}
\usepackage{graphicx}
\usepackage{subfigure}
\usepackage{booktabs} 

\usepackage{amsmath,amsfonts,bm}

\def\eqref#1{equation~\ref{#1}}

\def\floor#1{\lfloor #1 \rfloor}
\def\1{\bm{1}}

\def\XS{\rmX_{\rmS}}

\def\rvf{{\mathbf{f}}}

\def\rvw{{\mathbf{w}}}
\def\rvx{{\mathbf{x}}}
\def\rvy{{\mathbf{y}}}

\def\rmA{{\mathbf{A}}}
\def\rmB{{\mathbf{B}}}

\def\rmH{{\mathbf{H}}}
\def\rmI{{\mathbf{I}}}

\def\rmK{{\mathbf{K}}}

\def\rmR{{\mathbf{R}}}
\def\rmS{{\mathbf{S}}}

\def\rmW{{\mathbf{W}}}
\def\rmX{{\mathbf{X}}}

\DeclareMathAlphabet{\mathsfit}{\encodingdefault}{\sfdefault}{m}{sl}
\SetMathAlphabet{\mathsfit}{bold}{\encodingdefault}{\sfdefault}{bx}{n}

\newcommand{\E}{\mathbb{E}}

\DeclareMathOperator*{\argmax}{arg\,max}
\DeclareMathOperator*{\argmin}{arg\,min}

\usepackage{hyperref}
\usepackage{url}
\usepackage{lipsum}
\usepackage{algorithm, algorithmic}
\usepackage{amsmath}
\usepackage{setspace}
\usepackage{bbm}
\usepackage{amsthm}
\usepackage{amssymb}
\usepackage{multirow}
\usepackage{array}
\usepackage{mathtools}
\usepackage{makecell}
\newtheorem{thm}{Theorem}
\newcommand{\beq}{ \begin{equation} }
\newcommand{\eeq}{ \end{equation} }

\usepackage[accepted]{icml2024}

\usepackage{amsmath}
\usepackage{amssymb}
\usepackage{mathtools}
\usepackage{amsthm}

\usepackage[capitalize,noabbrev]{cleveref}

\theoremstyle{plain}
\newtheorem{theorem}{Theorem}[section]

\newtheorem{lemma}[theorem]{Lemma}

\theoremstyle{definition}

\theoremstyle{remark}

\newcommand{\ours}{BWS }
\usepackage[textsize=tiny]{todonotes}

\icmltitlerunning{BWS: Best Window Selection Based on Sample Scores 
for Data Pruning across Broad Ranges}

\begin{document}

\twocolumn[
\icmltitle{BWS: Best Window Selection Based on Sample Scores \\
for Data Pruning across Broad Ranges}



\icmlsetsymbol{equal}{*}

\begin{icmlauthorlist}
\icmlauthor{Hoyong Choi}{equal,comp}
\icmlauthor{Nohyun Ki}{equal,sch}
\icmlauthor{Hye Won Chung}{sch}
\end{icmlauthorlist}

\icmlaffiliation{comp}{Samsung Research, Seoul, South Korea}
\icmlaffiliation{sch}{School of Electronic Engineering, Korea Advanced Institute of Science and Technology (KAIST), Daejeon, South Korea}


\icmlcorrespondingauthor{Hye Won Chung}{hwchung@kaist.ac.kr}

\icmlkeywords{Machine Learning, ICML}

\vskip 0.3in
]



\printAffiliationsAndNotice{\icmlEqualContribution} 

\begin{abstract}

Data subset selection aims to find a smaller yet informative subset of a large dataset that can approximate the full-dataset training, addressing challenges associated with training neural networks on large-scale datasets. However, existing methods tend to specialize in either high or low selection ratio regimes, lacking a universal approach that consistently achieves competitive performance across a broad range of selection ratios. We introduce a universal and efficient data subset selection method, Best Window Selection (BWS), by proposing a method to choose the best window subset from samples ordered based on their difficulty scores. This approach offers flexibility by allowing the choice of window intervals that span from easy to difficult samples. Furthermore, we provide an efficient mechanism for selecting the best window subset by evaluating its quality using kernel ridge regression. Our experimental results demonstrate the superior performance of BWS compared to other baselines across a broad range of selection ratios over datasets, including CIFAR-10/100 and ImageNet, and the scenarios involving training from random initialization or fine-tuning of pre-trained models.

\end{abstract}

\section{Introduction}
\begin{figure*}[ht]
\centering
{\includegraphics[width=0.99\linewidth]{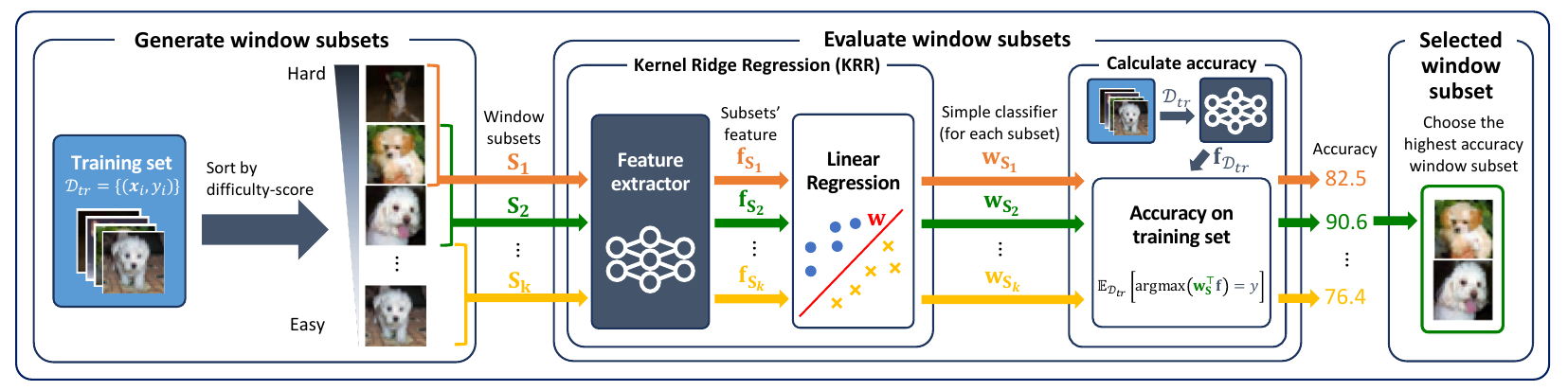}}
    \caption{
    Overview of the proposed method, Best Window Selection (BWS). BWS is composed of two parts, 1) generating window subsets and 2) evaluating window subsets. We first sort samples by a difficulty score (e.g., Forgetting \citep{forgetting}) and generate window subsets of a fixed size while varying their starting points. We then evaluate the window subsets, by solving kernel ridge regression on the input features of each window subset and obtaining simple (linear) classifiers associated with each window subset. Finally, we evaluate the performance of these classifiers on the full training dataset to identify the best window subset achieving the highest accuracy. 
    }\label{fig:main_figure}
\end{figure*}
In many machine learning tasks, the effectiveness of deep neural networks often relies on large-scale datasets that include a vast number of samples, enabling them to achieve state-of-the-art performances. However, working with such large datasets presents several challenges, including the high computational costs, storage requirements, and potential concerns related to privacy \citep{green,energy}. 
Data subset selection emerges as a promising approach to address these issues. This involves the careful selection of a smaller, yet highly informative, subset from the original large dataset. The goal is to find a subset with a specified selection ratio that approximates the performance of the entire dataset or incurs minimal performance loss.

Data subset selection has two primary approaches: score-based selection and optimization-based selection. 
Score-based selection involves defining a specific score to measure each sample's influence \citep{infl_func}, difficulty \citep{forgetting,EL2N}, or consistency \citep{c-score} in training neural networks.
The primary goal is to identify the most valuable or influential samples within the dataset while pruning the remaining samples that have minimal impact on the model's generalization ability. 
On the other hand, optimization-based selection approaches find the optimal subset of a fixed size that can best approximate  the full dataset training in terms of loss gradient or curvature by solving the associated optimization problem \citep{CRAIG,Adacore,LCMat, CREST}.  The original optimization, which is NP-hard, is commonly approximated by submodular functions and a greedy algorithm is adopted to sequentially select the samples up to the size limit of the subset.

While the prior approaches successfully reduce dataset size in specific scenarios, there is not a single selection method that universally outperforms other baselines across broad selection ratios. To illustrate this, we conduct a benchmark comparison between two methods: Forgetting score \citep{forgetting} representing the score-based selection approach, and LCMat \citep{LCMat} representing the optimization-based selection approach. We evaluate the test accuracy of models trained with different subset sizes of datasets, including CIFAR-10/100 \citep{CIFAR} and ImageNet \citep{ImageNet}, ranging from 1\% to 90\%, as selected by these two methods (Table \ref{tab:build_up}). 
Score-based methods, which prioritize samples of high influence or difficulty, tend to initially select rare yet influential samples while excluding typical or easy samples. These methods demonstrate competitive performance, nearly matching the full-dataset training, when the selection ratio is sufficiently high (e.g., over 40\% for CIFAR-10). However, they suffer significant performance degradation as the selection ratio decreases.
In contrast, optimization-based methods tend to select representative samples that best approximate the full dataset training. Consequently, they achieve competitive performance even with very low selection ratios. However, their performance gains are limited as the selection ratio increases due to lack of diversity in sample selection. These findings show the variability in the criteria for an effective data subset, depending on the selection ratio, and highlight that previous methods may not be general enough to cover the entire spectrum of selection ratios.

Our key contribution is the development of a universal and efficient data selection method capable of maintaining competitive performance across a wide range of selection ratios. We introduce the Best Window Selection (BWS) method, illustrated in Fig. \ref{fig:main_figure}. The key idea involves ordering samples based on their difficulty-based sample scores and
offering flexibility in choosing a `window subset' from the ordered samples.
Here, the window subset is defined as a subset consisting of samples with a contiguous ranking of difficulty. By allowing the starting point (the ranking of the hardest data in the subset) of each window subset to vary, we enable the selection of easy, moderate, or hard data subsets. 
We first demonstrate the existence of the best window that achieves the highest test accuracy for each subset size, and reveal that the optimal starting point for the best window varies depending on both the subset size and dataset.
We then present a computationally-efficient method for selecting the best window subset without the need to evaluate models trained with each subset. 
We achieve this by solving a kernel ridge regression problem using samples from each window, evaluating the corresponding solution's performance on the full training dataset, and selecting the best performing window subset.

We evaluate our selection method, BWS on CIFAR-10/100 and ImageNet, demonstrating that BWS consistently outperforms other baselines, including both score-based and optimization-based approaches, across a wide range of selection ratios ranging from 1\% to 90\%. 
For CIFAR-10, BWS achieves a 15-30\% improvement in test accuracy compared to Forgetting  \citep{forgetting} in the low selection ratios of 1-10\%. It also demonstrates competitive performance in the high selection ratio regime, reaching up to 93\% test accuracy with only a 40\% data subset. BWS also consistently outperforms optimization-based techniques such as LCMat \citep{LCMat} and AdaCore \cite{Adacore},
despite requiring significantly lower computational costs. 
Furthermore, we empirically verify that BWS is effective across different model architectures, including pre-trained ViT \citep{ViT}. Another significant advantage of our method is its resilience to label noise, enhancing its robustness in sample selection.
Our code is publicly available at \href{https://github.com/NohyunKi/BWS}{https://github.com/NohyunKi/BWS}.

\section{Related Works}

\paragraph{Score-based selection}
Some initial works in score-based selection use validation/test sets to quantify the effect of each training instance. Data Shapley \citep{shapley, D-shapley, beta-shapley} evaluates the value of each instance by measuring the average change in validation accuracy when that instance is excluded from the dataset. Influence Function \citep{infl_func, TracIn} approximates how a model's prediction changes as individual training examples are visited. 
In the absence of a validation set, score-based selection quantifies the learning difficulty or consistency of samples during neural network training. Forgetting \citep{forgetting} and EL2N \citep{EL2N} introduce a difficulty score to measure a data point's learning  difficulty. Memorization \citep{memorization} and C-score \citep{c-score} aim to predict the accuracy on a sample when the full dataset is utilized, except for that sample. CG-score \citep{CG-score} evaluates data instances without model training by calculating the analytical gap in generalization errors when an instance is held out. These score-based methods prioritize difficult or influential samples for data subset selection. While they effectively select a subset approximating the full-dataset performance, their performance degrades significantly as the selection ratio decreases, as achieving high performance solely with  difficult samples becomes challenging.

\paragraph{Optimization-based selection}
Optimization-based selection involves formulating an optimization problem to select a coreset of a given size that can effectively approximate the diverse characteristics of the full dataset. These methods include coreset selection to approximate the training distribution by herding \citep{herding} or k-center algorithms \citep{k-center}.
Recent approaches have sought subsets of samples approximating loss gradients or curvature by CRAIG \citep{CRAIG}, CREST \citep{CREST}, and AdaCore \citep{Adacore}. While these methods have proven effective, they are computationally demanding and necessitate full-dataset sampling at each epoch.
LCMat \citep{LCMat} addresses this computational challenge by aligning both gradients and Hessians without requiring periodic full-dataset sampling. However,  these methods often struggle to choose diverse samples, and their performance does not match that of score-based approaches, in the intermediate to high selection ratio regimes.

In contrast to these approaches, we develop a universal selection method capable of consistently identifying a high-performance subset across a wide range of selection ratios. While recent methods like Moderate-DS \citep{mds} and CCS \citep{CCS} have also aimed for universality across various selection ratios, our method outperforms these approaches, over a broad range of selection ratios, as demonstrated in Section \ref{sec:experiment}.
Moderate-DS selects samples closest to the median of the features of each class, while CCS prunes a $\beta$\% of hard examples, with $\beta$ being a hyperparameter, and then selects samples with a uniform difficulty score distribution. Importantly, our method does not require hyperparameter tuning, such as setting $\beta$ in CCS, since we assess the quality of window subsets and efficiently find the best one using kernel ridge regression.

\section{Motivation}
\subsection{No single method prevails over the entire range}\label{sec:failure}

We conduct an evaluation of existing data selection methods across a wide range of selection ratios. Specifically, we benchmark two representative methods: Forgetting score \citep{forgetting}, representing difficulty score-based selection, and LCMat \citep{LCMat}, representing optimization-based selection. We assess the test accuracy of models trained on subsets of CIFAR-10/100 and ImageNet, with selection ratios ranging from 1\% to 90\%, as summarized in  Table \ref{tab:build_up}. For the Forgetting score approach, we sort the samples in descending order based on their scores, defined as the number of times during training the decision of that sample switches from a correct one to incorrect one,  and select the top-ranking (most difficult) samples. In contrast, for LCMat, we employ an optimization to identify a subset that best approximates the loss curvature of the full dataset. We employ ResNet18 \citep{resnet} for CIFAR-10 and ResNet50 for CIFAR-100 and ImageNet.

We can observe that the most effective strategy varies depending on the selection ratios, and there is no single method that consistently outperforms others across the entire range of selection ratios. Specifically, for CIFAR-10 with low subset ratios (1-30\%), the optimization-based selection (LCMat) performs better than the difficulty score-based selection (Forgetting). In this regime, the `Forgetting' even underperforms random selection. However, as the subset ratio increases beyond 40\%, the `Forgetting' outperforms both the LCMat and random selection. Similar trends are observed for CIFAR-100 and ImageNet. Interestingly, for CIFAR-100, there is an intermediate regime where neither the `Forgetting' nor LCMat outperform random sampling.

These findings emphasize that the desired properties of data subsets change depending on the selection ratios. In cases of low selection ratios (sample-deficient regime), it is more beneficial to identify a representative subset that closely resembles the full dataset in terms of average loss gradients or curvature during training. However, as the selection ratio increases (sample-sufficient regime), preserving the high-scoring, rare or difficult-to-learn samples becomes more critical, as these samples are known to enhance the generalization capability of neural networks and cannot be fully captured by a representative subset that reflects only the average behavior of the full dataset.

\begin{table}[t]
\vspace{-0.7em}
\caption{Test accuracy across various selection ratios for the CIFAR-10/100 and ImageNet datasets, with subsets selected using random sampling, Forgetting score \citep{forgetting}, and LCMat \citep{LCMat}. The best performance among the three is highlighted in \textbf{bold}.}
\label{tab:build_up}
\begin{center}
\begin{small}
\resizebox{0.99\linewidth}{!}{
\begin{tabular}{cc|ccccccccc|c}
\toprule
\multicolumn{2}{c|}{Selection ratio} & 1\% & 5\% & 10\% & 20\% & 30\% & 40\% & 50\% & 75\% & 90\% & 100\% \\
\midrule
& Random & 39.10 & \textbf{67.14} & \textbf{78.43} & 86.87 & 89.91 & 91.66 & 92.83 & 94.40 & 95.08 & \\
CIFAR-10 & Forgetting & 30.08 & 42.39 & 54.31 & 79.19 & 89.13 & \textbf{93.41} & \textbf{94.49} & \textbf{95.31} & \textbf{95.14} & 95.40 \\
& LCMat & \textbf{41.53} & 66.86 & 77.48 & \textbf{87.34} & \textbf{90.72} & 92.45 & 93.38 & 94.90 & 95.19 & \\
\midrule
& Random & 5.89 & 23.76 & 42.03 & 55.03 & \textbf{65.98} & \textbf{69.23} & \textbf{73.84} & 76.53 & 78.29 & \\
CIFAR-100 & Forgetting & 7.01 & 20.69 & 34.22 & 50.95 & 61.54 & 68.92 & 72.65 & \textbf{78.55} & \textbf{79.69} & 78.81 \\
&  LCMat & \textbf{8.43} & \textbf{28.51} & \textbf{42.81} & \textbf{55.77} & 64.39 & 67.22 & 73.11 & 77.51 & 78.47  & \\
\midrule
& Random & \textbf{6.14} & \textbf{33.17} & 45.87 & 59.19 & 65.94 & 68.23 & 70.14 & 73.74 & 74.83 & \\
ImageNet & Forgetting & 4.78 & 28.18 & 45.84 & \textbf{60.75} & \textbf{67.48} & \textbf{70.26} & \textbf{72.73} & \textbf{74.63} & \textbf{75.53} & 75.85 \\
&  LCMat & 6.01 & 32.26 & \textbf{46.08} & 59.02 & 65.28 & 68.50 & 70.30 & 74.13 & 74.81 & \\
\bottomrule
\end{tabular}}
\end{small}
\end{center}
\end{table}

\subsection{Theoretical analysis}\label{sec:theory}
To validate this experimental finding, we provide a theoretical analysis of optimal subset selection, revealing similar change of trends in the desirable subsets depending on the selection ratios. 
We consider a binary classification problem by solving a linear regression problem, as detailed below:
Data samples $\rvx_1, \rvx_2, \dots \rvx_n\in \mathbb{R}^d$ are generated from a multivariate normal distribution, 
$\mathcal{D}=\frac{1}{\sqrt{d}} \mathcal{N}(0,\rmI_d)$. 
The label $y_i$ of sample $\rvx_i$ is determined by the sign of its first element. Specifically, if $(\rvx_i)_1>0$ then $y_i=1$; and if $(\rvx_i)_1<0$, then $y_i=-1$. We define the score of each sample as $1/|(\rvx_i)_1|$. Samples closer to the decision boundary $(\rvx)_1=0$ have higher scores, while those farther from the boundary have lower scores. We select a label-balanced subset of size $m$, denoted by $(\rmX_{\rmS}, \rvy_\rmS)\in\mathbb{R}^{d\times m}\times \{-1,1\}^m$, and use it to solve a linear regression problem to find $\rvw_{\rmS} = \argmin_{\rvw\in \mathbb{R}^d} \|\rvy_{\rmS}-\rmX_{\rmS}^\top\rvw\|_2^2$. 
For a new sample $\rvx'$, our decision will be $+1$ if $\rvw_{\rmS}^\top \rvx'>0$ and $-1$ otherwise. Thus, we consider $\rvw_{\rmS}$ to be a better solution when the value of its first element, $(\rvw_{\rmS})_1$, is larger.
For the above setup, we analyze the solution $\rvw_{\rmS}$ depending on the subset size $|\rmS|$. 

A similar problem setup was analyzed in \citep{beating}, demonstrating that the optimal selection strategy varies depending on the subset ratio. Specifically,  \citet{beating} considers a max margin classifier trained on a data subset selected by the teacher-perceptron model, providing a comprehensive set of equations enabling numerical computation of the generalization error for various subset data distributions. In contrast, our contribution lies in providing a closed-form solution for the optimal linear classifier, as summarized in the theorem below. This theorem shows the transition of the optimal sample selection strategy between sample-deficient and sample-sufficient regimes.

\begin{thm}[Informal]\label{thm:opt_reg}
If the subset size is as small as $|\rmS|=m\ll  \sqrt{d/\ln d}$, then the first coordinate of $\rvw_{\rmS}$ is approximated as $(\rvw_{\rmS})_1 \approx \sum_{i=1}^m|(\rvx_i)_1|$. On the other hand, if $|\rmS|=m\gg d^2\ln{d}$, it can be approximated as $(\rvw_{\rmS})_1 \approx (\sum_{i=1}^m|(\rvx_i)_1|)/(\sum_{i=1}^m|(\rvx_i)_1|^2)$. 
\end{thm}
A more formal statement and the proof of Thm. \ref{thm:opt_reg} is available in Appendix \ref{sec:app:proof}.
From Thm.\ref{thm:opt_reg}, it is evident that the characteristics of the desirable data subset $\rmX_{\rmS}$ vary depending on the subset size regime. In the sample-deficient regime $(m \ll \sqrt{d/\ln d})$, it is more advantageous to include samples that are farther from the decision boundary (easy samples) in $\rmX_{\rmS}$ to train a better classifier, resulting in a higher value of  $(\rvw_{\rmS})_1$. Conversely, in the sample-sufficient regime $(m \gg d^2\ln d)$, it is more beneficial to include samples closer to the decision boundary (difficult samples) to increase  $(\rvw_{\rmS})_1$. We conjecture that the relatively wide gap between two distinct regimes ( $[\sqrt{d/\ln{d}},d^2\ln{d}]$) may be attributed to the loose analysis. We anticipate that a more precise boundary will occur at $m=\Theta(d)$, where $m\ll d$ ($m\gg d$) corresponds to the sample-deficient (sufficient) regime. 
We provide empirical results that support this theoretical analysis and our conjecture in Appendix \ref{sec:toy}.

Having identified the distinct properties of desirable data subsets depending on the subset size, the remaining question is how to design a universal data selection method capable of performing well across a wide range of selection ratios.

\section{Window Subset: Flexible Subset Selection}
\subsection{Desirable difficult level for data subsets in moderate selection ratios? Hard, but not the hardest}\label{sec:train_split}

The underlying rationale for difficulty score-based selection methods like Forgetting \citep{forgetting} and EL2N \citep{EL2N} is that training models on a subset consisting of challenging data will enable the models to learn (or memorize) the atypical features of hard samples, while still retaining the capacity to learn typical  features of easier samples. However, as shown in our empirical findings in Sec. \ref{sec:failure} and further supported by our theoretical analysis in Sec. \ref{sec:theory}, this assumption may not hold when the subset ratio is extremely small. This leads us to our next question: Is it still feasible for models trained on \textit{hard} instances to effectively learn \textit{easier} instances, without having been exposed to these samples during training, at \textit{moderate} selection ratios?

\begin{figure}[tb!]
\begin{center}
\centerline{\includegraphics[width=0.85\linewidth]{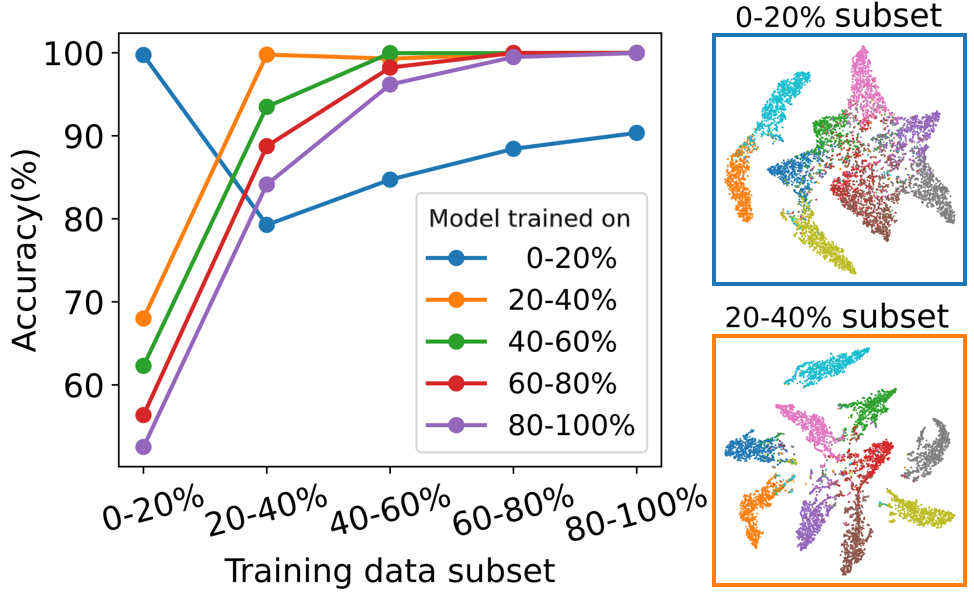}}
\caption{
Results on ``training set split" experiment on CIFAR-10 dataset, when five different models are trained by five different data subsets, divided by their difficulty rankings, $[0,20]\%$ (hardest) to $[80,100]\%$ (easiest). Model accuracies ($y$-axis) are evaluated on all five subsets ($x$-axis) separately. 
Right figures visualize the t-SNE of test samples' features extracted from models trained by the hardest $[0,20]\%$ subset (top) and the $[20,40]\%$ subset (bottom).
}
\label{fig:train_split}
\end{center}
\end{figure}

To investigate this, we design a ``training set split" experiment on CIFAR-10 dataset. We divide the training dataset into five subsets and observe the impact of training on each subset on the accuracy across the other subsets. In detail, we sort the CIFAR-10 training instances by forgetting score \cite{forgetting} and divide them into five subsets based on consecutive ranking intervals: the hardest 20\% (rankings within $[0,20]\%$), $[20,40]\%$, and so on, up to the easiest 20\% ($[80,100]\%$). We train five different ResNet18 models, each on one of these subsets, and then evaluate their classification accuracies on all five subsets separately.

The results, presented in Fig. \ref{fig:train_split}, reveal that models trained on harder data subsets generally perform better across all subsets, with the exception of the model trained solely on the hardest 20\%. For instance, the model trained on the $[20,40]\%$-ranked subset effectively classifies instances not only within its training range but also those in the easier $[40,100]\%$ range. 
This suggests that training with harder instances helps the model learn both the unique features of these challenging instances and the common, representative features of the entire dataset. This finding supports the rationale behind existing score-based selection methods, which prioritize selecting challenging data for training.

Yet, this pattern does not hold for the model trained exclusively on the \textit{hardest} 20\% subset. This model exhibits a significant drop in accuracy across all the easier subsets, except for the hardest subset it was trained on. This indicates that a model trained with only the most challenging instances lacks generalizability to easier samples. 

We support this claim by analyzing the feature spaces of models trained with the hardest $[0,20]\%$ subset and the subsequent $[20,40]\%$ subset. Our focus is on demonstrating that the model trained with the hardest 20\% subset struggles to effectively create a feature space for classification. We extract features of CIFAR-10 test samples from each model and visualize their t-SNE \citep{TSNE} in Fig. \ref{fig:train_split} (right). The figure reveals that the feature space generated by the model trained on the hardest subset does not efficiently cluster test samples by class. We further quantify this using the \textit{neural collapse} property \citep{neuralcollapse}, which compares within-class feature variability to inter-class feature variability. Let $f_{k,i}$ be the feature of the $i$-th data in the $k$-th class, $\mu_k=\frac{1}{n}\sum_{i=1}^nf_{k,i}$ be the feature mean of class $k$, and $\mu_G=\frac{1}{K}\sum_{k=1}^K\mu_k$ be the global mean feature. The within-class covariance $\Sigma_w$ is defined by $\frac{1}{Kn}\sum_{k=1}^K\sum_{i=1}^n (f_{k,i}-\mu_k)(f_{k,i}-\mu_k)^\top$, and the inter-class covariance $\Sigma_B$ by $\frac{1}{K}\sum_{k=1}^K(\mu_k-\mu_G)(\mu_k-\mu_G)^\top$. The trace, $\text{tr}(\Sigma_W\Sigma_B^\dagger)$, then measures the clusterability of features with respect to their classes, with a lower value indicating better clustering.
The $\text{tr}(\Sigma_W\Sigma_B^\dagger)$ values for models trained on $[0, 20]\%$ (the hardest 20\%), $[20, 40]\%$, and so on, up to $[80, 100]\%$, and the full dataset, are 9.33, 1.68, 1.99, 2.60, 3.35, and 1.04, respectively. Notably, there is a significant increase in $\text{tr}(\Sigma_W\Sigma_B^\dagger)$ for the hardest subset, suggesting poor feature learning for classification. 

In summary, training with harder data generally benefits learning both representative and atypical features, aiding in better model generalization. However, when the subset ratio is moderate and the subset consists of the hardest samples, the model may suffer significant performance drop and fail to establish an effective feature learning for classification.

\subsection{The existence of a high-performing window subset}\label{sec:sliding_window}
\begin{figure}[t]
\centering
    \subfigure[CIFAR-10 \label{fig:window_CIFAR10}]{\includegraphics[width=0.48\linewidth]{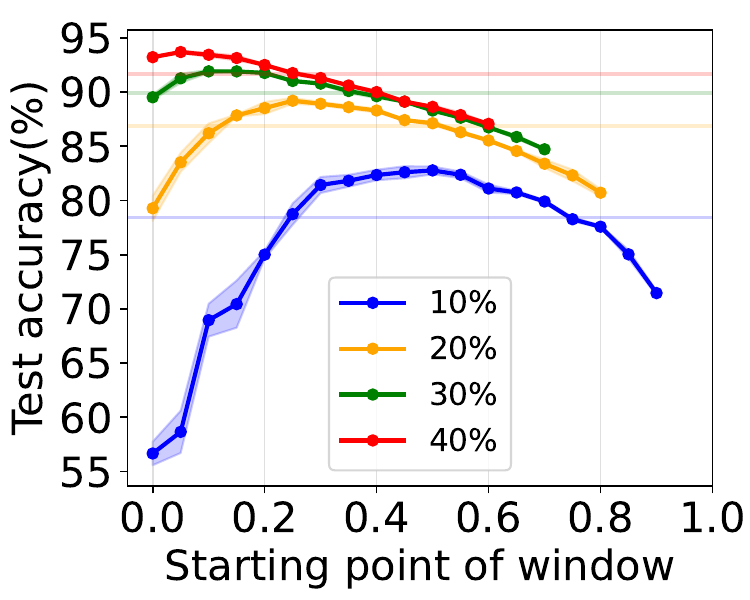}}
    \subfigure[CIFAR-100 \label{fig:window_CIFAR100}]{\includegraphics[width=0.48\linewidth]{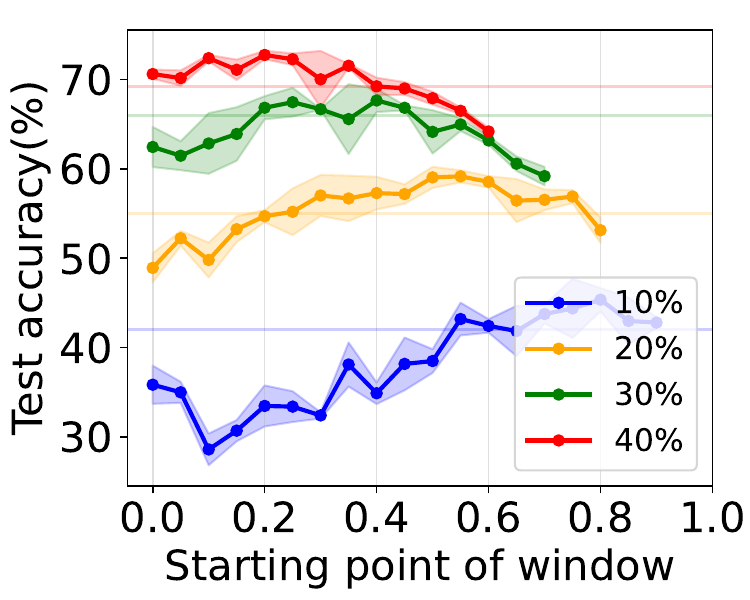}}
    \caption{
    Sliding window experiments to measure the test accuracy of the models trained by window subsets while changing the starting point of the windows in CIFAR-10 (left) and  CIFAR-100 (right) dataset. 
    Samples are sorted in descending order by their difficulty scores. The horizontal lines are results from random selection. For each subset ratio, there exists the best window, and its starting point shifts toward left as the subset ratio increases.
    Results for ImageNet dataset is also reported in Appendix 
    \ref{sec:sliding_window_app}
    }
    \label{fig:window_exp}
\end{figure}

Section \ref{sec:train_split} implies that for each subset ratio, there is a proper difficulty level of the subset for better model generalization. Expecting that a subset composed of samples of proper difficult level will perform well, we consider the window selection method, similar to \citep{selmatch}, that selects a window subset from samples ordered by their difficulty scores. 
In detail, we sort the samples in descending order based on their difficulty scores and select a starting point, such as $s\%$ for a given window size of $w\%$, to choose continuous intervals of samples within $[s, s+w]\%$.
This approach has two merits: 1) flexibility and 2) computational-efficiency.
The flexibility in choosing the starting point $s$\% of the window allows us to opt for easy, moderate, or hard data subsets depending on the choice of the starting point. 
The search space of window selection method is confined to the number of possible starting points for the windows, making the window selection method computationally much more efficient compared to a general subset selection where the search space scales as $\genfrac(){0pt}{1}{n}{m} \approx \exp(cn)$ for some constant $c>0$ when the subset size $m$ is a constant fraction of $n$.

We first explore the performance of the window selection approach while varying the starting point and illustrate the existence of the best window subset. We sort the samples from CIFAR-10/100 in descending order based on their Forgetting scores \citep{forgetting}, and select windows of different sizes, ranging from $10\%$ to $40\%$, by adjusting the starting point from $0$ to $(100-w)\%$ with a step size of $5\%$. 
We then train ResNet18 for CIFAR-10 and ResNet50 for CIFAR-100 using the windows subsets and plot the resulting test accuracies in Fig.~\ref{fig:window_exp}.

We can observe that, for each subset ratio, there exists an optimal starting point, and this optimal point shifts towards lower values (indicating more difficult samples) as the window subset size increases. Specifically, for CIFAR-10, the optimal window subset of size $10\%$ falls within the interval $[50,60]\%$, while for a window size of $40\%$, it falls within $[5,45]\%$. Similar trends are observed for CIFAR-100, albeit with distinct optimal starting points depending on the dataset. For CIFAR-100, with a window size of $10\%$, the best window subset comprises samples from $[80,90]\%$, primarily consisting of easy samples. It is important to note that the $10\%$ subset for CIFAR-100 includes only 50 samples per class, whereas for CIFAR-10, it includes 500 samples per class. Consequently, the optimal $10\%$ window for CIFAR-100 ($[80,90]\%$) tends to include more easy and representative samples capable of capturing the representative features of each class. 

The observation that the optimal starting point of the window subset varies based on both the subset size and the dataset introduces a new challenge in window selection: How can we efficiently identify the best window subset without having to evaluate models trained on each subset? We address this crucial question by introducing a proxy task to estimate the quality of window subsets.

\section{Best Window Selection (BWS)}\label{sec:methodology}

\begin{algorithm}[t]
\small{
\setstretch{1.1}
\caption{BWS: Best Window Selection Method}\label{alg1}
    \textbf{Input} Dataset $\{(\rvx_i, y_i)\}_{i=1}^n$ sorted by difficulty scores from the hardest to easiest, subset size $m$, and step size $t$.
\begin{algorithmic}
\STATE Train a feature extractor $f(\cdot)$ by $m$  randomly chosen samples from the dataset.
\STATE Extract the features of the samples by using $f(\cdot)$ and denote them by $\rvf_i=[f(\rvx_i),1]$.
\FOR{$k\in\{0, t, 2t, 3t \dots, \floor{(n-m)/t}t\}$}
    \STATE Define a window subset ${\rmS} = \{(\rvf_i,y_i)\}_{i=k}^{k+m-1}$.
    \FOR{$c \in\{ 1, 2, \dots C\}$} 
        \STATE For the samples in $\rmS$ with label $c$, set the label equal to 1. For others, set the label to 0.
        \STATE Solve the linear regression problem Eq.\ref{eq:regression} with the window subset ${\rmS}$. Let $\rvw_{\rmS}(c)$ be the solution.
    \ENDFOR
    \STATE Obtain $\rvw_{\rmS} \in \mathbb{R}^{(d+1)\times C}$ by $\rvw_{\rmS}:=[\rvw_{\rmS}(1),\dots \rvw_{\rmS}(C)]$. 
    \STATE Calculate the accuracy of $\rvw_{\rmS}$ by  \\
    $\frac{1}{n}\sum_{i=1}^n \mathbbm{1} (\argmax_c(\rvw_{\rmS}^\top\rvf_i)_c=y_i)$.
\ENDFOR
\end{algorithmic}
\textbf{Output} Window subset ${\rmS}$ for which the accuracy of $\rvw_{\rmS}$ is maximized.
}
\end{algorithm}

\begin{table}[t]
\caption{
Comparison between the window subsets chosen by the sliding window experiment in Fig. \ref{fig:window_exp} (left) and BWS (using the KRR as a proxy task) (right) in terms of their starting points and test accuracies. The chosen windows align well between the two. 
}
\label{tab:window_result}
  \centering
  \resizebox{\linewidth}{!}{  
\begin{tabular}{c|c|c|c|c}
\toprule
\multirow{2}{*}{Ratio} & \multicolumn{2}{c|}{Sliding window experiment} & \multicolumn{2}{c}{BWS} \\
\cmidrule{2-5}
& Starting point & Test accuracy & Starting point & Test accuracy \\
\midrule
10\% & 50\% & 82.67 & 55\% & 82.29 \\
\midrule
20\% & 25\% & 89.06 & 30\% & 88.74 \\
\midrule
30\% & 15\% & 91.80 & 15\% & 91.80 \\
\midrule
40\% & 5\% & 93.59 & 5\% & 93.59 \\
\bottomrule
\end{tabular}}
\end{table} 

Our goal is to develop a computationally-efficient method capable of assessing and identifying the best window subset without requiring the training of a model on every potential subset. To achieve this goal, we propose to solve a kernel ridge regression (KRR) problem by using each window subset and evaluate the performance of the corresponding solution on the full training datasets. 
Using KRR for a proxy task is motivated by the observation that the kernel regression with the model-related kernels can provide a good approximation to the original model  \citep{neal2012bayesian,lee2018deep, jacot2018neural, Equivalence_NN_KRR}, while providing computational efficiency compared to training the actual models. 
We provide further justifications of this proxy task in Appx. \ref{justification_krr}.
Alg. \ref{alg1} outlines the main steps.

Let $\rvf_i:=[f(\rvx_i),1]\in \mathbb{R}^{d+1}$ be the feature vector of $\rvx_i$ obtained by a feature extractor $f(\cdot)$. The details of the feature extractor is available in the end of this section. 
For each window subset $\rmS=\{(\rvf_i, y_i)\}_{i=1}^m$ composed of $m$ samples, define $\rmX_\rmS:=[\rvf_1,\dots, \rvf_m]$  and $\rvy_\rmS:=[y_1,\dots, y_m]$. Then, we denote the problem of kernel ridge regression, and the corresponding solution, using the subset $\rmS$ by
\begin{align}
    \rvw_{\rmS} &:= \argmin_{\rvw} \|\rvy_{\rmS}-\rmX_{\rmS}^\top\rvw\|_2^2 + \lambda\|\rvw\|_2^2,\label{eq:regression}\\
    \rvw_{\rmS} &= (\lambda\rmI_{d+1}+\rmX_{\rmS}{\rmX_{\rmS}}^\top)^{-1}\rmX_{\rmS}\rvy_{\rmS} \nonumber\\
    &= \rmX_{\rmS}(\lambda\rmI_{m}+\rmX_{\rmS}^\top\rmX_{\rmS})^{-1}\rvy_{\rmS}\label{eq:solution}.    
\end{align}
We set $\lambda=1$ to prevent singularity in matrix inversion. The matrix inversion in Eq. \ref{eq:solution} can be performed efficiently in a lower dimension between $d+1$ and $m$.

\begin{figure*}[!tb]
\centering
    {\includegraphics[width=0.95\linewidth]{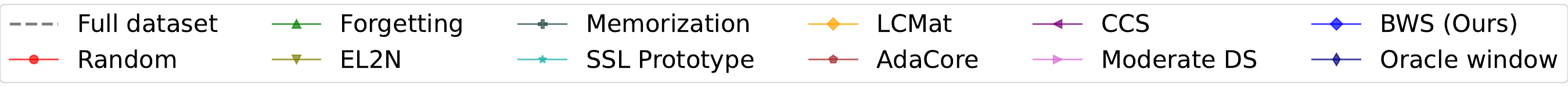}} \\
    \vspace{-0.6em}
    \subfigure[CIFAR-10 \label{fig:pruning_CIFAR10}]{\includegraphics[width=0.32\linewidth]{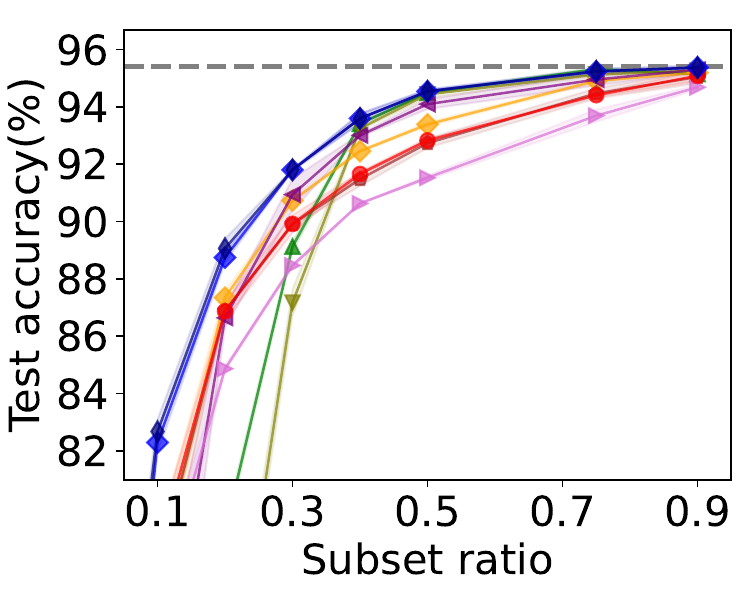}}
    \subfigure[CIFAR-100 \label{fig:pruning_CIFAR100}]{\includegraphics[width=0.32\linewidth]{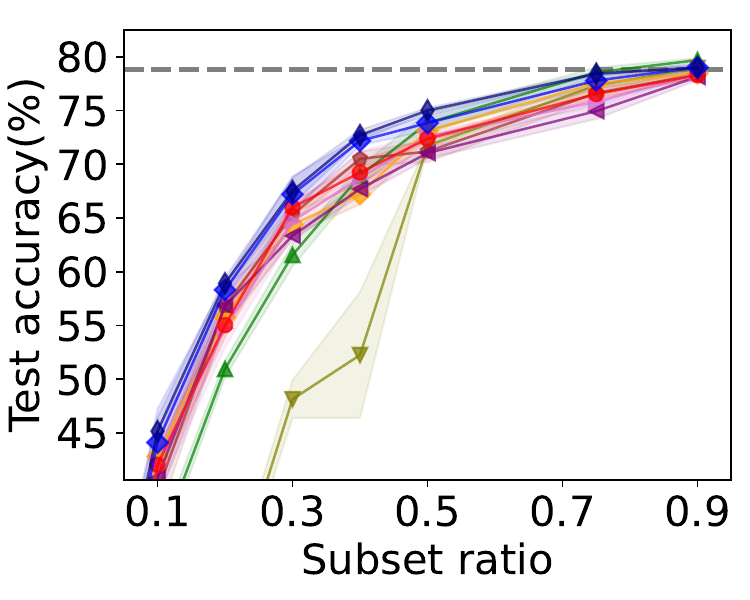}}
    \subfigure[ImageNet \label{fig:pruning_ImageNet}]{\includegraphics[width=0.32\linewidth]{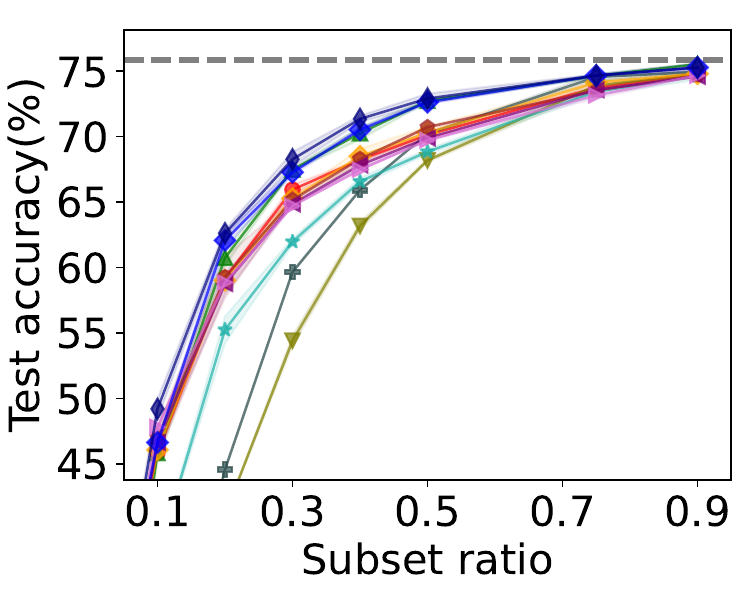}}
    \caption{
    (a, b, c) \textbf{Data pruning experiments.} Test accuracy of the models trained with data subsets of varying ratios in CIFAR-10/100, and ImageNet dataset, selected by different methods. Our method (BWS) outperforms other baselines across a wide range of selection ratios and achieves the accuracy as high as the Oracle window. Full results are reported in Table \ref{tab:CIFAR10_acc}--\ref{tab:ImageNet_acc}.
    }
    \label{fig:pruning_exp}
\end{figure*}

Our algorithm finds the best window subset by evaluating the performance of $ \rvw_{\rmS} $, corresponding to each window subset $\rmS$, on classifying the training samples $\{(\rvx_i, y_i)\}_{i=1}^n$ as described in Alg.  \ref{alg1}. To apply $ \rvw_{\rmS} $ for $C$-class classification problem, we find $ \rvw_{\rmS}(c)\in \mathbb{R}^{d+1}$ for each class $c\in\{1,\dots,C\}$, classifying whether a sample belongs to class $c$ or not, and simply place the vectors in columns of  $\rvw_\rmS\in \mathbb{R}^{(d+1)\times C}$. Then, we evaluate the performance of $ \rvw_{\rmS} $ by calculating the classification accuracy $\frac{1}{n}\sum_{i=1}^n \mathbbm{1}(\argmax_c(\rvw_{\rmS}^\top\rvf_i)_c=y_i)$ on the full training set.

In Table~\ref{tab:window_result}, we compare the performances of window subsets chosen by the sliding window experiment in Fig. \ref{fig:window_exp} and BWS (using the KRR as a proxy task) on CIFAR-10 dataset. We compare the starting points and test accuracies of the window subsets chosen by the two different methods for each subset ratio. We can observe that window subsets chosen by KRR align well with those chosen by the sliding window experiment. 
This observation demonstrates the effectiveness of our algorithm, which can efficiently replace the need to train models on each window subset and evaluate them on test dataset. 
BWS also finds the near optimal starting points for CIFAR-100 and ImageNet datasets across a broad range of subset ratios (from 1\% to 90\%). The detailed results are available in the Appendix \ref{sec:full_result}.

\paragraph{Feature extractor}
When $|\rmS|=m$, we randomly select $m$ samples from the full dataset, and use these samples to train a neural network for a few epochs to generate a feature extractor $f(\cdot)$. 
For CIFAR-10, we train ResNet18 for 20 epochs, and for CIFAR-100/ImageNet, we train ResNet50 for 20 epochs. The rationale behind training a feature extractor with random samples matching the window subset size is to simulate the situation where the model is trained with a limited window subset of the same size, enabling effective quality evaluation for window subsets.

\paragraph{Computational complexity}
The computational complexity of Algorithm \ref{alg1} includes training a feature extractor and solving the regression problem for $(\floor{(n-m)/t})$-subsets. Training the feature extractor is relatively efficient since it involves only a few epochs. Solving the regression requires matrix inversion, which takes $O(\min(d, m)^3)$ steps, with $d=512$ for ResNet18 and $2048$ for ResNet50.
This cost is significantly lower than other optimization-based baselines. For example, running BWS for the CIFAR-10 dataset with ResNet-18 and a step size of 5\% takes less than 11 seconds.
Detailed comparisons are provided in Appendix \ref{sec:cost}.

\section{Experiments}\label{sec:experiment}

To demonstrate the effectiveness of our method,  we conduct data pruning experiments. 
We select a subset of the dataset using each selection method while pruning the rest of the samples, and evaluate the performance of the model trained with each subset. We perform these experiments using ResNet18 for CIFAR-10 and ResNet50 for CIFAR-100 and ImageNet. Baselines include 1) two difficulty score-based selection: Forgetting and EL2N, 2) two optimization-based selection: AdaCore and LCMat, and 3) two  universal selection methods: Moderate DS score and CCS. 
We also add SSL Prototype \citep{beating} and memorization score \citep{memorization} as baselines on the ImageNet experiment, since these scores are known to achieve competitive performances especially on the large-scale datasets  \citep{beating}.
More details about the baselines and experiments are available in Appx. \ref{sec:exp_details}. The full experimental results are available in  Appx.  \ref{sec:full_result}.

\subsection{Experimental Results}\label{sec:pruning}
\paragraph{Data pruning experiments}

In Fig. \ref{fig:pruning_exp}, we present the test accuracies of models trained with data subsets of varying ratios, selected by different methods. The reported values are mean, and the shaded regions are std. across three (two) independent runs for CIFAR-10/100 (ImageNet). 
The Oracle window curve represents the results obtained using the window subset of the highest test accuracy found by the sliding window experiment as in Fig. \ref{fig:window_exp}, and BWS represents the results obtained using Alg. \ref{alg1}. We can observe that our method, BWS, consistently outperforms all other baselines across almost all selection ratios, and achieves the performance near the Oracle window. In the case of CIFAR-10/100, the difficulty score-based methods, Forgetting and EL2N, perform well in high ratio regimes but experience significant performance drop as the selection ratio decreases. The optimization-based methods, LCMat and AdaCore, achieve better performance than the difficulty score-based methods in low selection ratios but underperform in high selection ratios. Detailed numbers are reported in Table \ref{tab:CIFAR10_acc}--\ref{tab:ImageNet_acc}.

\begin{figure}[!tb]
\centering
    {\includegraphics[width=0.96\linewidth]{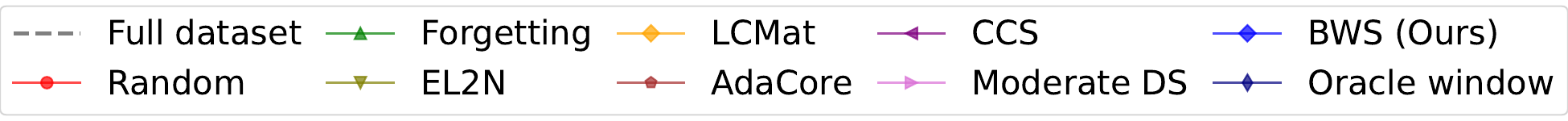}} \\
    \vspace{-0.6em}
    \subfigure[CIFAR-10 with ViT \label{fig:vit_main}]
    {\includegraphics[width=0.49\linewidth]{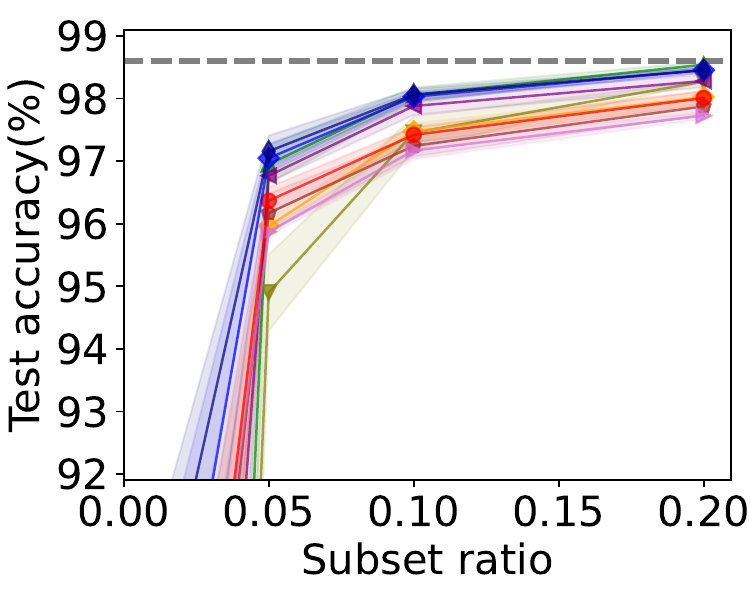}}
    \subfigure[Noisy CIFAR-10 \label{fig:noise_main}]{\includegraphics[width=0.49\linewidth]{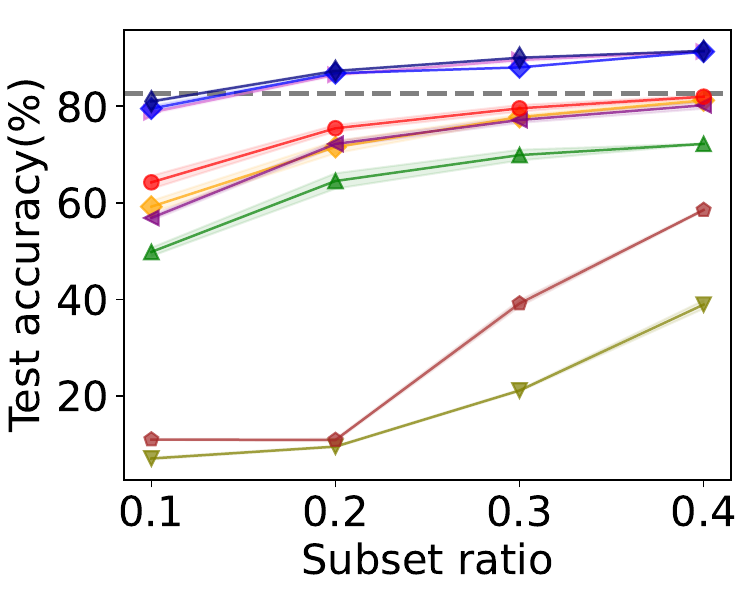}}
    \caption{(a) \textbf{Cross architecture experiment.} Test accuracy of the model fine-tuned with subsets of varying ratios in the CIFAR-10 dataset, selected by different methods. We utilize the Vision Transformer (ViT) architecture, pretrained on the ImageNet dataset. 
    (b) \textbf{Robustness to label noise.} Data pruning experiments with CIFAR-10, including 20\% label-noise.
    For both experiments, BWS surpasses other baselines for a wide range of selection ratios.
    }
    \label{fig:pruning_addition_exp}
\end{figure}

\paragraph{Cross-architecture experiments}
To test the robustness of our method across changes in model architectures, we conduct data pruning experiments on CIFAR-10 while using different architectures during sample scoring and training. 
Window subsets are constructed using samples ordered by their Forgetting scores, calculated on ResNet18, and then the best window selection (Alg. \ref{alg1}) and the model training are conducted using a simpler CNN/EfficientNet-B0 or a larger Vision Transformer (ViT) \citep{ViT}, pre-trained on the ImageNet. The results on the ViT are presented in Fig. \ref{fig:vit_main}, while those on CNN and EfficientNet-B0 are shown in Fig. \ref{fig:cross_arch_exp_app} of Appx. \ref{sec:cross_arch_exp_app}. In all cases, our method consistently achieves competitive performances across all selection ratios, demonstrating its robustness to changes in neural network architectures during data subset selection. 

\paragraph{Robustness to label noise}
Additionally, we demonstrate that BWS is robust against label noise in subset selection.
Existing sample selection methods, which rely on difficulty-based sample scores \citep{forgetting, EL2N}, are susceptible to a particular limitation: they often assign high scores to samples corrupted by label noise, as these samples are inherently hard to learn. 
This poses the risk of unintentionally selecting noisy samples during the selection phase.
On the contrary, our BWS algorithm adopts a different approach by solving a proxy task using kernel ridge regression rather than solely relying on high or low difficulty-based scores.
We test the robustness of BWS in the presence of label noise by corrupting randomly chosen 20\% samples of CIFAR-10 dataset by random label noise. To further enhance the robustness of our method, we modify Alg. \ref{alg1} to evaluate the solution of kernel ridge regression using only the low-scoring 50\% samples from the training dataset, which will rarely
include label-noise samples, instead of the full dataset. In Fig. \ref{fig:noise_main}, we compare the performance of this modified version of BWS with other baselines. While difficulty score-based selection and optimization-based selection methods suffer from performance degradation due to label noise, our method, along with another label noise-robust method, Moderate DS, achieves performance even higher than what is achievable with the full training dataset, which includes the 20\% label noise. 
Further experiment results with higher noise ratio are provided in Appendix \ref{sec:label_noise}.

\begin{figure}[!tb]
\centering
    \subfigure[Two half-width windows\label{fig:two_window_main}]{\includegraphics[width=0.49\linewidth]{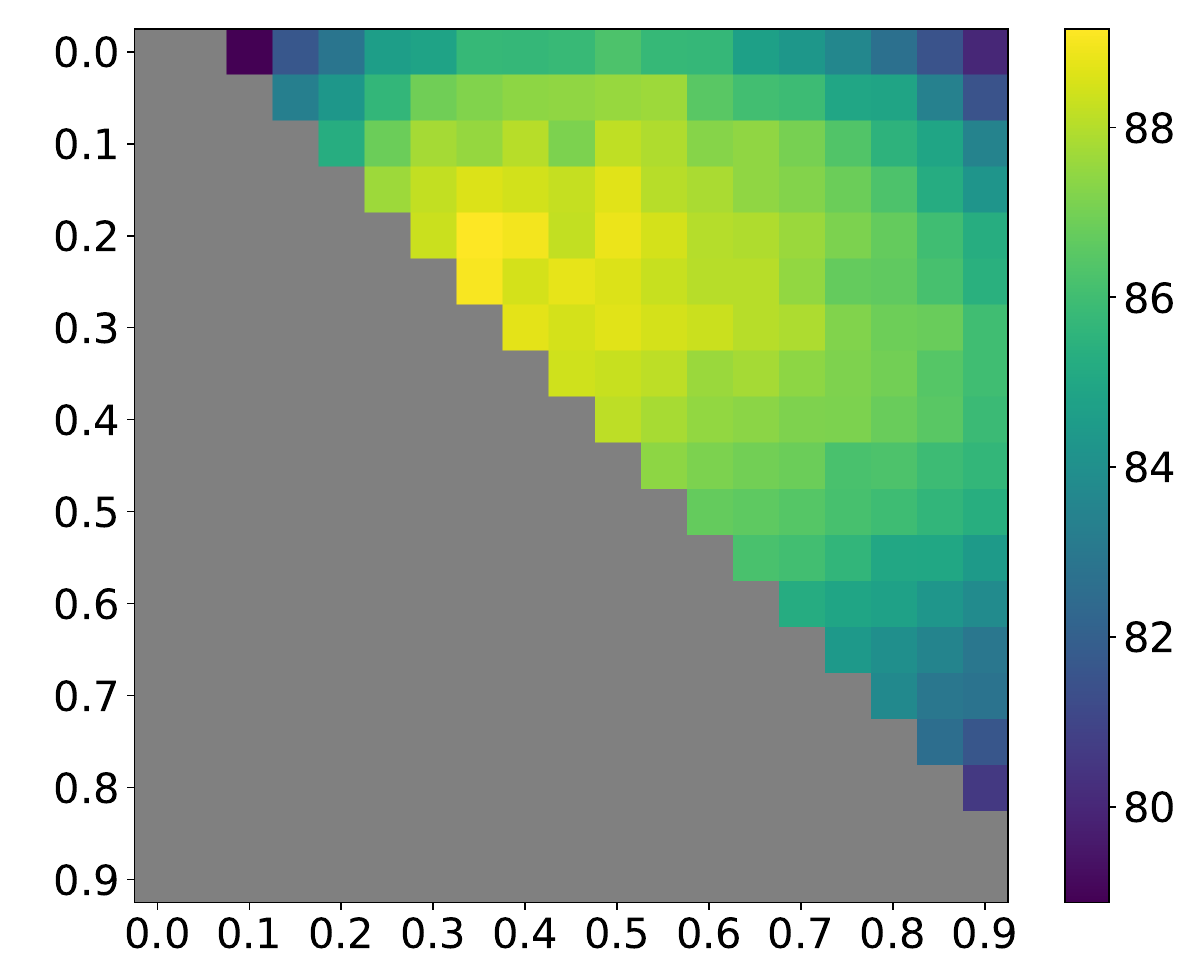}\label{fig:sub:two_half}}
    \subfigure[Wider window\label{fig:wider_window_main}]{\includegraphics[width=0.49\linewidth]{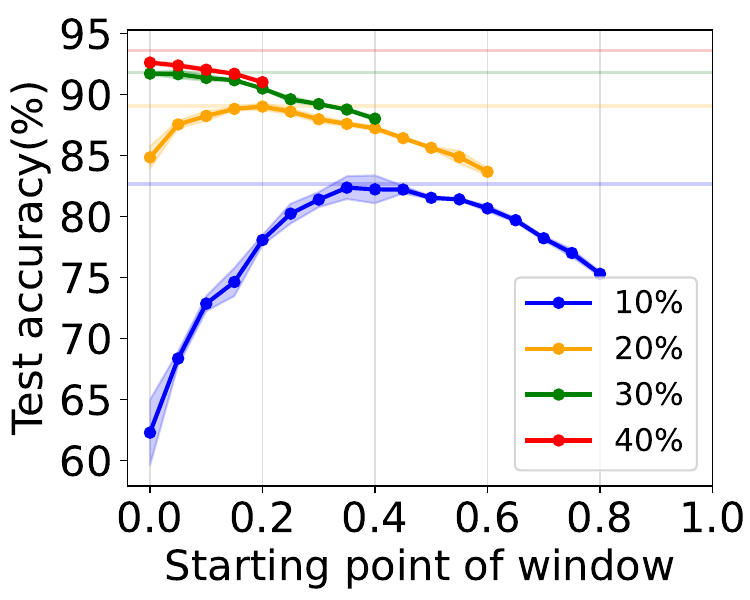}\label{fig:sub:wider}}
    \caption{
    (a) Test accuracies of the models trained with two half-width windows of varying starting points. Each axis indicates the starting points of each widow, and brighter color indicates higher accuracy. The best result is observed near the diagonal, contiguous windows.
    (b) Test accuracy of the models trained with wider windows. Horizontal lines are the results of oracle window subset. 
    At high ratio, oracle window outperforms wider windows. 
    }
    \label{fig:window_ablation}
\end{figure}

\subsection{Ablation study}
BWS operates by sorting training instances based on their difficulty scores, creating window subsets, and selecting the best window by a proxy task. To assess the importance of each component, we conduct several ablation studies.

\paragraph{Different types of window subsets}
Our method employs a window type that includes samples from a contiguous range of difficulty scores while changing the starting point. We explore two more generalized window types: a union of two half-width windows and a wider window where the samples are randomly selected from a wider range. For two half-width windows, given a subset of size $w$, we search over all combinations of two half-width windows, denoted by $[x_1,x_1+w/2] \cup [x_2,x_2+w/2]$, while varying their starting points $x_1 \in [0, 100-w]$ and $x_2 \in [x_1+w/2, 100-w/2]$ with a step size of $5\%$. For wider windows, we consider a window that is $c$ times wider than the subset size $w$, denoted as $[x_1, x_1+c\cdot w]$ while varying the starting point $x_1$ within the range $[0, 100-c\cdot w]$ with a step size of 5\%. 
These ablation studies are conducted with CIFAR-10 on ResNet18, to see whether the generality in subset selection can bring meaningful gain possibly at the cost of computation. 

In Table \ref{tab:diff_window}, we present the maximum test accuracies achieved by the two non-contiguous (two half-width/wider windows) and contiguous window types. Remind that two half-width window type includes all the contiguous windows. 
We observe that for every subset ratio, the performance of the best contiguous window subset almost matches that of two half-width windows, and outperforms wider widows. Moreover, Fig. \ref{fig:sub:two_half} shows that the best composition of two half-width windows occur when the two windows are close to each other (the diagonal positions in the figure). The sliding window experiment for wider windows in Fig. \ref{fig:sub:wider} shows that the best contiguous window (horizontal lines) achieves better performance than wider windows, especially in high ratios. 
 These results support our use of contiguous window subsets in choosing the near-optimal subset in a computationally efficient manner across a broad range of selection ratios. Further results are reported in Appx. \ref{sec:two_window}--\ref{sec:wider_window}.

\begin{table}[t]
\vspace{-0.7em}
\centering
\caption{The maximum test accuracy achieved by each window type in CIFAR-10. 
The best contiguous window nearly matches two half-width windows and outperforms wider windows. 
}\label{tab:diff_window}
\vspace{0.3em} 
\begin{small}
\resizebox{0.99\linewidth}{!}{
\begin{tabular}{c|ccc}
\toprule
\makecell{Selection \\ Ratio} & \makecell{Two half-width\\windows} & \makecell{Twice \\ wider window} & \makecell{Best contiguous \\ window} \\
\midrule
10\% & 83.04 & 82.37 & 82.67 \\
20\% & 89.16 & 89.01 & 89.06 \\
30\% & 92.02 & 91.72 & 91.80 \\
40\% & 93.67 & 92.62 & 93.59 \\
\bottomrule
\end{tabular}}
\end{small}
\end{table}

\paragraph{Different types of proxy task}

\begin{table}
\vspace{-0.7em}
\centering
\caption{Test accuracy of the models trained by window subsets of CIFAR-10 selected by different proxy tasks. Our method achieves the better performance, and the best window subsets selected by ours aligns better with those of oracle windows.
}
\label{tab:diff_proxy}
\vspace{0.5em} 
\begin{small}
\resizebox{0.99\linewidth}{!}{
\begin{tabular}{c|c|ccccccccc}
\toprule
Proxy task & Subset ratio & 1\% & 5\% & 10\% & 20\% & 30\% & 40\% & 50\% & 75\% & 90\% \\
\midrule
\multirow{2}{*}{SVP} & Test accuracy & 46.25 & 71.35 & 80.95 & 88.06 & 90.68 & 91.63 & 93.36 & 94.75 & 95.37 \\
    & Window index & 80\% & 65\% & 60\% & 40\% & 30\% & 25\% & 15\% & 5\% & 0\% \\
\midrule
Gradient  & Test accuracy & 39.45 & 70.40 & 82.24 & 88.42 & 90.68 & 91.63 & 92.72 & 94.30 & 94.82 \\
difference    & Window index & 50\% & 45\% & 40\% & 35\% & 30\% & 25\% & 20\% & 10\% & 5\% \\
\midrule
Gradient  & Test accuracy & 36.33 & 60.46 & 74.77 & 87.79 & 91.77 & 93.59 & 94.54 & 95.23 & 95.37 \\
  similarity  & Window index & 30\% & 25\% & 20\% & 15\% & 10\% & 5\% & 0\% & 0\% & 0\% \\
\midrule
\multirow{2}{*}{BWS} & Test accuracy & 46.10 & 70.70 & 82.29 & 88.74 & 91.80 & 93.59 & 94.54 & 95.23 & 95.37 \\
    & Window index & 90\% & 70\% & 55\% & 30\% & 15\% & 5\% & 0\% & 0\% & 0\% \\
\midrule
\multirow{2}{*}{Oracle} & Test accuracy & 47.17 & 72.89 & 82.67 & 89.06 & 91.80 & 93.59 & 94.54 & 95.23 & 95.37 \\
    & Window index & 85\% & 55\% & 50\% & 25\% & 15\% & 5\% & 0\% & 0\% & 0\% \\
\bottomrule
\end{tabular}}
\end{small}
\end{table}

We also evaluate the effectiveness of our proxy task, kernel ridge regression in Alg. \ref{alg1},  by comparing it with three different variants: 1) Selection via proxy (SVP) \citep{selectionviaproxy}, utilizing a smaller model (ConvNet) for choosing the best window, 2) Gradient $\ell_2$-norm difference, which finds a window subset minimizing the $\ell_2$-norm difference between the average gradients of the full dataset and the window subset, and 2) Gradient cosine similarity, which finds a window subset maximizing the cosine similarity between the average gradients of the full dataset and the window subset. The last two methods are inspired by gradient-matching strategies used in optimization-based coreset selection \citep{CRAIG, CREST}. Table \ref{tab:diff_proxy} presents the test accuracies achieved by models trained on window subsets selected by each method, along with the corresponding starting points of the chosen windows. The last row shows the result with the oracle window. Our method achieves better test accuracy compared to the three variants, and the window selected by our method aligns better with the oracle selection. In particular, SVP tends to select easier subsets possibly due to the limited capacity of the simple network used in the proxy task.
This result demonstrates that the best subset cannot be effectively chosen by using a simpler network or matching the average gradients; it requires a proxy task such as kernel ridge regression, with model-related kernels, to evaluate the quality of window subsets for classification tasks. We also perform an ablation study to show the robustness of our method across various difficulty scores  in Appx. \ref{sec:ablation_score}.

\section{Conclusion and Discussion}

We introduced the Best Window Selection (BWS), a universal and efficient data subset selection method capable of achieving competitive performance across a wide range of selection ratios. Our experimental results demonstrate that BWS effectively identifies the best window subset from samples ordered by difficulty-based scores, utilizing a simple proxy task based on kernel ridge regression. This method outperforms previous data subset selection approaches, which often excel within a limited range of selection ratios.

Subset selection has become a crucial technique in the big data era, allowing for the reduction of large datasets with minimal information loss. However, current efforts, including BWS, mainly focus on sample selection for supervised learning on curated datasets designed for classification tasks with well-defined labels. The next stage for subset selection may involve addressing challenges associated with much larger and more complex datasets. For instance, DataComp \citep{datacomp} proposes a new benchmark for subset selection by providing a web-scale multimodal dataset as the full training set. This setup challenges researchers to develop strategies for selecting subsets that benefit diverse downstream test sets capable of zero-shot generalization.

We believe that the insights gained through BWS--specifically, the shifts in the desired dataset characteristics based on selection ratio and the methodology for efficiently identifying the optimal subset using a simple proxy task--may provide valuable perspectives for designing data filtering or selection strategies for these large-scale datasets.


\section*{Impact Statement}

This paper addresses the performance degradation seen in existing data subset selection methods when the selection ratio varies widely. We introduce a methodology specifically designed to effectively counter this challenge. Our proposed universal data subset selection method delivers consistent, competitive performance across various selection ratios. This is particularly valuable in practical situations where computational and storage resources for training can vary, necessitating flexible sample selection based on the required subset ratios.

\section*{Acknowledgements}

This research was supported by the National Research Foundation of Korea under grant 2021R1C1C11008539.
\bibliography{main}
\bibliographystyle{icml2024}

\newpage
\appendix
\onecolumn
\section{Proof of Theoretical Analysis}\label{proof}

\subsection{Linear ridge regression}
The solution of the linear ridge regression problem is derived as follows.
\begin{align}
    L(\rvw)&=\|\rvy-\rmX^\top\rvw\|_2^2+\lambda\|\rvw\|_2^2\nonumber\\
    \dfrac{\partial L}{\partial \rvw} &= 2\rmX\rmX^\top\rvw - 2\rmX\rvy + 2\lambda\rvw = 0
    \quad \Rightarrow \quad \rvw = (\lambda\rmI + \rmX\rmX^\top)^{-1}\rmX\rvy\nonumber\\
   \therefore \rvw_\rmS &= (\lambda\rmI + \XS\XS^\top)^{-1}\XS\rvy_\rmS = \XS(\lambda\rmI + \XS^\top\XS)^{-1}\rvy_\rmS\nonumber
\end{align}

\subsection{Proof of Theorem \ref{thm:opt_reg}}\label{sec:app:proof}
In this section, we provide the detailed proof of Theorem \ref{thm:opt_reg} in Sec. \ref{sec:theory}.
We assume that $n=poly(d)$ data inputs $\rvx_1, \rvx_2, \dots \rvx_n$ are sampled from normalized multivariate normal distribution, $\mathcal{D}=\frac{1}{\sqrt{d}} \mathcal{N}(0,\rmI_d) = \frac{1}{\sqrt{d}} (\mathcal{N}_1, \mathcal{N}_2 \dots \mathcal{N}_d)$ where $\{\mathcal{N}_k\}$ are $i.i.d.$ normal distributions. Remind that the label $y_i$ of sample $\rvx_i$ is determined by the sign of its first element, i.e., 
 if $(\rvx_i)_1>0$ then $y_i=1$, and if $(\rvx_i)_1<0$, then $y_i=-1$.
We select a subset of size $m$, denoted by $(\rmX_{\rmS}, \rvy_\rmS)\in\mathbb{R}^{d\times m}\times \{-1,1\}^m$. 

We first provide a high-level proof idea of Theorem \ref{thm:opt_reg}.
Note that the optimal $\rvw_{\rmS} = \argmin_{\rvw\in \mathbb{R}^d} \|\rvy_{\rmS}-\rmX_{\rmS}^\top\rvw\|_2^2$ can be written as $\rvw_{\rmS}= \XS(\XS^\top\XS)^{-1}\rvy_{\rmS}$ when $m\leq d$, and $\rvw_{\rmS}= (\XS\XS^\top)^{-1} \XS \rvy_{\rmS}$ when $m\geq d$. Let us first consider the case of $m\leq d$. Due to the properties of high dimensional multivariate normals, we have $\|\rvx_i\|\in [1\pm \sqrt{7\ln n/2d}]$ for all $i\in[n]$ and $|\rvx_i^\top \rvx_j|\leq \sqrt{7\ln n/2d}$ for all $i\neq j\in[n]$ with high probability. Thus, $\|\XS^\top\XS- \rmI_m\|_F\leq \sqrt{\frac{m^2(7\ln n)}{2d}}$ where $\rmI_m$ is the identity matrix of size $m$. When $m\ll \sqrt{d/\ln d}$, we have $(\XS^\top\XS)\approx (\XS^\top\XS)^{-1}\approx \rmI_m$, and thus $\rvw_{\rmS}= \XS(\XS^\top\XS)^{-1}\rvy_{\rmS}\approx \XS \rvy_{\rmS}$, which implies that $(\rvw_{\rmS})_1 \approx \sum_{i=1}^m|(\rvx_i)_1|$.
Let us next consider the case of $m\geq d$. Note that the diagonal terms of $\XS\XS^\top$ are $\sum_{i=1}^m |(\rvx_i)_k|^2=\Theta(m/d)$ for $k\in[d]$ and the off-diagonal terms are $\sum_{i=1}^m (\rvx_i)_{k} (\rvx_i)_{l}=O(\sqrt{m\ln d}/d)$ for $k\neq l\in[d]$ with high probability. The eigenvalues of $\XS\XS^\top$ can be shifted from its diagonal entries $(\sum_{i=1}^m |(\rvx_i)_1|^2,\dots, \sum_{i=1}^m |(\rvx_i)_d|^2)$  by at most $\frac{\sqrt{m\ln d}}{d}d=\sqrt{m\ln d}$ by the effect of its off-diagonal entries. Thus, when $m/d\gg \sqrt{m\ln d}$, i.e., $m\gg d^2\ln d$, we can have $\XS\XS^\top\approx \rm{diag}(\sum_{i=1}^m |(\rvx_i)_1|^2,\dots, \sum_{i=1}^m |(\rvx_i)_d|^2)$ and  $(\XS\XS^\top)^{-1}\approx \rm{diag}((\sum_{i=1}^m |(\rvx_i)_1|^2)^{-1},\dots, (\sum_{i=1}^m |(\rvx_i)_d|^2)^{-1})$. Since $\rvw_{\rmS}= (\XS\XS^\top)^{-1} \XS \rvy_{\rmS}$, the first coordinate value of $\rvw_{\rmS}$ is $(\rvw_{\rmS})_1 \approx (\sum_{i=1}^m|(\rvx_i)_1|)/(\sum_{i=1}^m|(\rvx_i)_1|^2)$.

To more formally state and prove Theorem \ref{thm:opt_reg}, we provide Theorem \ref{thm:poorcondition} to explain the regime of low selection ratio ($m = o( \sqrt{d/\ln d})$) and Theorem \ref{thm:richcondition} for the high selection ratio ($m = \omega(d^2\ln{d})$).
To prove the two theorems, we use the following three lemmas, including the tail bounds on chi-square and Gaussian distributions, and Gershgorin theorem, which are stated as below:

\begin{lemma}[Chi-square tail bound]\label{lemma:chi-square}
    If $\rvx \sim \chi^2(d)$, then $\mathbb{P}(\chi^2(d) \geq d + 2\sqrt{dt} + 2t) \leq e^{-t}$ and $\mathbb{P}(\chi^2(d) \leq d - 2\sqrt{dt}) \leq e^{-t}$.  
\end{lemma}

\begin{lemma}[Gaussian tail bound]\label{lemma:gaussiantail}
     If $\rvx \sim \mathcal{N}(0, 1)$, then $\mathbb{P}(|\rvx| \geq t)\leq e^{\frac{-t^2}{2}}$.
\end{lemma}

\begin{lemma}[Gershgorin circle theorem]\label{lemma:Gershgorin}
Let $A\in\mathbb{C}^{d\times d}$ be a matrix with its $(i,j)$-th entry equal to $a_{ij}$. Let $r_i := \sum_{j\neq i}|a_{ij}|$ and $D_i:=D_{r_i}(a_{ii})$ be a closed ball centered $a_{ii}$ with radius $r_i$. Then, every eigenvalue of $A$ is contained in $\cup_{i}D_i$
\end{lemma}

Gershgorin circle theorem restricts the eigenvalues of a matrix in a union of disks, whose centers are diagonal elements, and the radius is the sum of off-diagonal elements.

Now, we provide Theorem \ref{thm:poorcondition}, which will be used to explain why selecting low-scoring (easy) data samples results in a good performance when the subset size $|\rmS|$ is small.
\begin{thm}[Sample-deficient regime]\label{thm:poorcondition}
If $m = o\left(\sqrt{d/\ln d}\right)$, then $\|(\XS^\top\XS)^{-1}-\rmI_m\|_2 \leq  m\sqrt{\frac{7\ln{n}}{2d}}$ with high probability as $d\to\infty$.
\end{thm}
\begin{proof}
At first, we prove two properties of the high dimensional multivariate normal distribution, which state that the norm of every $\rvx_i$ is almost equal to $1$, and every two independent vectors are almost orthogonal for large enough $d$.
For any $1 \leq i \neq j \leq n$, with probability $1-O(\frac{1}{n})$, we have
\begin{align}
&1-\sqrt{\dfrac{7\ln{n}}{2d}} \leq \|\rvx_i\|_2 \leq 1+\sqrt{\dfrac{7\ln{n}}{2d}}, \;\;\text{and}\label{eqn:norm}\\
&|\rvx_i^\top \rvx_j| < \sqrt{\dfrac{7\ln{n}}{2d}}.\label{eqn:inner}
\end{align}
The first property (Eq. \ref{eqn:norm}) can be proved by Lemma \ref{lemma:chi-square}.
Let $t=3\ln{n}$ for Lemma \ref{lemma:chi-square}. Then,
\beq 
\mathcal{P}(\chi^2(d) \geq d + 2\sqrt{3d\ln{n}}+6\ln{n}) \leq \frac{1}{n^3} \quad \text{and} \quad 
\mathcal{P}(\chi^2(d) \leq d - 2\sqrt{3d\ln{n}}) \leq \frac{1}{n^3}. \nonumber
\eeq 

Since $2\sqrt{3d\ln{n}}+6\ln{n} \leq \sqrt{13d\ln{n}}$ for large enough $d$, with probability $1-O(\frac{1}{n^3})$ we have
\beq\label{eq:sandwich}
1-\sqrt{\frac{13\ln{n}}{d}} \leq \frac{1}{d}\chi^2(d) \leq 1+\sqrt{\frac{13\ln{n}}{d}}
\; \xRightarrow{\text{d $\rightarrow \infty$}} \;
1-\sqrt{\frac{7\ln{n}}{2d}} \leq \sqrt{\frac{1}{d}\chi^2(d)} \leq 1+\sqrt{\frac{7\ln{n}}{2d}}.
\eeq

Since $\rvx_i \sim \frac{1}{\sqrt{d}}\mathcal{N}(0, \rmI_d)$ and $\|\rvx_i\|_2^2 = \frac{1}{d}\chi^2(d)$, for $\forall i \in [n]$,  with probability $ 1-O(\frac{1}{n^2})$, Eq. \ref{eqn:norm} follows.

The proof of the second property (Eq. \ref{eqn:inner}) also utilizes Lemma \ref{lemma:chi-square}.
Let $\rvx_i=\dfrac{1}{\sqrt{d}}(\mathcal{N}_{i1}, \mathcal{N}_{i2}, \dots \mathcal{N}_{id})$ and $\rvx_j=\dfrac{1}{\sqrt{d}}(\mathcal{N}_{j1}, \mathcal{N}_{j2}, \dots \mathcal{N}_{jd})$ where $\mathcal{N}_{ik}, \mathcal{N}_{jk}$ are $i.i.d.$ normals $\mathcal{N}(0, 1)$. Then,
\begin{align*}
\rvx_i^{\top}\rvx_j & = \frac{1}{d}\sum_{k=1}^d {\mathcal{N}_{ik}}{\mathcal{N}_{jk}} = \frac{1}{d}\sum_{k=1}^d{\frac{({\mathcal{N}_{ik}+ {\mathcal{N}_{jk}}})^2-({\mathcal{N}_{ik}-{\mathcal{N}_{jk}}})^2}{4}}\\
& = \frac{1}{2d}\sum_{k=1}^d\left[\left(\frac{\mathcal{N}_{ik}+\mathcal{N}_{jk}}{\sqrt{2}}\right)^2-\left(\frac{\mathcal{N}_{ik}-\mathcal{N}_{jk}}{\sqrt{2}}\right)^2\right] = \frac{1}{2d}\sum_{k=1}^d[(\mathcal{N}_{k}^{'})^2-(\mathcal{N}_{k}^{''})^2]\\
& = \frac{1}{2d}(\chi_1^2(d)-\chi_2^2(d)),     
\end{align*}
where $\mathcal{N}_{k}^{'}$ and $\mathcal{N}_{k}^{''}$ are $i.i.d.$ normals, and $\chi_1^2(d)$ and $\chi_2^2(d)$ are $i.i.d$ chi-squares. 

As shown in Eq. \ref{eq:sandwich}, with probability $1-O(\frac{1}{n^3})$, 
\beq
1-\sqrt{\frac{7\ln{n}}{2d}} \leq  \sqrt{\frac{1}{d}\chi_1^2(d)} \quad \text{and} \quad \sqrt{\frac{1}{d}\chi_2^2(d)}  \leq  1+\sqrt{\frac{7\ln{n}}{2d}}.
\eeq
Thus, we have
\beq
\left|\frac{1}{2d}({\chi_1^2(d)-\chi_2^2(d)})\right| \leq \sqrt{\frac{7\ln{n}}{2d}}.
\eeq
By applying a union bound, for $\forall i \neq j \in [n]$,  with probability $ 1-O(\frac{1}{n})$, we have
$
|\rvx_i^\top \rvx_j| \leq \sqrt{\frac{7\ln{n}}{2d}}\nonumber.
$
From Eq. \ref{eqn:norm} and Eq. \ref{eqn:inner}, we obtain that $\|\XS^\top\XS-\rmI_m\|_F^2 \leq m^2\left(\frac{7\ln{n}}{2d}\right)$.

Let $\rmA = \XS^\top\XS$, then we derive the bound on $\|\rmI-\rmA^{-1}\|_2$ from the bounds of $\|\rmI-\rmA\|_2$ and $\|\rmA^{-1}\|_2$. First, note that
\beq
\|\rmI-\rmA\|_2 \leq \|\rmI-\rmA\|_F \leq m\sqrt{\frac{7\ln{n}}{2d}} \quad \text{and} \quad m\sqrt{\frac{7\ln{n}}{2d}} \rightarrow 0
\text{ as } d\rightarrow \infty\nonumber
\eeq
since $m = o(\sqrt{d/\ln d})$ and $n=poly(d)$. Moreover, we have
\beq
\begin{split}
&\|\rmA^{-1}\|_2 =\|(\rmI-(\rmI-\rmA))^{-1}\|_2 = \|\rmI+(\rmI-\rmA)+(\rmI-\rmA)^2+\dots\|_2\\
&\leq \|\rmI\|_2+\|(\rmI-\rmA)\|_2+\|(\rmI-\rmA)^2\|_2+\dots \leq 1+\sum_{k=1}^\infty\left(m\sqrt{\frac{7\ln{n}}{2d}}\right)^k \leq 2\nonumber.
\end{split}
\eeq
Finally, we have
\beq
\|\rmI-\rmA^{-1}\|_2 \leq \|\rmA^{-1}\|_2\|\rmI-\rmA\|_2 \leq m\sqrt{\frac{7\ln{n}}{2d}}.\nonumber
\eeq
\end{proof}

We next provide Theorem \ref{thm:richcondition}, which explains why selecting high-scoring (difficult) data samples results in a good performance when the subset size $|\rmS|$ is large ($m = \omega(d^2\ln{d})$).
Assume that we select the subset $\XS$ by observing the first element of each data, $(\rvx_i)_1$.
Suppose that we select the data samples whose first elements are $\frac{a_1}{\sqrt{d}}, \frac{a_2}{\sqrt{d}}, \dots \frac{a_m}{\sqrt{d}}$ where $a_i \in \Theta(1)$, and let $a := \frac{\sum_{i=1}^m a_i^2}{m}$.
The elements of the other coordinates are independent normals, i.e., $(\rvx_i)_k \sim \mathcal{N}(0,1)$
for $k \geq 2$. We will prove that $(\frac{d}{m}\XS\XS^\top)^{-1}$ can be approximated by a diagonal matrix of which the first element is equal to $\frac{1}{a}$ and other elements are equal to 1.

\begin{thm}[Sample-sufficient regime]\label{thm:richcondition}
Let $\rmB = \rm{diag}(a, 1, 1, \dots 1) \in \mathbb{R}^{d\times d}$. If $m = \omega(d^2\ln{d})$, then
$\|(\frac{d}{m}\XS\XS^\top)^{-1}-\rmB^{-1}\|_2 \leq c'd^2\dfrac{\ln{d}}{m}$ for some constant $c'>0$ with high probability as $d\to\infty$.
\end{thm}
\begin{proof}
The elements of $\frac{d}{m}\XS\XS^\top$ are expressed as follows, where $k \neq l \in [d]\text{\textbackslash}\{1\}$:

\begin{align*}
    & \frac{d}{m}(\XS\XS^\top)_{11} = \frac{1}{m}\sum_{i=1}^m a_i^2 = a\\
    & \frac{d}{m}(\XS\XS^\top)_{k1} = \frac{1}{m}\sum_{i=1}^m a_i\mathcal{N}_i=\mathcal{N}\left(0, \frac{\sum_{i=1}^m a_i^2}{m^2}\right) = \mathcal{N}\left(0, \frac{a}{m}\right)\\
    & \frac{d}{m}(\XS\XS^\top)_{kk}= \frac{d}{m}\sum_{i=1}^m(\rvx_i)_k^2=\frac{1} {m}\mathcal{N}_{ik}^2=\frac{1}{m}\chi^2(m) \\
    & \frac{d}{m}(\XS\XS^\top)_{kl} = \frac{d}{m}\sum_{i=1}^m(\rvx_i)_k(\rvx_i)_l=\frac{1}{m}\sum_{i=1}^m\mathcal{N}_{ik}\mathcal{N}_{il}. 
\end{align*}

By Gaussian tail bound (Lemma \ref{lemma:gaussiantail}), 
if $\rvx \sim \mathcal{N}(0, {a}/{m})$, then we have
\beq
\mathbb{P}(|\rvx| \geq 2\sqrt{{a\ln{d}}/{m}}) \leq {1}/{d^2}.\nonumber
\eeq
Note that $\frac{1}{m}\chi^2(m)= \frac{1}{m}\|\rvx\|_2^2$ for  $\rvx \sim \mathcal{N}(0,\rmI_m)$. By Lemma \ref{lemma:chi-square}, we have a result similar to Eq. \ref{eqn:inner},
\beq
\mathbb{P}(|\|\rvx\|_2-1| \geq \sqrt{{7\ln{d}}/{2m}}) \leq {1}/{d^2}.\nonumber
\eeq
For $\mathcal{N}_{ik}\mathcal{N}_{il}$, by applying the result of Eq. \ref{eqn:inner}, we can also show that
\beq
\mathbb{P}(\|\rvx\|_2 \geq \sqrt{{7\ln{d}}/{2m}}) \leq {1}/{d^3}.\nonumber
\eeq
Combining the above three bounds, we obtain that for $\forall k \neq l \in [d]$, with probability $1-O(\frac{1}{d})$,
\beq
\left|\frac{d}{m}(\XS\XS^\top)_{k1}\right| \leq 2\sqrt{\frac{a\ln{d}}{m}}, \quad
    \left|\frac{d}{m}(\XS\XS^\top)_{kk}-1\right| \leq \sqrt{\frac{7\ln{d}}{2m}}, \; \text{and} \quad \left|\frac{d}{m}(\XS\XS^\top)_{kl}\right| \leq \sqrt{\frac{7\ln{d}}{2m}}.\nonumber
\eeq
Thus, we obtain $\|\frac{d}{m}(\XS\XS^\top)-\rmB\|_F \leq cd^2\frac{\ln{d}}{m}$ for some constant $c>0$, with probability $1-O(\frac{1}{d})$.
Let $\rmA=\frac{d}{m}(\XS\XS^\top)$, then
\beq
\begin{split}
\|\rmA^{-1}-\rmB^{-1}\|_2 &\leq \|\rmA^{-1}\|_2\|\rmI-\rmA\rmB^{-1}\|_2 \leq \|\rmA^{-1}\|_2\|\rmB-\rmA\|_2\|\rmB^{-1}\|_2\nonumber\\ 
    &\leq \|\rmA^{-1}\|_2\|\rmB-\rmA\|_F\|\rmB^{-1}\|_2 \leq \|\rmA^{-1}\|_2 \;cd^2\dfrac{\ln{d}}{m} \cdot 1.
 \end{split}
\eeq
It is remaining to prove that $ \|\rmA^{-1}\|_2$ is bounded.  The eigenvalues of $\rmA = \frac{d}{m}(\XS\XS^\top)$ are almost equal to the diagonal elements by utilizing Gershgorin circle theorem.
Since $m \in \omega(d^2\ln{d})$, 
\begin{center}
    $r_1=\sum_{j\neq i}|a_{1j}| < 2d\sqrt{\frac{a\ln{d}}{m}} \ll a$ \quad and \quad $r_k=\sum_{j\neq i}|a_{kj}| < d \sqrt{\frac{7\ln{d}}{2m}} \ll 1$.
\end{center}
Therefore, every eigenvalues of $\rmA$ are close to either $a$ or $1$ with probability $1-O(\frac{1}{d})$.
Thus, $\|\rmA^{-1}\|_2$ is bounded above by $\max(\frac{1}{a}, 1)$ plus some some constant, which shows that $\|\rmA^{-1}\|_2$ is bounded. Therefore, we have
\beq
\|\rmA^{-1}-\rmB^{-1}\|_2 \leq \|\rmA^{-1}\|_2 \; cd^2\dfrac{\ln{d}}{m} \leq c'd^2\dfrac{\ln{d}}{m} \quad \text{ for some constant } c'>0.    \nonumber
\eeq
\end{proof}

\subsection{Toy experiment}\label{sec:toy}
\begin{figure*}[!tb]
\centering
    \subfigure[Plots for entire regime]
    {\includegraphics[width=0.49\linewidth]{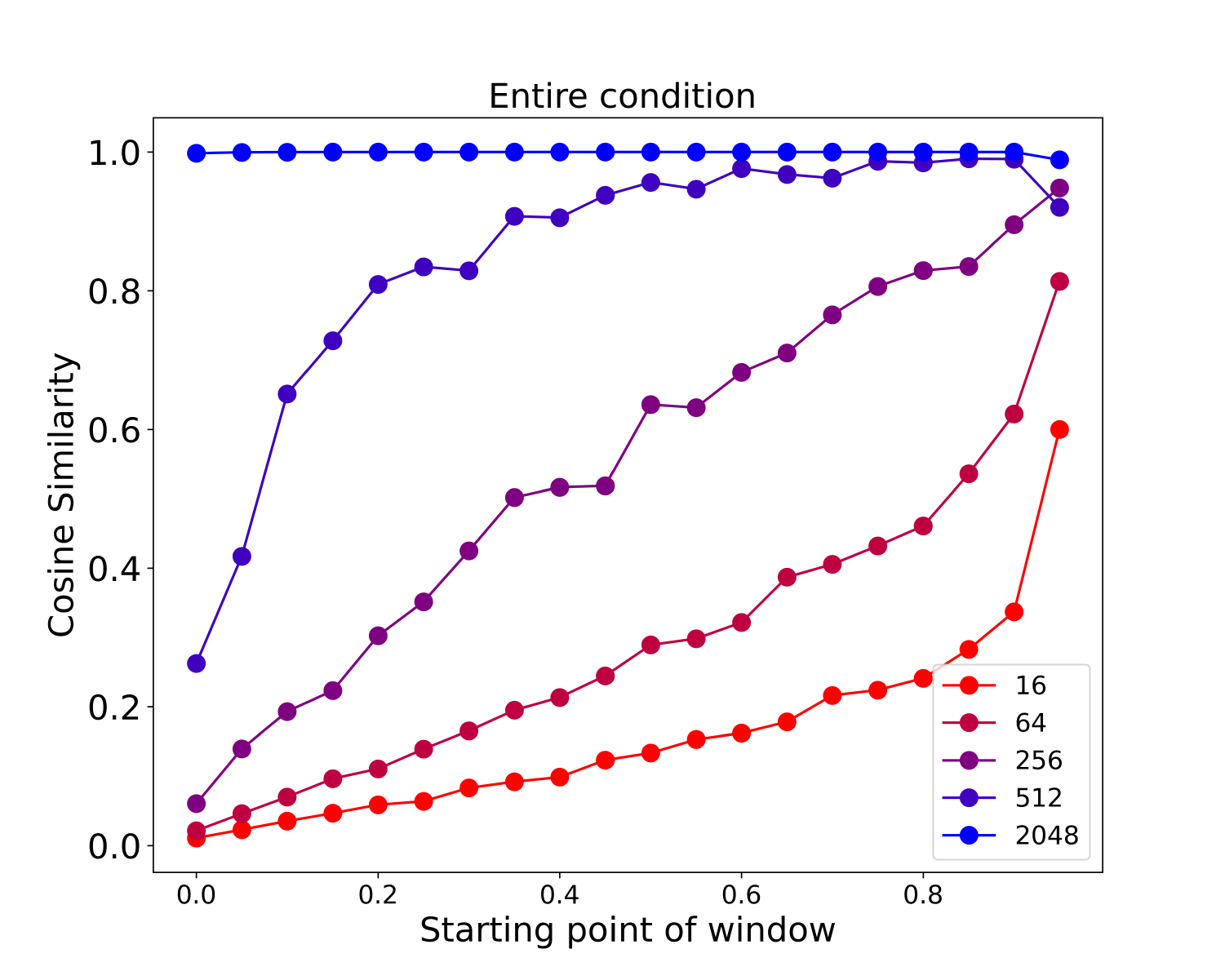}\label{fig:entire}}
    \subfigure[Plots focused on sample-sufficient regime]
    {\includegraphics[width=0.49\linewidth]{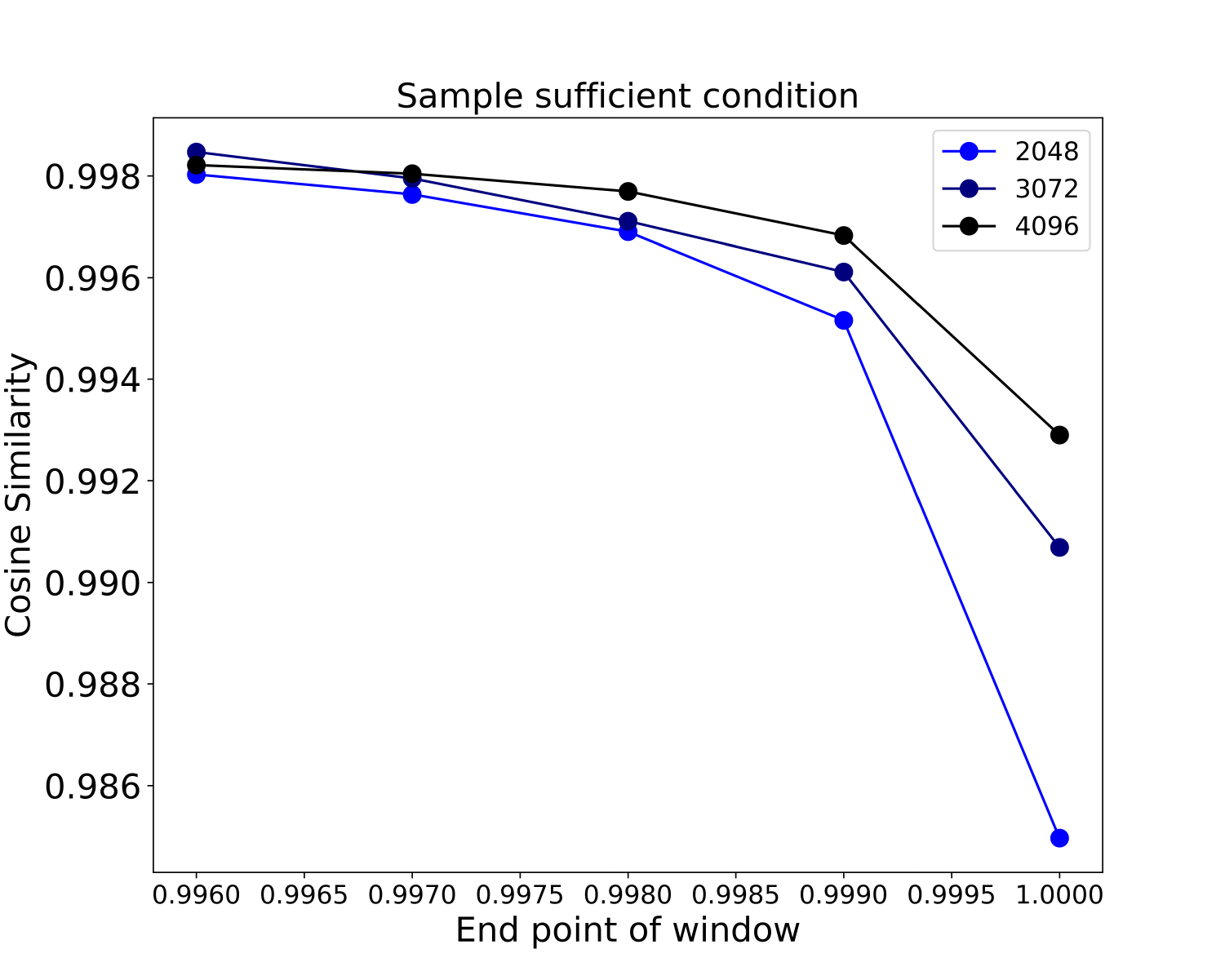}\label{fig:sufficient}}
    \caption{Results of window sliding experiment at the setting of theoretical analysis. In our setting, the dimension $d$ is $256$, the number of full dataset $n$ is $256,000$, and the subset size $m$ are selected among $16, 64, 256, 512, 2048, 3072, \text{and } 4096$.
    Left figure covers the entire regime ($m=16,64,256,512, \text{and } 2048$), while the right figure focuses on sample-sufficient regime.}
    \label{fig:toy}
\end{figure*}

To validate our theoretical analysis, we conduct a window sliding experiment similar to the one in Sec. \ref{sec:methodology}, at the setting of the theoretical analysis  in Sec. \ref{sec:app:proof}, while varying the subset sizes and the starting points of the window subsets at $d=256$ and $n=256,000$. The results are shown in Fig. \ref{fig:toy}. Fig. \ref{fig:entire} shows the plot for both the sample-deficient and sufficient regimes, including $m=16,64,256,512,2048$, while Fig. \ref{fig:sufficient} shows focused plots for sample sufficient regime where $m=2048,3072, 4096$. 
The x-axis in Fig. \ref{fig:entire} is the starting point of the window subset, which identifies the ranking of the hardest sample in the window subset, while that in Fig. \ref{fig:sufficient} is the end point of the window subset, which identifies the ranking of the easiest sample in the window subset.
The y-axis is the cosine similarity between $\rvw$ and $\hat{e}_1=(1,0,\dots, 0)$, where $\rvw$ is the solution of the regression problem, and $\hat{e}_1$ is the unit vector with its first coordinate equal to 1, which is the true decision boundary. A higher cosine similarity implies a better solution.
Red lines show the results when subset size $m$ is smaller, and the blue or black lines show the result of larger subset sizes.

In the sample-deficient regime where $m \leq d$ (red lines), the cosine similarity increases as the starting point of the window increases, meaning that it is better to use easy samples to learn the linear classifier. On the other hand, in the sample-sufficient regime where $m > d$ (blue and black lines), the cosine similarity is larger for windows having a smaller end point, meaning that it is better to include difficulty samples to learn a better classifier. 
This result coincides with the theoretical analysis, which claims that the inclusion of easier (harder) data samples results in a better solution for a smaller (larger) subset size, respectively.
As we conjectured at Sec. \ref{sec:theory}, the transition of a desirable selection strategy occurs near $m =\Theta(d)$.

\section{Discussions on Using Kernel Ridge Regression as a Proxy}\label{justification_krr}

In Algorithm~\ref{alg1}, we use the kernel ridge regression as a proxy for training neural networks to evaluate the performance of window subsets. In this section, we provide some theoretical rationale behind the use of the kernel ridge regression. 

Our use of the kernel ridge regression as a proxy for training neural networks can be partly explained by the recent progress in theoretical understanding of training neural networks using kernel methods. In particular, some recent works \citep{neal2012bayesian,lee2018deep, jacot2018neural, Equivalence_NN_KRR} have shown that training and generalization of neural networks can be approximated by two associated kernel matrices: the Conjugate Kernel (CK) and Neural Tangent Kernel (NTK). The Conjugate Kernel is defined by the gram matrix of the derived features produced by the final hidden layer of the network, while NTK is the gram matrix of the Jacobian of in-sample predictions with respect to the network weights. These two kernels also have fundamental relations in terms of their eigenvalue distributions as analyzed in \cite{CKandNTK}. Our proxy task is motivated by the observation that the kernel regression with these model-related kernels can provide a good approximation to the original model (under some assumptions such as enough width, random initialization, and small enough learning rate, etc.). 
As an example, the work by \citet{Equivalence_NN_KRR} provides the following theorem, which connects the training of a neural network with kernel ridge regression using NTK. 

\begin{thm}[Informal version of \citet{Equivalence_NN_KRR}]\label{thm:equivalence}
Consider a fully connected neural network with sufficiently large width $d_1=d_2= \dots d_L$ where $d_l$ is the number of nodes in $l$th layer. Given a training dataset $\{(\rvx_i, y_i)\}_{i=1}^n \subset \mathbb{R}^d \times \mathbb{R}$ with normalized inputs $\|\rvx_i\|_2=1$, the network is trained by gradient descent with a sufficiently small learning rate to minimize the square-loss $\sum_{i=1}^n(f_{nn}(\rvx_i)-y_i)^2$, where
$f_{nn}$ is the trained network.
With a kernel function of the network $K(\cdot, \cdot)$, NTK of training data $\mathbf{H} \in \mathbb{R}^{n\times n}$ is defined by $\mathbf{H}_{ij}=K(\rvx_i, \rvx_j)$. And, for a test data $\rvx_{te}$, the kernel between the test data and the training dataset $\rmX = [\rvx_1, \rvx_2, \dots \rvx_n] \in \mathbb{R}^{d \times n}$ is defined by $\rmK(\rvx_{te}, \rmX) \in \mathbb{R}^n$ where $\rmK(\rvx_{te}, \rmX)_i = K(\rvx_{te}, \rvx_i)$.
Let $f_{ntk}(\rvx_{te}) = (\rmK(\rvx_{te}, \rmX))^\top\mathbf{H}^{-1}\rvy$. Then,
\begin{center}
    $|f_{nn}(\rvx)-f_{ntk}(\rvx)| \leq \epsilon$
\end{center}
\end{thm}

Theorem \ref{thm:equivalence} justifies that the kernel regression with NTK can provide a good approximation to the neural network training under the specified assumptions. However, calculating the NTK for the entire neural network requires high computational cost as it involves computing the Jacobian with respect to the network weights.

To address such a problem, Conjugate Kernel is often considered as a promising replacement of NTK. For example, the work by \citet{FRePo} utilizes the kernel ridge regression based on Conjugate Kernel for dataset distillation.
We also use the Conjugate Kernel in our kernel ridge regression (Eq. \ref{eq:regression}), by defining the kernel matrix as $\rmX_{\rmS}^\top\rmX_{\rmS}$ where $\mathbf{X}_\mathbf{S}=[\mathbf{f}_1,\dots,\mathbf{f}_m]$ is composed of features produced by the exact target network of our consideration (ResNet18 for CIFAR-10 and ResNet50 for CIFAR-100/ImageNet). By considering the features from the target network, we can obtain the (approximate) network predictions that are linear in these derived features. 
In detail, the output of the CK-regression for the test example $\rvx_{te}$ can be written as  $f_{ntk}(\rvx_{te})=\rvx_{te}^\top\XS(\XS^\top\XS)^{-1}\rvy_{\rmS}=\rvx_{te}^\top\rvw$ where $\rvw=\XS(\XS^\top\XS)^{-1}\rvy$ in Eq. \ref{eq:solution} for $\lambda=0$.

Of course, this kernel approximation of the neural network models, which assumes a fixed feature extractor, does not exactly match our situation where the selected subset not only affects the linear classifier but also the feature extractor itself during the training. However, this is still a good proxy that can reflect the network architecture of our interest in a computationally-efficient manner. Also, our analysis in Table \ref{tab:window_result} shows that this proxy finds the best window subset that aligns well with the result from the actual training of the full model.


\section{Implementation Details}\label{sec:exp_details}
\subsection{Baseline details}
We benchmark our BWS algorithm against eight different state-of-the-art methods, Forgetting score \citep{forgetting}, EL2N score \citep{EL2N}, AdaCore \citep{Adacore}, LCMat \citep{LCMat}, Moderate DS \citep{mds}, CCS \citep{CCS}, SSL Prototype score \citep{beating}, and Memorization score \citep{memorization}.

In Forgetting and EL2N scores, the scores are derived by averaging the results of five independent training runs using the full CIFAR-10/100 dataset. Specifically, Forgetting scores are obtained at the 200th epoch (full training), while EL2N scores are captured at the 20th epoch. 
For our ImageNet experiments, pre-calculated Forgetting, EL2N, SSL prototype, and Memorization scores are sourced from \href{https://github.com/rgeirhos/dataset-pruning-metrics}{https://github.com/rgeirhos/dataset-pruning-metrics} \citep{beating}.

In the AdaCore and LCMat methodology, subset selection is conducted only once at the 10th epoch, to ensure a fair comparison to other baselines. Both AdaCore and LCMat implementations are sourced from \href{https://github.com/SJShin-AI/LCMat}{the LCMat repository}.
For Moderate DS, models are trained using the full dataset, and the individual data features are extracted from the models. These features are defined as the outputs of the penultimate layer, with dimensions being 512 for ResNet18 and 2048 for ResNet50. The CCS algorithm employs the aforementioned Forgetting score. Within the CCS approach, we consistently set the hyperparameter \(\beta\) to zero across all data selection ratios.
We also provide the results of CCS with optimal $\beta$ obtained by grid search in Appendix \S\ref{sec:ccs}.

All computational tasks utilized consistent network architectures, as detailed in Section \ref{sec:experiment}: ResNet18 for CIFAR-10 and ResNet-50 for both CIFAR-100 and the ImageNet dataset. Additional experimental specifications related with learning algorithms are reported in Table \ref{tab:detail} of Appendix \S\ref{sec:experiment_details}.

Details of the baselines are summarized below:
\begin{itemize}
    \item EL2N score: The Error L2-Norm (EL2N) score of data $(\rvx_i, y_i)$ is defined as $\E[\|f(\rmW(t),\rvx_i)-y_i\|_2]$, where $f(\rmW(t),\rvx)$ is the output of the neural network for the sample $(\rvx,y)$ at the $t$-th epoch.
    
    \item Forgetting score: The Forgetting score is defined as the number of times during training (until epoch $T$) that the decision of the sample switches from a correct one to an incorrect one. $\text{Forgetting}(\rvx_i, y_i) $ is defined as
    \beq
    \sum_{t=2}^T \mathbbm{1}\{\argmax f(\rmW(t-1), \rvx_i) = y_i\} (1-\mathbbm{1}\{\argmax f(\rmW(t),\rvx_i) = y_i\}).
    \eeq

    \item AdaCore: Adaptive Second order Coresets (AdaCore) is an algorithm that solves the optimization problem, which finds a subset that imitates the full gradient preconditioned with the Hessian matrix: 
    \beq
        S^* \in \text{argmin}_{S \subset V} \sum _{i \in V} \min_{j \in S}  \| \rmH_i(w_t)^{-1} \mathbf{g}_i(w_t) - \rmH_j(w_t)^{-1} \mathbf{g}_j(w_t) \|, \text{ s.t. } |S| \leq r
    \eeq
    where $\mathbf{g}_i(w_t) = \nabla l(w_t, (\mathbf{x}_i, y_i))$ and $\mathbf{H}_i(w_t) = \nabla^2 l(w_t, (\mathbf{x}_i, y_i))$ represent the gradient and Hessian of the loss function for the data point $(\mathbf{x}_i, y_i)$ using the model parameter $w_t$ at the $t$-th epoch of training, respectively. Let $V$ represents the full dataset and $S$ be the coreset of size $r$.
    In the AdaCore method, when employing the cross entropy loss with a softmax layer as the final layer, the gradient $\mathbf{g}_i(w_t)$ is approximated by $p_i - y_i$, where $p_i$ is the softmax output for the data point $(\mathbf{x}_i, y_i)$. Moreover, to reduce computational complexity, the Hessian $\mathbf{H}_i(w_t)$ is approximated using only its diagonal.

    \item LCMat: Loss-Curvature Matching (LCMat) is an algorithm that solves the optimization problem, which finds a subset that matches the loss   curvature of full dataset. Due to the intractability of utilizing the loss curvature, an alternative optimization problem is suggested as follows: 
    \beq
        S^* \in \text{argmin}_{S \subset V} \sum _{i \in V} \min_{j \in S} \| \mathbf{g}_i(w_t) - \mathbf{g}_j(w_t) \| + \dfrac{1}{2} \rho \sum_{k \in \mathcal{K}} |\lambda_{i,k} - \lambda_{j,k}| , \text{ s.t. } |S| \leq r
    \eeq
    where $\mathbf{g}_i(w_t) = \nabla l(w_t, (\mathbf{x}_i, y_i))$ and $\lambda_{i,k} = [\mathbf{H}_i(w_t)]_{kk} = \nabla^2_{kk} l(w, (\mathbf{x}_i, y_i))$ denote the gradient and the $k$-th diagonal element of the Hessian of the loss function for the data point $(\mathbf{x}_i, y_i)$ with the model parameter $w_t$ at the $t$-th epoch of training, respectively. Let $V$ represent the full dataset, $S$ the coreset with size $r$ and $W$ the model parameter space. $\mathcal{K} = \argmax_{|\mathcal{K}| =K} \sum_{j\in\mathcal{K}} \text{Var}_i(\lambda_{i,k})$ is a set of indices for $K$ sub-dimensions on $W$, where the dimension variance is high.
    In LCMat, when employing the cross entropy loss with a softmax layer as the final layer, the gradient $\mathbf{g}_i(w)$ is approximated by $p_i - y_i$, where $p_i$ is the softmax output for the data point $(\mathbf{x}_i, y_i)$. 
    
    \item Moderate DS:
    For each class of a given dataset, Moderate Coreset calculates the distance between features and the feature mean of the class, which is  defined as $d_i = \|\rvf_i-\frac{\sum_{j \in S} \rvf_j}{|S|}\|_2$ where $S$ is the set of features whose label is the same as $f_i$. Then, the data points with distances closest to the distance-median($median(\{d_i\}_{i \in S})$) are selected.
    
    \item CCS:
    Coverage-Centric Coreset Selection (CCS) is an algorithm based on difficulty-based score, which considers overall data coverage upon a distribution as well as important data. CCS prunes $\beta\%$ hardest data first and splits the remained data into $k$ subsets $\{\rmB_i\}_{i=1}^k$ based on evenly divided score ranges $\{\rmR_i\}_{i=1}^k$.  Then, CCS selects the same number of samples from each score range to make the score distribution of the selected samples uniform. 

    \item SSL Prototype score: 
    The work by \citet{beating} conducts $k$-means clustering of samples in the embedding space of a model pre-trained on the ImageNet dataset. It then defines a self-supervised prototype metric (SSL Prototype score) as the  Euclidean distance to its nearest cluster centroid, or prototype. Points located closer to the center have lower SSL scores.  

    \item Memorization score:
    Memorization score \citep{memorization} calculates the influence of each training example $(\rvx_i, y)$ on the classification accuracy of that same example $(\rvx_i, y)$, and is defined as follow
    \beq
     \text{mem}((\rvx_i, y_i)) = \mathbf{P}(h_{T}(\rvx_i)=y_i)-\mathbf{P}(h_{T\setminus{\{(\rvx_i,y_i)\}}}(\rvx_i)=y_i)
     \eeq
     where $h_S(\cdot)$ is a model trained on the set $S$ and $T$ is the training dataset.

\end{itemize}

\subsection{Experiment details}
\paragraph{Data pruning experiment} \label{sec:experiment_details}
We conduct experiments with three public datasets, CIFAR-10/100 and ImageNet by training ResNet networks \citep{resnet} of different depths. ResNet18 is used for CIFAR-10 and ResNet50 is used for CIFAR-100 and ImageNet dataset. The implementation  is based on the ResNet network in torchvision \citep{torch}. 
Since CIFAR-10/100 images are smaller than ImageNet images, we replace the front parts of the ResNet (convolution layer with $7 \times 7$ kernel and $2 \times 2$ stride, max pooling layer with $3 \times 3$ kernel and $2 \times 2$ stride) with a single convolution layer with $3 \times 3$ kernel and $1 \times 1$ stride for the small size images. The details on hyperparameters and optimization methods used in training are summarized in Table \ref{tab:detail}. 

Our experiments report the averaged results from three runs on CIFAR-10/100 and two on ImageNet, with shaded regions representing standard deviations. Networks are trained on datasets curated based on specific selection ratios and methods. 
Crucially, our data selection ensures equal selection from each class by preserving the portion data in each class.
\begin{table}[tb!]
\centering
\caption{Details for the experiments used in the training of the dataset.}\label{tab:detail} 
\begin{tabular}{l|lll}
\toprule
& CIFAR-10 & CIFAR-100 & ImageNet
\\
\midrule
Architecture & ResNet18 \hspace{1.3cm} & ResNet50 & ResNet50 \\
Batch size & 128 & 128 & 256 \\
Epochs & 200 & 400 & 90 \\
Initial Learning Rate & 0.05 & 0.2 & 0.1 \\
Weight decay & 5e-4 & 5e-4 & 1e-4 \\
Learning Rate Scheduler & \multicolumn{2}{l}{Cosine annealing scheduler} & Step scheduler \\
Optimizer & \multicolumn{3}{l}{SGD with momentum 0.9} \\

\midrule    

\multirow{3}{*}{Data Augmentation} & \multicolumn{2}{c}{ Random Zero Padded Cropping (4 pixels)} & Random Resized Cropping  \\
& \multicolumn{3}{l}{Random left-right flipping (probability 0.5)} \\ 
& \multicolumn{3}{l}{Normalize by dataset's mean, variance} \\ 
\bottomrule
\end{tabular}
\end{table}

\paragraph{Cross-architecture robustness} \label{sec:cross_details}
We conduct cross-architecture experiments on the CIFAR-10 dataset, training three distinct networks: a simple CNN, EfficientNet-B0, and a Vision Transformer (ViT) pretrained on the ImageNet dataset.

For the simple CNN, we design an architecture comprising three convolutional layers with a $3 \times 3$ kernel and $1 \times 1$ stride (channels: 64, 128, 256). This is paired with two max-pooling layers with a $2 \times 2$ kernel. The convolutional layers are interspersed with these max-pooling layers. Following the convolutional layers, the network is connected to two fully connected layers (channels: 128, 256). Each convolutional layer is equipped with a batch normalization layer followed by a non-linear ReLU activation layer. We set the initial learning rate to 0.05 and weight decay to 1e-4. Other details are the same as CIFAR-10 case in Table \ref{tab:detail}.
For EfficientNet-B0, we closely follow the implementation details of \citet{efficientnet}, by setting the learning rate to 1e-4 and a weight decay of the same magnitude. Other implementation specifications are the same with the details in Table \ref{tab:detail} of CIFAR-10.
For the ViT, we adhere to the implementation specifications as detailed in \citet{ViT}. We obtain a ViT model pretrained on the ImageNet dataset using the timm module in PyTorch, which we subsequently fine-tune on the CIFAR-10 dataset for 10 epochs. For fine-tuning a model pre-trained on ImageNet to adapt to the CIFAR-10 dataset, we resize the data to fit the 224x224 pixel dimensions.
We set the initial learning rate to 1e-4 and weight decay to 1e-4. We do not use a learning rate scheduler. Other details are the same as CIFAR-10 case in Table \ref{tab:detail}.

Within our algorithm, BWS, we utilize a Forgetting score sourced from the ResNet18 architecture. Furthermore, we establish a feature extractor using either simple CNN, EfficientNet-B0, or ViT architecture, repectively. For the CNN and EfficientNet-B0 architecture, we execute training for 20 epochs, while for the ViT setup, we fine-tune for 3 epochs.
We report the averaged results from three independent runs on the three networks, with the shaded regions indicating the standard deviations. Similar to previous experiments, data samples are selected to ensure a balanced portion of each class, preserving the original class ratios within the CIFAR-10 dataset.

\paragraph{Robustness to label noise}
We generate  a noise version of the CIFAR-10 dataset with symmetric label noise at levels of 20\% and 40\%, respectively. To evaluate this noisy dataset, we compute the EL2N score using a ResNet18 model, averaging the outcomes over five independent runs. The EL2N score is selected due to its lower computational cost compared to the Forgetting score, especially when it is required to re-calculate the new EL2N score for the entire noisy dataset.
In our analysis, we apply Algorithm \ref{alg1} to the CIFAR-10 dataset, where the samples are ordered by their EL2N scores. To calculate the classification accuracy of $\rvw_{\rmS}$, we specifically use the lower-scoring 50\% of the samples, represented as $\frac{1}{n}\sum_{i=\frac{n}{2}}^{n} \mathbbm{1}(\argmax_c(\rvw_{\rmS}^\top\rvx_i)_c=y_i)$, instead of the typical approach $\frac{1}{n}\sum_{i=1}^n \mathbbm{1}(\argmax_c(\rvw_{\rmS}^\top\rvx_i)_c=y_i)$. This adjustment was made to exclude noisy samples from the quality evaluation of window subsets and thus prevent overfitting to noise in the data.

\paragraph{Ablation on different window selection methods}
The formal definitions of Gradient difference and Gradient similarity are as follows:
\begin{itemize}
    \item Gradient difference: minimizing the difference between the gradients of the full training dataset ($V$) and window subset ($S$). 
    \beq
        \text{Gradient Difference}(V, S) = \left\| \frac{\sum_{i \in V} \nabla f_{\rvw}(\rvx_i)}{|V|} -\frac{\sum_{i \in S} \nabla f_{\rvw}(\rvx_i)}{|S|}\right\|_2
    \eeq
    \item Gradient similarity: maximizing the cosine similarity between the gradients of the full training dataset ($V$) and window subset ($S$).
    \beq
        \text{Gradient Similarity}(V, S) = 
        \frac{\sum_{i \in V} \nabla f_{\rvw}(\rvx_i)}{\left\|\sum_{i \in V} \nabla f_{\rvw}(\rvx_i)\right\|_2} \cdot \frac{\sum_{i \in S} \nabla f_{\rvw}(\rvx_i)}{\left\|\sum_{i \in S} \nabla f_{\rvw}(\rvx_i)\right\|_2} 
    \eeq
\end{itemize}

\subsection{Computational cost}\label{sec:cost}
\begin{table}[tbh]
\centering
\caption{Time cost (in seconds) for subset selection of each algorithm across selection ratios.}
\label{tab:computational_cost}
\resizebox{\textwidth}{!}{
\begin{tabular}{{cc|ccccccccc}}
\toprule
\multicolumn{2}{c|}{Selection ratio}& 1\% & 5\% & 10\% & 20\% & 30\% & 40\% & 50\% & 75\% & 90\% \\
\midrule
\multirow{3}{*}{CIFAR-10} & BWS (Ours) & 4.3 & 4.8 & 7.2 & 8.5 & 9.7 & 10.4 & 10.4 & 9.0 & 6.3 \\
& LCMat & 520 & 1197 & 2260 & 4213 & 5800 & 7173 & 8450 & 10320 & 10631\\
& AdaCore & 224 & 699 & 1256 & 2273 & 3186 & 3977 & 4649 & 5637 & 5839 \\
\midrule
\multirow{3}{*}{CIFAR-100} & BWS (Ours) & 14.6 & 55.6 & 57.3 & 62.2 & 64.7 & 64.0 & 61.7 & 45.7 & 30.2 \\
& LCMat & 1465 & 1468 & 1471 & 1478 & 1483 & 1489 & 1493 & 1501 & 1504 \\
& AdaCore & 1295 & 1300 & 1304 & 1309 & 1315 & 1320 & 1324 & 1331 & 1334 \\
\midrule
\multirow{3}{*}{ImageNet} & BWS (Ours) & 1423 & 2910 & 3941 & 6141 & 7590 & 8616 & 9029 & 10118 & 6499 \\
& LCMat & 238451 & 239027 & 239694 & 240934 & 242003 & 242864 & 243594 & 244854 & 245245 \\
& AdaCore & 213733 & 214309 & 214963 & 216181 & 217067 & 217866 & 218521 & 219593 & 219924 \\
\midrule
\multicolumn{2}{c|}{GPU} & \multicolumn{9}{c}{Nvidia A100 40GB} \\
\bottomrule
\end{tabular}}
\end{table}
We compare the computational cost of our BWS algorithm, as detailed in Algorithm \ref{alg1}, with other optimization-based coreset selection baselines, namely LCMat and AdaCore. We assume that the sample scores, used for sorting, is readily available and that the feature extractor is also pre-provided for both ours and optimization-based methods. We report  and compare the time taken to select the subset for each algorithm.

In Table \ref{tab:computational_cost}, we detail the time required to select subsets from various datasets using the different methods. Clearly, our method outperforms optimization-based techniques in terms of time cost for subset selection. As we have previously described in Sec. \ref{sec:methodology}, our strategy, which selects the best window subset from a continuous interval of samples sorted by their scores, greatly reduces the search space compared to the general optimization techniques, leading to improved efficiency.

\section{Detailed Review of Related Works}\label{sec:additional_related}
In this section, we provide additional related works on data subset selection regarding various perspectives.

When there is no validation set, some score-based selection methods, such as EL2N \citep{EL2N} and Forgetting \citep{forgetting}, suffer from performance degradation when the dataset includes label-noise samples, since these methods often assign high scores to label-noisy samples, as label-noise samples are inherently hard to learn. Some recent methods adopt more cautious measures to distinguish hard-to-learn but clean-label samples, known to be valuable to enhance the generalization ability of neural networks, from label-noise samples.
For instance, Cartography \citep{cartography} utilizes two measures, confidence mean and confidence variance of data sample, to distinguish hard-to-learn samples from mere label-noise samples. Second-Split Forgetting \citep{secondsplit} achieves this goal by observing the learning time and forgetting time of each sample while training a model.
AUM \citep{AUM} observes the logit value of a given label and the next largest logit value, and uses the gap to separate noisy data samples and ambiguous data samples.

Another important issue that has been recently explored in data subset selection is the computational overhead in quantifying the data value. 
There are several recent attempts to valuate data without training of a neural network.
For example, CG-score \citep{CG-score} evaluates data instances without model training by calculating the analytical gap in generalization errors when an instance is held out. 
LAVA \citep{lava} evaluates the value of each data instance without a model training by using a proxy function, the class-wise Wasserstein distance between training and validation set, for the validation performance. DAVINZ \citep{DAVINZ} utilizes the Neural Tangent Kernel (NTK) of a network at initialization for calculating the contribution of each data instance to domain-aware generalization error bound.

In optimization-based selection, many works have utilized the effect of data on the model training. MaxMargin \citep{MaxMargin}, IWeS \citep{leaveraging}, and Selection-Via-Proxy \citep{selectionviaproxy} use the confidence of a model to identify uncertain data during optimization. Maximum Margin Coresets \citep{MaxMargin} selects data with the smallest margin in an SVM setting, IWeS \citep{leaveraging} selects examples using importance sampling with a sampling probability based on the confidences of two models, and Selection-Via-Proxy \citep{selectionviaproxy} applies confidence-based methods to a small proxy model to perform data selection. \citet{convexprogram} solve a convex linear programming problem to find high-value data that contributes much to the loss and optimization, and Glister \citep{glister} finds data that contributes significantly to the loss during the training of a neural network. GradMatch \citep{grad-match} finds a subset whose gradient matches better with the gradient of the full dataset.

Additionally, in active learning, where data samples are selectively labeled for semi-supervised learning, Neural-Preconditioning \citep{preconditioning} obtains the label of data that dominates the eigenspace of the NTK, and  \citet{reducing_labeling} obtain the label of the least confident data to reduce labeling costs.
The works by \citet{MaxMargin, leaveraging, selectionviaproxy} utilize the confidence of a model to identify uncertain data during the optimization, and the works by \citet{convexprogram, glister} find the data that contributes much to the loss.

\section{Comparison with CCS}\label{sec:ccs}

CCS \citep{CCS} prunes the hardest $\beta$\% samples, divides the remaining data into non-overlapping $k$ ranges based on the difficulty scores, and uniformly assigns budgets to each range. Samples are then chosen from each range within the budget. If a range has fewer data than the assigned budget, the remaining budget is iteratively reassigned to other ranges.
While methodologies based on difficulty scores suffer from a drastic performance drop at low subset ratios, CCS achieves high performance across a broad range by selecting diverse data with appropriate $\beta$ and $k$. However, CCS does not propose an efficient method to find the desired $\beta$ and reports the result with the optimal $\beta$ obtained by grid search, which may require high computational cost.
Since choosing the best performing model among the models trained on each subset chosen by different $\beta$ is not fair for comparison to other baselines, including BWS, we reported the result of CCS obtained by setting the hyperparameter $\beta$ as 0 in the main experimental results.
In this section, we additionally report the results of CCS with the optimal $\beta$ found by grid search  and compare the performance with those of BWS and oracle window in Table \ref{tab:CCS_optimal}. In Table \ref{tab:CCS_beta}, we also report the optimal $\beta$ for CCS across different selection ratios. 
For the subset ratios of 10\%, 20\%, 30\%, and 50\%, we utilize the $\beta$ values reported in the original paper (marked with $^\dagger$), and for other subset ratio, we conduct a grid search to find the best $\beta$ by exploring it with a step size of 10\%.
 
 Several key observations emerge from the results. First, CCS after hyperparameter tuning, achieves performance comparable to the Oracle Window at lower selection ratios (1-10\%). This is a natural consequence, given both methods' ability to discard the top hardest samples in favor of easier ones at low ratios. However, at higher ratios (20\%-90\%), CCS's efficacy decreases relative to both Oracle Window and BWS, even after tuning $\beta$. This decline can be attributed to CCS's strategy of selecting samples across a uniform score distribution after pruning the hardest ones. Even CCS adjusts $\beta$ to 0 or lower values (e.g., 10 or 20) for ratios beyond 30\%, the sample selection with uniform score distribution makes CCS incorporate not only hard samples but also easy samples, which are less effective in high subset ratios. In comparison, Optimal Window or BWS, which select samples from a contiguous difficulty range, focus on selecting harder samples as the subset ratio increases, resulting in better performance.
Moreover, we analyze the computational costs associated with CCS and BWS as summarized in Table \ref{tab:CCS_time_cost}. Since CCS requires the repeated training of deep neural networks in the process of tuning the hyperparameter $\beta$, it requires significant computational overhead compared to BWS. On the other hand, BWS circumvents the need for hyperparameter tuning, by solving a simple proxy task to identify the best window subset, which considerably shortens the time requirement.

\begin{table}[ht]
\centering
\caption{Time cost (in seconds) to compute CCS and BWS.}
\label{tab:CCS_time_cost}
\begin{tabular}{{cc|ccc}}
\toprule
\multicolumn{2}{c|}{Selection ratio}& 5\% & 30\% & 75\%  \\
\midrule
\multirow{2}{*}{CIFAR-10} & CCS & 812 & 3897 & 3654  \\
& BWS & 13 & 59 & 130  \\
\midrule
\multirow{2}{*}{CIFAR-100} & CCS & 3909 & 18767 & 17594 \\
& BWS & 75 & 182 & 339 \\
\bottomrule
\end{tabular}
\end{table}

\begin{table}[ht]
\centering
\caption{Optimal $\beta$ (\%) at different selection ratios in various dataset.}
\label{tab:CCS_beta}
\begin{tabular}{c|ccccccccc}
\toprule
Selection ratio & 1\% & 5\% & 10\% & 20\% & 30\% & 40\% & 50\% & 75\% & 90\% \\
\midrule
CIFAR-10 & 80 & 50 & 30$^\dagger$ & 10$^\dagger$ & 10$^\dagger$ & 10 & 0$^\dagger$ & 0 & 0 \\
CIFAR-100 & 99 & 80 & 50$^\dagger$ & 40$^\dagger$ & 20$^\dagger$ & 20 & 20$^\dagger$ & 20 & 0 \\
ImageNet & 80 & 30 & 30$^\dagger$ & 20$^\dagger$ & 20$^\dagger$ & 10 & 10$^\dagger$ & 10 & 0 \\
\bottomrule
\end{tabular}
\end{table}

\begin{table}[hbt]
\centering
\caption{Test accuracy of CCS with optimal $\beta$ at different selection ratios.}
\label{tab:CCS_optimal}
\begin{tabular}{{cc|ccccccccc}}
\toprule
\multicolumn{2}{c|}{Selection ratio}& 1\% & 5\% & 10\% & 20\% & 30\% & 40\% & 50\% & 75\% & 90\% \\
\midrule
\multirow{3}{*}{CIFAR-10} & CCS & 46.58 & 72.12 & 81.99 & 88.89 & 91.74 & 93.10 & 94.09 & 94.95 & 95.29 \\
& Oracle & 47.17 & 72.89 & 82.67 & 89.06 & 91.80 & 93.59 & 94.54 & 95.23 & 95.37 \\
& BWS & 46.10 & 70.70 & 82.29 & 88.74 & 91.80 & 93.59 & 94.54 & 95.23 & 95.37 \\
\midrule
\multirow{3}{*}{CIFAR-100} & CCS & 11.01 & 31.11 & 45.52 & 56.09 & 64.26 & 68.51 & 70.80 & 75.83 & 78.13 \\
& Oracle & 10.63 & 30.39 & 45.16 & 58.91 & 67.51 & 72.70 & 75.00 & 78.42 & 79.00 \\
& BWS & 8.43 & 29.25 & 44.11 & 58.30 & 67.20 & 72.17 & 73.83 & 77.78 & 79.00 \\
\midrule
\multirow{3}{*}{ImageNet} & CCS & 8.13 & 31.40 & 45.10 & 57.12 & 62.65 & 67.54 & 69.58 & 73.10 & 74.59 \\
& Oracle & 7.97 & 33.58 & 48.84 & 62.83 & 68.22 & 71.33 & 72.74 & 74.73 & 75.25 \\
& BWS & 7.02 & 33.31 & 46.72 & 62.32 & 67.16 & 70.47 & 72.68 & 74.73 & 75.25 \\
\bottomrule
\end{tabular}
\end{table}

\section{Additional Experiments}
\begin{figure}[t]
\centering
    {\includegraphics[width=0.33\linewidth]{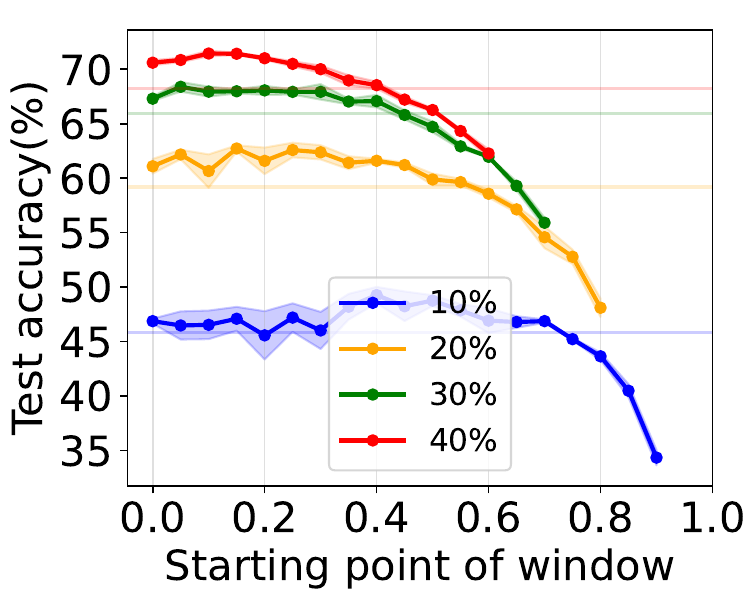}}
    \caption{
    Sliding window experiments in ImageNet dataset to measure the test accuracy of the models trained by window subsets while changing the starting point of the windows. 
    Samples are sorted in descending order by their difficulty scores. The horizontal lines are results from random selection. For each subset ratio, there exists the best window, and its starting point shifts toward left as the subset ratio increases.
    }
    \label{fig:window_exp_app}
\end{figure}

\subsection{Sliding window experiment for ImageNet dataset}\label{sec:sliding_window_app}

We investigate the efficacy of the window selection approach by varying the starting points and demonstrate the existence of an optimal window subset. In this process, we arrange the ImageNet samples in descending order based on their Forgetting scores \citep{forgetting}, and then select windows of varying sizes, from $10\%$ to $40\%$. The starting point for these windows is adjusted from $0$ to $(100-w)\%$, incrementing in steps of $5\%$. Subsequently, we train a ResNet50 model using these window subsets and present the resulting test accuracies in Fig.~\ref{fig:window_exp_app}.
Consistent with the observations in Fig.~\ref{fig:window_exp}, within the ImageNet dataset, we note that for each subset ratio, there is an optimal starting point. Notably, this optimal point shifts progressively towards lower values, which correspond to more difficult samples, as the size of the window subset increases.

\subsection{Cross architecture robustness}\label{sec:cross_arch_exp_app}

\begin{figure*}[tb]
\centering
    \subfigure[CNN \label{fig:pruning_CNN}]{\includegraphics[width=0.28\linewidth]{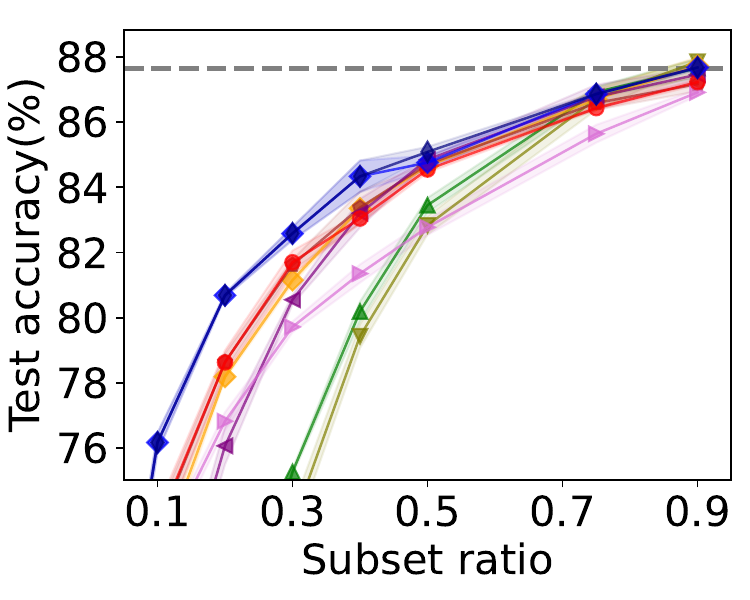}}
    \subfigure[EfficientNet-B0 \hspace{0.9cm}\label{fig:eff_exp}]{\includegraphics[width=0.42\linewidth]{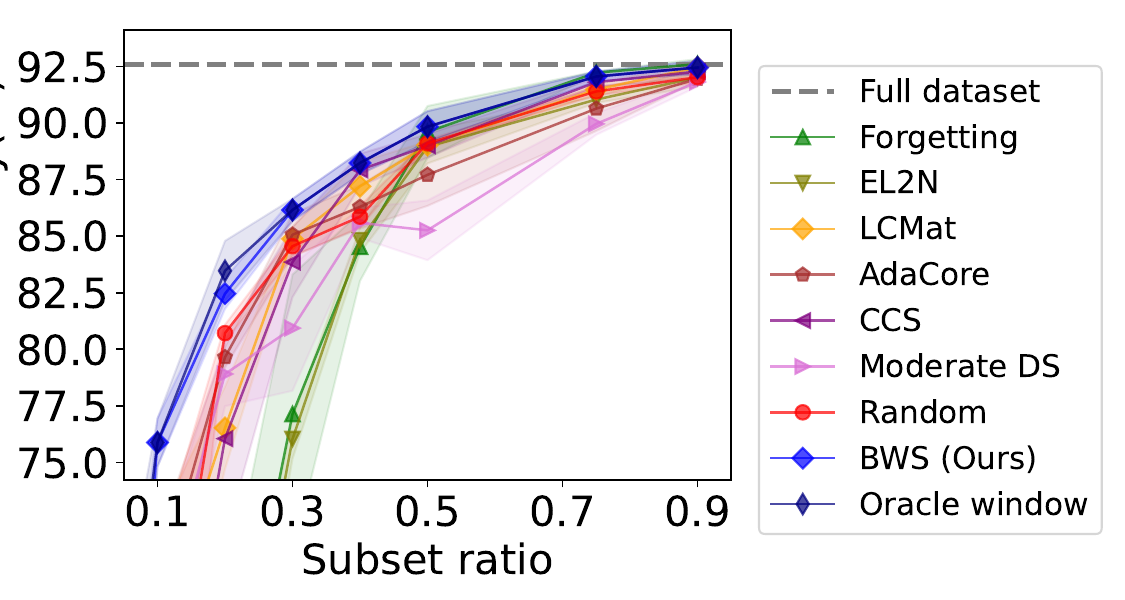}}
    \caption{Cross-architecture experiments with CNN (left), and EfficientNet-B0 (right), where samples scores are calculated using ResNet18 model. Full results are reported in Table \ref{tab:cnn_acc}--\ref{tab:eff_acc}.
    }
    \label{fig:cross_arch_exp_app}
\end{figure*}

To test the robustness of our method across changes in neural network architectures, we conduct data pruning experiments on CIFAR-10 while using different architectures during sample scoring and training. The window subsets are constructed using samples ordered by their Forgetting scores, calculated on ResNet18 architecture. Then, the best window selection (Alg. \ref{alg1}) and the model training are conducted using a simpler CNN architecture or EfficientNet-B0 architecture. The results on the CNN architecture are presented in Fig. \ref{fig:pruning_CNN}, and those on the EfficientNet-B0 are shown in Fig. \ref{fig:eff_exp}. In all cases, our method (BWS) consistently achieves competitive performances across all selection ratios, demonstrating its robustness to changes in neural network architectures during data subset selection.

\subsection{Robustness to label noise}\label{sec:label_noise}

\begin{figure*}[!tb]
\centering
    \subfigure[20\% label-noise CIFAR-10\label{fig:noise}]{\includegraphics[width=0.28\linewidth]{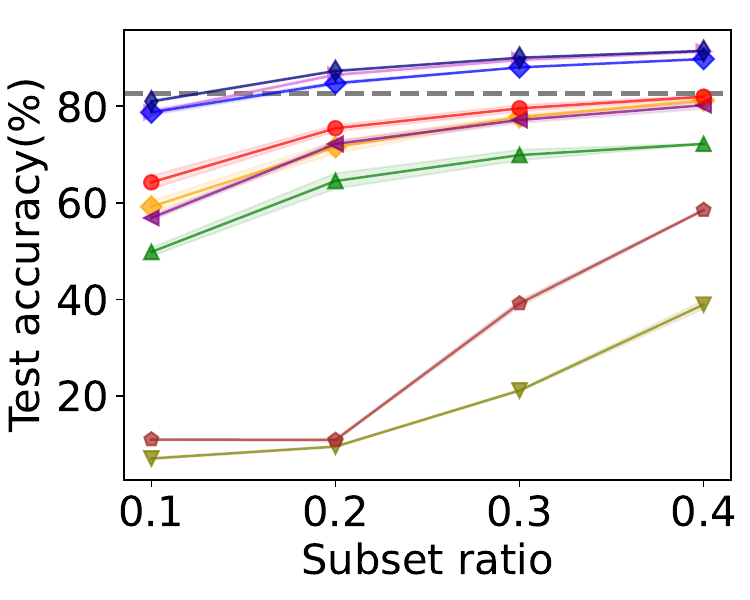}}
    \subfigure[40\% label-noise CIFAR-10 \label{fig:noise40}]{\includegraphics[width=0.42\linewidth]{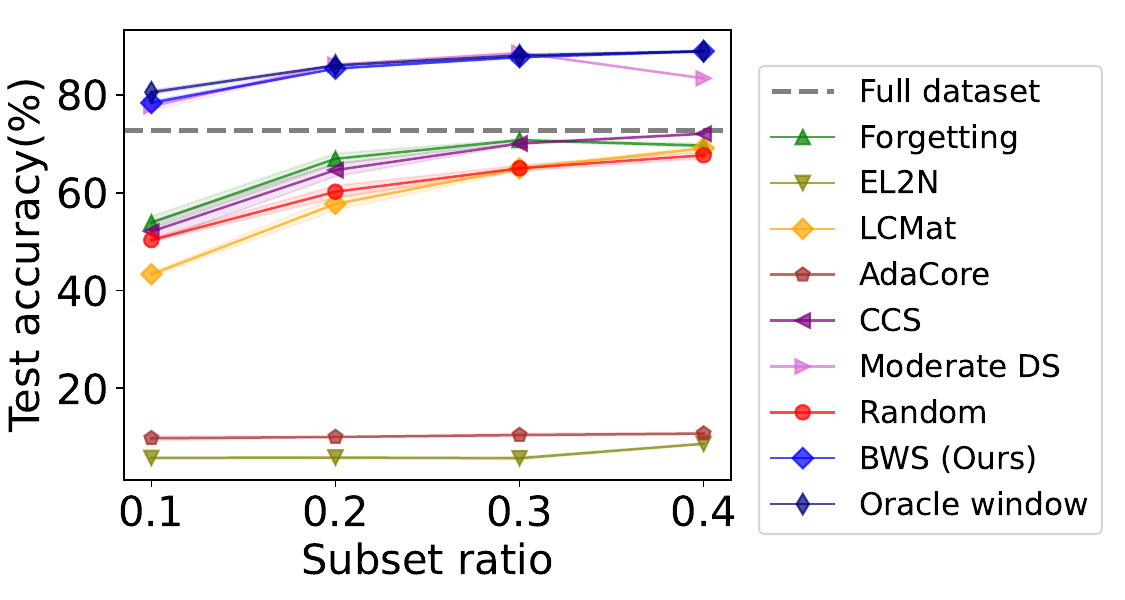}}
    \caption{Data pruning experiments with CIFAR-10, including (a) 20\% label-noise, (b) 40\% label-noise. 
    Our method (BWS) attains the accuracy of the full training dataset despite the presence of label noise.
    Full results are reported in Table \ref{tab:noise_acc}-\ref{tab:noise40_acc}.
    }\label{fig:noise_exp}
\end{figure*}
We test the robustness of BWS in the presence of label noise in the training dataset. We corrupt randomly chosen 20\% and 40\% samples of CIFAR-10 by random label noise. It has been previously reported that the difficulty score-based selection methods are susceptible to label noise since such methods tend to assign high scores to label-noise samples \citep{forgetting, EL2N}. Thus, these methods often ends up prioritizing the label-noise samples in the selection process, leading to suboptimal results. On the other hand, our algorithm offers flexibility in choosing window subsets with varying levels of difficulty by changing the starting point, and adopts an approach to select the best window by solving a proxy task using the kernel ridge regression. To further enhance the robustness of our method, we can modify Alg. \ref{alg1} to evaluate the solution of kernel ridge regression using only the low-scoring 50\% samples from the training dataset, which will rarely
include label-noise samples, instead of the full dataset. We use EL2N \citep{EL2N} as the difficulty score to align the samples in our algorithm. 
In Fig. \ref{fig:noise_exp}, we compare the performance of this modified version of BWS with other baselines. While difficulty score-based selection and optimization-based selection methods suffer from performance degradation due to label noise, our method, along with another label noise-robust method, Moderate DS, achieves performance even higher than what is achievable with the full training dataset, which includes the 20\% or 40\% label noise, respectively.

\section{Ablation studies}

\subsection{Ablation study on window type: two half-width sliding windows}\label{sec:two_window}

\begin{figure}[t]
\centering
    \subfigure{\includegraphics[width=0.3\linewidth]{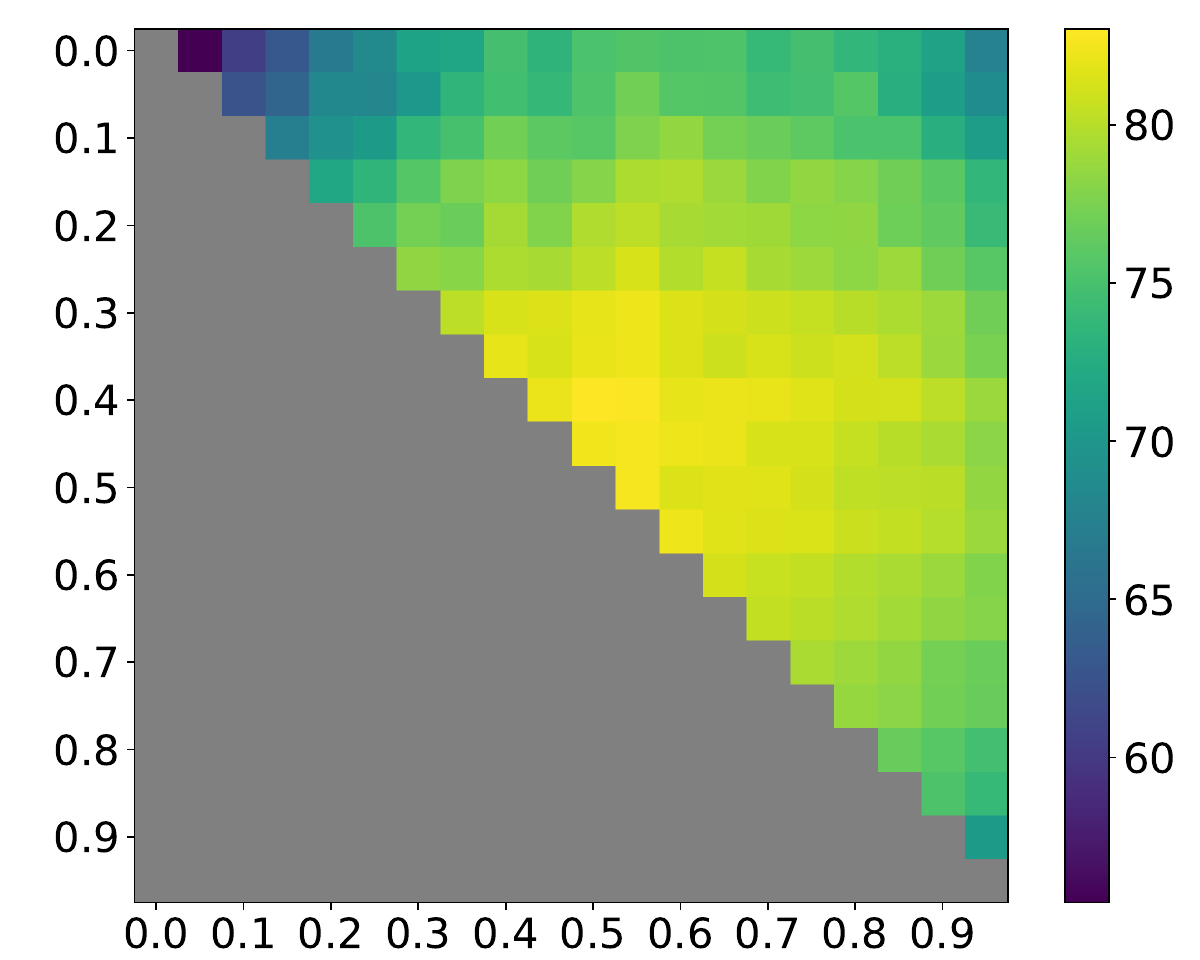}}
    \subfigure{\includegraphics[width=0.3\linewidth]{figs/two_window_exp/0.2_grid_acc.pdf}} \\
    \subfigure{\includegraphics[width=0.3\linewidth]{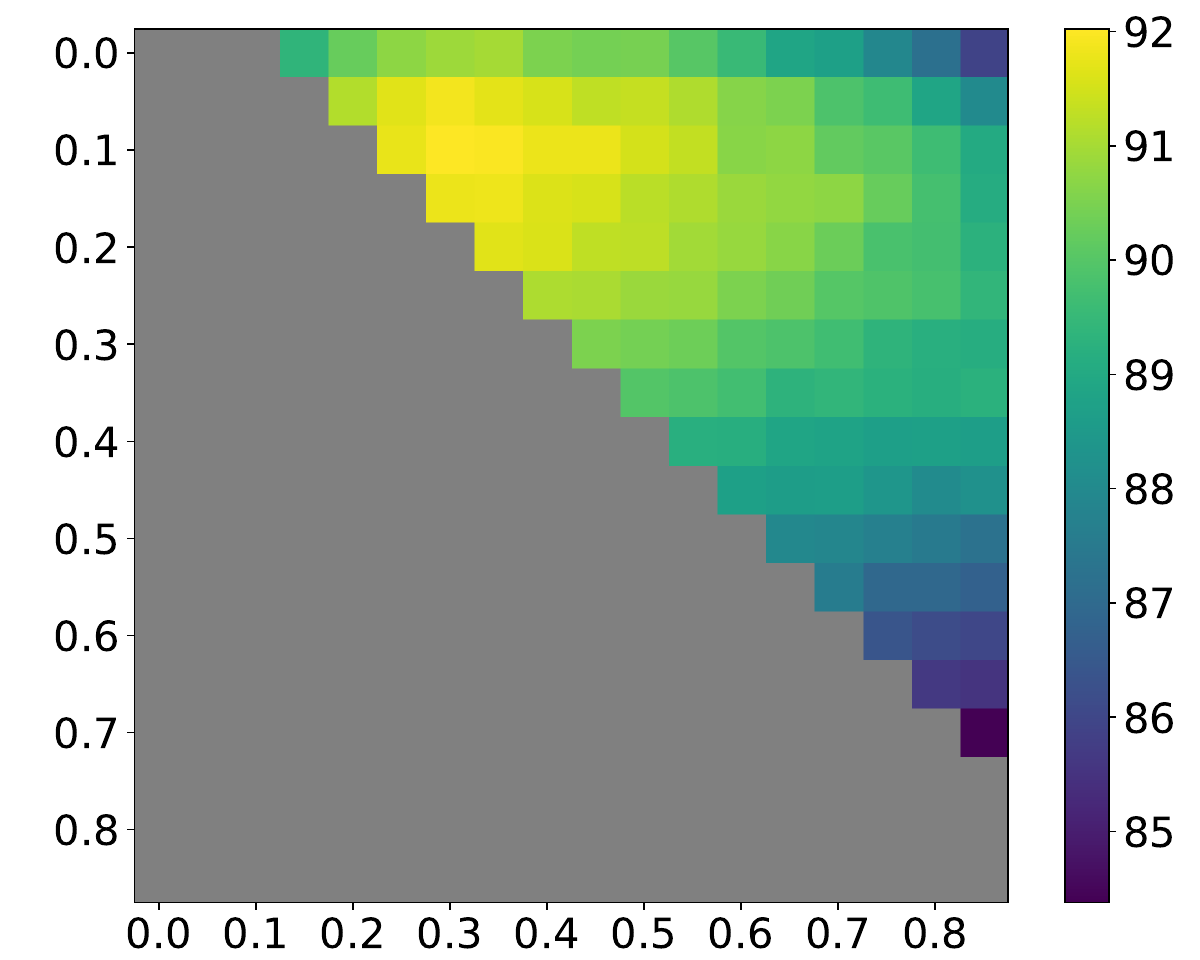}}
    \subfigure{\includegraphics[width=0.3\linewidth]{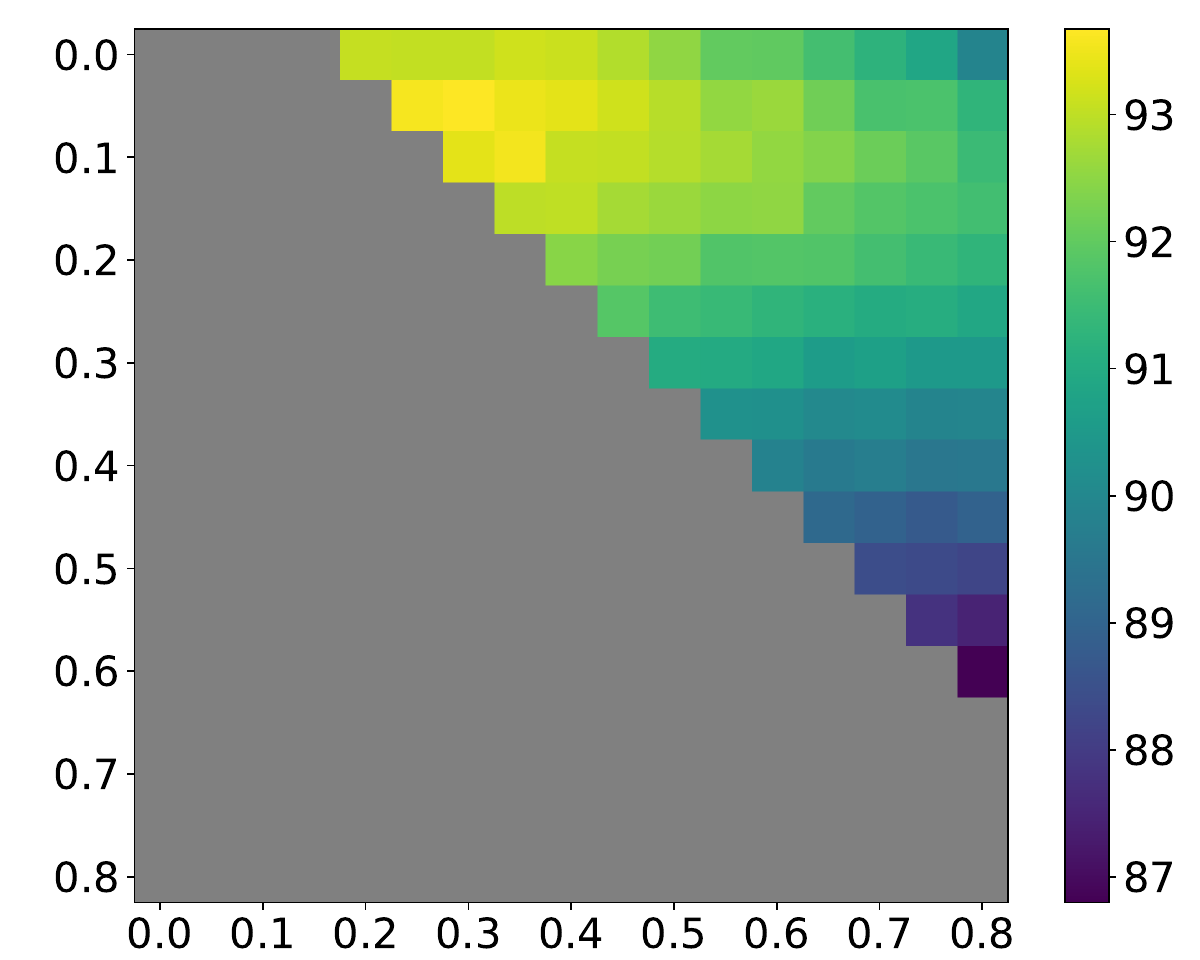}}
    \caption{
    Test accuracy of the models trained with two half-width windows of subset ratios 10\% (top left), 20\% (top right), 30\% (bottom left), and 40 \%(bottom right) with varying starting points. 
    The numbers in axes indicate the starting points of each interval, and the color indicates the test accuracy for each composition of half-width windows. We note that a contiguous window, a point near the diagonal, attains performance levels comparable to the best performance.
    }
    \label{fig:two_window_exp}
\end{figure}

BWS sorts samples in a dataset by their difficulty scores and then selects the optimal window subset from one continuous single-interval regime. Thus, the window selection chooses the samples of similar difficulty level. 
To further examine possible benefits from non-contiguous subset selection, we conduct an additional experiment on the CIFAR-10 dataset by finding the optimal two half-width windows while varying their starting points. In detail, we sort the samples from CIFAR-10 in descending order based on Forgetting score \citep{forgetting} and for a subset of size $w\%$, we search over all combinations of two half-width windows, denoted by $[x_1, x_1+w/2] \cup [x_2, x_2+w/2]$ while varying their starting points $(x_1, x_2)$ in $x_1\in[0, 100-w]$ and $x_2\in [x_1+w/2, 100-w/2]$ with a step size of 5\%. We train ResNet18 on each subset and evaluate the corresponding test accuracies. The full results are presented in Fig. \ref{fig:two_window_exp}, and in Table \ref{tab:two_window} we report the top five results (the compositions of half-width windows and their test accuracies) for subset ratios ranging from 10 to 40\%. We highlight the cases where the two half-width windows are contiguous to each other with bold letters. 

\begin{table}[!tbh]
\caption{
Top-five test accuracies and their corresponding half-width window compositions on CIFAR-10 dataset. We highlight the cases where the two half-width windows are contiguous to each other with bold letters. 
}
\label{tab:two_window}
\vspace{0.2em} 
  \centering
  \resizebox{\textwidth}{!}{  
\begin{tabular}{c|c|ccccc}
\toprule
\multicolumn{1}{c|}{Ratio} & Ranking & 1st & 2nd & 3rd & 4th & 5th \\
\midrule
\multirow{2}{*}{10\%} & Half-width windows & 40-45\%, 50-55\% & 40-45\%, 55-60\% & 45-50\%, 55-60\% & \textbf{50-55\%, 55-60\%} & \textbf{45-50\%, 50-55\%} \\ 
 & Test Acc  & 83.04 & 82.87 & 82.71 & \textbf{82.67} & \textbf{82.46} \\ 
 \midrule
\multirow{2}{*}{20\%} & Half-width windows & 20-30\%, 35-45\% & \textbf{25-35\%, 35-45\%} & 20-30\%, 40-50\% & 20-30\%, 50-60\% & 25-35\%, 45-55\% \\ 
 & Test Acc  & 89.16 & \textbf{89.06} & 88.98 & 88.84 & 88.77 \\ 
 \midrule
\multirow{2}{*}{30\%} & Half-width windows & 10-25\%, 30-45\% & 10-25\%, 35-50\% & 5-20\%, 30-45\% & 15-30\%, 35-50\% & \textbf{15-30\%, 30-45\%} \\ 
 & Test Acc  & 92.02 & 91.98 & 91.90 & 91.84 & 91.80 \\ 
 \midrule
\multirow{2}{*}{40\%} & Half-width windows & 5-25\%, 30-50\% & \textbf{5-25\%, 25-45\%} & 10-30\%, 35-55\% & 5-25\%, 35-55\% & 5-25\%, 40-60\% \\ 
 & Test Acc  & 93.67 & \textbf{93.59} & 93.54 & 93.46 & 93.40 \\ 

\bottomrule
\end{tabular}}
\end{table}

We can observe that for every considered subset ratio, the top-five best performing cases include contiguous windows or windows near to each other with the gap of only 5\%, even though we allowed flexibility in choosing the two half-width windows far away from each other. This result further supports our use of window selection, which only considers subsets from a continuous interval of samples based on difficulty scores, in choosing near-optimal subset in an efficient manner across a broad range of selection ratios. 
\begin{figure*}[t]
\centering
    \subfigure[One and a half times wider window]{\includegraphics[width=0.33\linewidth]{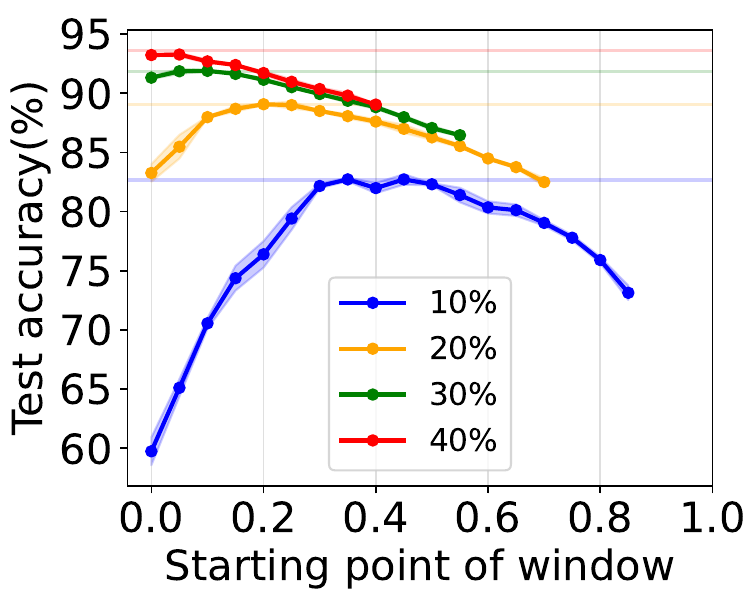}}
    \subfigure[Twice wider window]{\includegraphics[width=0.33\linewidth]{figs/wider_window_exp/pruning_wider_window_2.pdf}}
    \caption{
     Test accuracy using subsets randomly sampled from windows of (a) one and a half times ($\times 1.5$) and (b) twice ($\times 2$) larger than the subset ratio, while varying the starting points of these windows. The horizontal lines represent the results from the oracle window, which is the maximum test accuracy obtained in sliding window experiments, for each subset ratio. Our observations indicate that at lower subset ratios, there are subsets whose performance is comparable to that of the oracle window, but, at higher subset ratios, the performance of all subsets consistently falls short of the oracle window.
    }
    \label{fig:wider_window_exp}
\end{figure*}

\subsection{Ablation study on window type: wider sliding windows}\label{sec:wider_window}

We also conduct an additional experiment on the CIFAR-10 dataset to explore non-contiguous sample selection by considering random  selection from wider windows. 
By arranging the samples in descending order according to difficulty scores and selecting a starting point, denoted as $s\%$, for a given subset ratio of $w\%$, we randomly choose samples within the range $[s, s+c \cdot w]\%$ , where $c$ is a constant greater than 1.
In particular, we sort the CIFAR-10 samples in descending order based on their Forgetting scores \cite{forgetting}, and then select windows of various sizes, ranging from 10\% to 40\%, by adjusting the starting point from 0 to $(100-cw)\%$ in 5\% increments. The window width is $cw$ for a given ratio $w$\% and a constant $c$ equal to either 1.5 or 2. 
Subsequently, we randomly select $w$\% of the data from the window, train the ResNet18 network with this subset, and plot the resulting test accuracies in Fig. \ref{fig:wider_window_exp}.

We observe that training with a wider window, regardless of the constant $c$, results in test accuracy curves similar to those shown in Fig. \ref{fig:window_CIFAR10}. However, the sliding window experiment for wider windows in Fig. \ref{fig:wider_window_exp} shows that the best contiguous window (horizontal lines) achieves better performance than wider windows, especially in high ratios. Thus, this result supports the use of a contiguous window subset in sample selection across a broad range of selection ratios.

\subsection{Ablation study on difficulty scores}\label{sec:ablation_score}

\begin{table}[!thb]
\centering
\caption{
Test accuracy of the \ours algorithm at different data selection ratios, depending on the difficulty score. Due to the high correlation between the difficulty scores, there is a similar sorting order across the scores. Thus, similar window positions are selected by BWS, regardless of the specific difficulty score in use. This similarity in subset selection leads to consistently strong performance regardless of the chosen difficulty score.}
\label{tab:diff_score}
\resizebox{\textwidth}{!}{ 
\begin{tabular}{c|c|ccccccccc}
\toprule
Selection methods & Selection ratio & 1\% & 5\% & 10\% & 20\% & 30\% & 40\% & 50\% & 75\% & 90\% \\
\midrule
\multirow{2}{*}{BWS with C-score} & Test accuracy & 46.25 & 71.34 & 82.02 & 89.12 & 91.85 & 93.62 & 94.62 & 95.18 & 95.28 \\
& Window index & 85\% & 55\% & 45\% & 25\% & 15\% & 5\% & 0\% & 0\% & 0\% \\
\midrule
\multirow{2}{*}{BWS with EL2N} & Test accuracy & 45.02 & 71.87 & 81.79 & 88.87 & 91.59 & 93.39 & 94.44 & 95.06 & 95.32 \\
& Window index & 80\% & 60\% & 45\% & 25\% & 10\% & 5\% & 0\% & 0\% & 0\% \\
\midrule
BWS with forgetting & Test accuracy & 46.10 & 70.70 & 82.29 & 88.74 & 91.80 & 93.59 & 94.54 & 95.23 & 95.37 \\
(Ours) & Window index & 90\% & 70\% & 55\% & 30\% & 15\% & 5\% & 0\% & 0\% & 0\% \\
\midrule
\multirow{2}{*}{Oracle window} & Test accuracy & 47.17 & 72.89 & 82.67 & 89.06 & 91.80 & 93.59 & 94.54 & 95.23 & 95.37 \\
    & Window index & 85\% & 55\% & 50\% & 25\% & 15\% & 5\% & 0\% & 0\% & 0\% \\
\bottomrule
\end{tabular}}
\end{table}

In the implementation of our \ours algorithm, we employ the Forgetting score \citep{forgetting} as a difficulty score. To test the algorithm's adaptability to alternative difficulty scores, we examine its performance when configured with the EL2N score \citep{EL2N} and C-score \citep{c-score}.
Table \ref{tab:diff_score} presents a comparison of the results obtained by our \ours algorithm when utilizing the EL2N score and C-score, against those achieved with the Forgetting score. 
Regardless of the difficulty score used, all the results demonstrate competitive performances, closely approaching those of the oracle window, across a wide range of selection ratios.
We anticipate that the observed phenomenon arises due to a strong correlation between the difficulty scores. The rank correlation between the EL2N score (C-score) and the forgetting score used for comparison is notably high as 0.8836 (0.8500). This suggests that samples sorted by the different difficulty scores would likely follow a similar order of forgetting score. As a result, the best windows selected by BWS for the two different score cases exhibit similarity, as shown by Table \ref{tab:diff_score}.
This consistency shows that the effectiveness of \ours is not limited by the choice of the difficulty score, highlighting its robustness to the sample scores used in sorting.

\newpage

\section{Full Results}\label{sec:full_result}

Table \ref{tab:Starting_points} compares the starting points of the window subsets selected by BWS and those of the oracle window subsets. BWS finds the nearly optimal subsets at broad subset ratios for CIFAR-10/100 and ImageNet dataset.
In Table \ref{tab:CIFAR10_acc}-\ref{tab:noise40_acc}, the oracle window achieves the highest performance among the considered methodologies for almost every selection ratio and dataset, since it finds the best window by directly measuring and comparing the test accuracy of models trained by each window using the test dataset. Since the oracle window cannot be implemented in practice due to significant computational overhead, it is fair to compare the performances among the methods except the oracle window. Thus, we highlight the highest and the second-highest values among the rest of the methodologies except the oracle window in the tables. 

\color{blue}

\begin{table}[!ht]
\centering
\caption{The starting points (\%) of the window subsets obtained by BWS, and those of the oracle window subsets. BWS finds the nearly optimal subsets at broad subset ratios for all datasets.}
\label{tab:Starting_points}
\vspace{0.2em} 
\begin{tabular}{cc|ccccccccc}
\toprule
\multicolumn{2}{c|}{Selection ratio} & 1\% & 5\% & 10\% & 20\% & 30\% & 40\% & 50\% & 75\% & 90\% \\
\midrule

\multirow{2}{*}{CIFAR-10} & BWS(Ours) & 90 & 70 & 55 & 30 & 15 & 5 & 0 & 0 & 0 \\ 
& Oracle window & 85 & 55 & 50 & 25 & 15 & 5 & 0 & 0 & 0 \\ 
\midrule
\multirow{2}{*}{CIFAR-100} & BWS(Ours) & 85 & 95 & 75 & 60 & 25 & 10 & 0 & 0 & 0 \\ 
& Oracle window & 95 & 85 & 80 & 55 & 40 & 20 & 5 & 5 & 0 \\ 
\midrule
\multirow{2}{*}{ImageNet} & BWS(Ours) & 90 & 75 & 70 & 5 & 0 & 0 & 0 & 0 & 0 \\ 
& Oracle window & 85 & 65 & 40 & 15 & 5 & 10 & 10 & 0 & 0 \\ 

\bottomrule
\end{tabular}
\end{table}

\color{black}

\begin{table}[!ht]
\centering
\caption{Test accuracy of CIFAR-10 dataset on ResNet18. We highlight the highest values in bold and the second-highest values in underscore.}
\label{tab:CIFAR10_acc}
\vspace{0.2em} 
\resizebox{\textwidth}{!}{
\begin{tabular}{c|cccccccccc}
\toprule
Selection ratio & 1\% & 5\% & 10\% & 20\% & 30\% & 40\% & 50\% & 75\% & 90\% & 100\% \\
\midrule

Forgetting & 30.08$\pm$1.79 & 42.39$\pm$1.31 & 54.31$\pm$0.23 & 79.19$\pm$0.26 & 89.13$\pm$0.17 & \underline{93.41$\pm$0.05} & \underline{94.49$\pm$0.02} & \textbf{95.31$\pm$0.08} & 95.14$\pm$0.04 & \multirow{9}{*}{95.40$\pm$0.08}\\ 
EL2N & 15.27$\pm$0.36 & 27.01$\pm$0.76 & 41.27$\pm$0.62 & 71.67$\pm$0.82 & 87.17$\pm$0.48 & 93.24$\pm$0.06 & 94.43$\pm$0.13 & 95.13$\pm$0.05 & 95.26$\pm$0.11 & \\ 
LCMat & \underline{41.53$\pm$0.61} & 66.86$\pm$1.00 & 77.48$\pm$1.62 & \underline{87.34$\pm$0.22} & 90.72$\pm$0.06 & 92.45$\pm$0.05 & 93.38$\pm$0.07 & 94.90$\pm$0.06 & 95.19$\pm$0.01 & \\ 
AdaCore & 39.87$\pm$0.75 & 66.40$\pm$1.10 & 77.84$\pm$0.49 & 86.88$\pm$0.05 & 89.90$\pm$0.08 & 91.48$\pm$0.24 & 92.73$\pm$0.17 & 94.47$\pm$0.18 & 95.04$\pm$0.23 & \\ 
CCS & 31.86$\pm$0.72 & 58.89$\pm$1.43 & 72.61$\pm$3.59 & 86.64$\pm$0.35 & \underline{90.94$\pm$0.55} & 93.00$\pm$0.05 & 94.09$\pm$0.17 & 94.95$\pm$0.21 & \underline{95.29$\pm$0.09} & \\ 
Moderate DS & 40.67$\pm$0.50 & \underline{67.53$\pm$0.75} & 76.62$\pm$1.29 & 84.86$\pm$0.09 & 88.46$\pm$0.07 & 90.63$\pm$0.01 & 91.52$\pm$0.08 & 93.69$\pm$0.21 & 94.68$\pm$0.07 & \\ 
Random & 39.10$\pm$0.14 & 67.14$\pm$0.29 & \underline{78.43$\pm$0.72} & 86.87$\pm$0.31 & 89.91$\pm$0.31 & 91.66$\pm$0.06 & 92.83$\pm$0.04 & 94.40$\pm$0.05 & 95.08$\pm$0.19 & \\ 
BWS (Ours) & \textbf{46.10$\pm$2.68} & \textbf{70.70$\pm$0.53} & \textbf{82.29$\pm$0.35} & \textbf{88.74$\pm$0.18} & \textbf{91.80$\pm$0.03} & \textbf{93.59$\pm$0.17} & \textbf{94.54$\pm$0.06} & \underline{95.23$\pm$0.08} & \textbf{95.37$\pm$0.07} & \\ 
Oracle window & 47.17$\pm$0.25 & 72.89$\pm$1.05 & 82.67$\pm$0.43 & 89.06$\pm$0.34 & 91.80$\pm$0.03 & 93.59$\pm$0.17 & 94.54$\pm$0.06 & 95.23$\pm$0.08 & 95.37$\pm$0.07 & \\

\bottomrule
\end{tabular}
}
\end{table}

\begin{table}[!ht]
\centering
\caption{Test accuracy of CIFAR-100 dataset on ResNet50. We highlight the highest values in bold and the second-highest values in underscore.}
\label{tab:CIFAR100_acc}
\vspace{0.2em} 
\resizebox{\textwidth}{!}{
\begin{tabular}{c|cccccccccc}
\toprule
Selection ratio & 1\% & 5\% & 10\% & 20\% & 30\% & 40\% & 50\% & 75\% & 90\% & 100\% \\
\midrule

Forgetting & 7.01$\pm$0.50 & 20.69$\pm$1.13 & 34.22$\pm$1.27 & 50.95$\pm$0.78 & 61.54$\pm$1.06 & 68.92$\pm$0.87 & \textbf{73.84$\pm$0.95} & \textbf{78.55$\pm$0.44} & \textbf{79.69$\pm$0.19} & \multirow{9}{*}{78.81$\pm$0.13}\\ 
EL2N & 3.40$\pm$0.12 & 8.15$\pm$0.17 & 14.06$\pm$0.48 & 28.14$\pm$1.21 & 48.13$\pm$1.77 & 52.25$\pm$5.85 & 71.72$\pm$0.17 & 77.33$\pm$0.70 & 78.96$\pm$0.10 & \\ 
LCMat & \textbf{8.43$\pm$0.44} & \underline{28.51$\pm$0.65} & \underline{42.81$\pm$0.31} & 55.77$\pm$1.45 & 64.39$\pm$1.02 & 67.22$\pm$0.96 & 73.11$\pm$0.81 & 77.51$\pm$0.37 & 78.47$\pm$0.65 & \\ 
AdaCore & 5.56$\pm$0.14 & 22.76$\pm$1.20 & 39.56$\pm$2.53 & 56.81$\pm$1.60 & 65.30$\pm$0.64 & \underline{70.51$\pm$0.64} & 71.18$\pm$1.00 & 76.62$\pm$0.47 & 78.37$\pm$0.32 & \\ 
CCS & 7.49$\pm$0.66 & 24.34$\pm$0.35 & 40.81$\pm$2.11 & \underline{56.81$\pm$1.81} & 63.35$\pm$0.40 & 67.70$\pm$0.64 & 71.04$\pm$0.52 & 74.94$\pm$0.73 & 78.13$\pm$0.31 & \\ 
Moderate DS & 6.05$\pm$0.29 & 24.53$\pm$1.28 & 42.23$\pm$3.03 & 54.72$\pm$1.76 & 64.71$\pm$1.27 & 68.71$\pm$2.45 & 72.61$\pm$0.31 & 75.80$\pm$0.48 & 78.48$\pm$0.13 & \\ 
Random & 5.89$\pm$0.52 & 23.76$\pm$1.12 & 42.03$\pm$1.56 & 55.03$\pm$1.17 & \underline{65.98$\pm$0.50} & 69.23$\pm$1.04 & 72.37$\pm$0.13 & 76.53$\pm$0.52 & 78.29$\pm$0.22 & \\ 
BWS (Ours) & \underline{8.43$\pm$0.49} & \textbf{29.25$\pm$1.15} & \textbf{44.11$\pm$3.13} & \textbf{58.30$\pm$0.65} & \textbf{67.20$\pm$1.67} & \textbf{72.17$\pm$0.42} & \underline{73.83$\pm$0.72} & \underline{77.78$\pm$0.55} & \underline{79.00$\pm$0.29} & \\ 
Oracle window & 10.63$\pm$0.87 & 30.39$\pm$2.87 & 45.16$\pm$1.28 & 58.91$\pm$0.52 & 67.51$\pm$1.22 & 72.70$\pm$0.50 & 75.00$\pm$0.23 & 78.42$\pm$0.14 & 79.00$\pm$0.29 & \\ 

\bottomrule
\end{tabular}}
\end{table}

\begin{table}[!ht]
\centering
\caption{Test accuracy of ImageNet dataset on ResNet50. We highlight the highest values in bold and the second-highest values in underscore.}
\label{tab:ImageNet_acc}
\vspace{0.2em} 
\resizebox{\textwidth}{!}{
\begin{tabular}{c|cccccccccc}
\toprule

Selection ratio & 1\% & 5\% & 10\% & 20\% & 30\% & 40\% & 50\% & 75\% & 90\% & 100\% \\
\midrule

Forgetting & 4.78$\pm$0.10 & 28.18$\pm$0.46 & 45.84$\pm$0.67 & \underline{60.75$\pm$0.60} & \textbf{67.48$\pm$0.11} & \underline{70.26$\pm$0.48} & \textbf{72.73$\pm$0.09} & \underline{74.63$\pm$0.13} & \textbf{75.53$\pm$0.06} & \multirow{11}{*}{75.85$\pm$0.07}\\ 
EL2N & 2.10$\pm$0.08 & 9.80$\pm$0.03 & 20.42$\pm$0.47 & 41.14$\pm$0.04 & 54.42$\pm$0.39 & 63.19$\pm$0.29 & 68.19$\pm$0.13 & 73.91$\pm$0.36 & 74.79$\pm$0.27 & \\ 
Memorization & 0.52$\pm$0.04 & 9.70$\pm$0.21 & 23.80$\pm$0.31 & 44.58$\pm$0.09 & 59.66$\pm$0.06 & 65.92$\pm$0.04 & 70.22$\pm$0.02 & 74.56$\pm$0.24 & 74.94$\pm$0.16 & \\ 
SSL Prototype & 1.33$\pm$0.21 & 20.07$\pm$1.39 & 37.98$\pm$0.08 & 55.25$\pm$1.02 & 61.97$\pm$0.25 & 66.58$\pm$0.28 & 68.85$\pm$0.19 & 73.43$\pm$0.29 & 74.63$\pm$0.28 & \\ 
LCMat & 6.01$\pm$0.31 & 32.26$\pm$0.84 & 46.08$\pm$0.64 & 59.02$\pm$0.36 & 65.28$\pm$0.21 & 68.50$\pm$0.56 & 70.30$\pm$0.46 & 74.13$\pm$0.12 & 74.81$\pm$0.02 & \\ 
AdaCore & 6.01$\pm$0.44 & 31.52$\pm$0.58 & \underline{46.98$\pm$0.80} & 59.26$\pm$1.58 & 65.18$\pm$0.05 & 68.28$\pm$0.05 & 70.72$\pm$0.04 & 73.53$\pm$0.13 & 74.69$\pm$0.00 & \\ 
CCS & 5.04$\pm$0.40 & 31.83$\pm$0.62 & 46.64$\pm$1.08 & 58.77$\pm$0.80 & 64.85$\pm$0.12 & 67.82$\pm$0.24 & 69.89$\pm$0.24 & 73.57$\pm$0.12 & 74.59$\pm$0.03 & \\ 
Moderate DS & 5.97$\pm$0.60 & 32.47$\pm$0.21 & \textbf{47.83$\pm$0.11} & 58.86$\pm$0.14 & 64.71$\pm$0.01 & 67.47$\pm$0.03 & 69.73$\pm$0.08 & 73.16$\pm$0.25 & 74.67$\pm$0.07 & \\ 
Random & \underline{6.14$\pm$0.01} & \underline{33.17$\pm$0.11} & 45.87$\pm$0.07 & 59.19$\pm$0.04 & 65.94$\pm$0.38 & 68.23$\pm$0.00 & 70.14$\pm$0.31 & 73.74$\pm$0.14 & 74.83$\pm$0.08 & \\ 
BWS (Ours) & \textbf{7.61$\pm$0.84} & \textbf{33.96$\pm$1.08} & 46.64$\pm$0.20 & \textbf{62.08$\pm$0.51} & \underline{67.28$\pm$0.20} & \textbf{70.53$\pm$0.16} & \underline{72.63$\pm$0.14} & \textbf{74.67$\pm$0.10} & \underline{75.28$\pm$0.28} & \\ 
Oracle window & 7.89$\pm$0.27 & 33.98$\pm$0.57 & 49.21$\pm$0.76 & 62.62$\pm$0.30 & 68.27$\pm$0.56 & 71.35$\pm$0.24 & 72.91$\pm$0.38 & 74.67$\pm$0.10 & 75.28$\pm$0.28 & \\ 
\bottomrule

\end{tabular}}
\end{table}

\begin{table}[!ht]
\centering
\caption{Test accuracy of CIFAR-10 dataset by training a simple CNN architecture. We highlight the highest values in bold and the second-highest values in underscore.}
\label{tab:cnn_acc}
\vspace{0.2em} 
\resizebox{\textwidth}{!}{
\begin{tabular}{c|cccccccccc}
\toprule
Selection ratio & 1\% & 5\% & 10\% & 20\% & 30\% & 40\% & 50\% & 75\% & 90\% & 100\% \\
\midrule

Forgetting & 34.01$\pm$0.45 & 46.28$\pm$0.67 & 55.04$\pm$0.55 & 67.98$\pm$0.29 & 75.27$\pm$0.08 & 80.19$\pm$0.31 & 83.45$\pm$0.23 & \textbf{86.92$\pm$0.17} & 87.66$\pm$0.29 & \multirow{9}{*}{87.64$\pm$0.14}\\ 
EL2N & 16.60$\pm$0.86 & 30.58$\pm$0.27 & 42.90$\pm$0.18 & 62.90$\pm$0.15 & 73.67$\pm$0.50 & 79.43$\pm$0.36 & 82.83$\pm$0.28 & 86.72$\pm$0.35 & \textbf{87.86$\pm$0.09} & \\ 
LCMat & 46.42$\pm$0.23 & 65.74$\pm$0.65 & 72.54$\pm$0.42 & 78.19$\pm$0.07 & 81.15$\pm$0.11 & 83.36$\pm$0.05 & 84.65$\pm$0.01 & 86.70$\pm$0.29 & \underline{87.73$\pm$0.26} & \\ 
AdaCore & 46.72$\pm$0.21 & \underline{66.69$\pm$0.43} & 73.52$\pm$0.49 & \underline{78.64$\pm$0.27} & 81.62$\pm$0.16 & \underline{83.38$\pm$0.49} & 84.71$\pm$0.05 & 86.59$\pm$0.21 & 87.17$\pm$0.22 & \\ 
CCS & 39.50$\pm$0.96 & 59.86$\pm$0.21 & 68.89$\pm$0.44 & 76.07$\pm$0.57 & 80.55$\pm$0.14 & 83.21$\pm$0.44 & \textbf{84.85$\pm$0.08} & 86.78$\pm$0.37 & 87.46$\pm$0.21 & \\ 
Moderate DS & \underline{48.68$\pm$0.46} & 66.61$\pm$0.29 & 72.64$\pm$0.25 & 76.82$\pm$0.28 & 79.72$\pm$0.17 & 81.35$\pm$0.28 & 82.78$\pm$0.24 & 85.64$\pm$0.27 & 86.91$\pm$0.10 & \\ 
Random & 46.70$\pm$0.91 & 66.27$\pm$0.48 & \underline{73.65$\pm$0.53} & 78.63$\pm$0.37 & \underline{81.70$\pm$0.36} & 83.04$\pm$0.16 & 84.55$\pm$0.10 & 86.43$\pm$0.05 & 87.23$\pm$0.09 & \\ 
BWS (Ours) & \textbf{52.48$\pm$0.42} & \textbf{69.71$\pm$0.37} & \textbf{76.17$\pm$0.30} & \textbf{80.69$\pm$0.06} & \textbf{82.58$\pm$0.20} & \textbf{84.34$\pm$0.49} & \underline{84.76$\pm$0.22} & \underline{86.86$\pm$0.05} & 87.68$\pm$0.06 & \\ 
Oracle window & 53.08$\pm$0.29 & 69.71$\pm$0.37 & 76.17$\pm$0.30 & 80.69$\pm$0.06 & 82.59$\pm$0.20 & 84.34$\pm$0.49 & 85.09$\pm$0.16 & 86.86$\pm$0.05 & 87.68$\pm$0.06 & \\

\bottomrule
\end{tabular}}
\end{table}

\begin{table}[!ht]
\centering
\caption{Test accuracy of CIFAR-10 dataset by training EfficientNet-B0 architecture. We highlight the highest values in bold and the second-highest values in underscore.}
\label{tab:eff_acc}
\vspace{0.2em} 
\resizebox{\textwidth}{!}{
\begin{tabular}{c|cccccccccc}
\toprule
Selection ratio & 1\% & 5\% & 10\% & 20\% & 30\% & 40\% & 50\% & 75\% & 90\% & 100\% \\
\midrule

Forgetting & 26.22$\pm$0.74 & 31.84$\pm$2.06 & 39.56$\pm$3.81 & 62.36$\pm$6.14 & 77.14$\pm$6.16 & 84.53$\pm$1.51 & \underline{89.60$\pm$1.15} & \textbf{92.23$\pm$0.06} & \textbf{92.60$\pm$0.22} & \multirow{9}{*}{92.60$\pm$0.12}\\ 
EL2N & 14.67$\pm$0.28 & 24.97$\pm$0.84 & 32.32$\pm$2.63 & 60.11$\pm$2.11 & 76.03$\pm$0.88 & 84.82$\pm$0.56 & 88.95$\pm$0.74 & 91.04$\pm$0.15 & 92.01$\pm$0.40 & \\ 
LCMat & 29.12$\pm$2.16 & 54.45$\pm$4.23 & 67.35$\pm$4.97 & 76.52$\pm$1.82 & 84.88$\pm$0.53 & 87.20$\pm$1.08 & 89.02$\pm$0.16 & 91.47$\pm$0.20 & \underline{92.46$\pm$0.31} & \\ 
AdaCore & 31.50$\pm$2.29 & 52.96$\pm$5.90 & \underline{69.13$\pm$2.58} & 79.65$\pm$0.45 & \underline{85.06$\pm$0.93} & 86.29$\pm$0.96 & 87.71$\pm$1.37 & 90.64$\pm$0.99 & 91.95$\pm$0.10 & \\ 
CCS & 26.38$\pm$1.58 & 45.78$\pm$4.05 & 59.21$\pm$6.03 & 76.05$\pm$4.40 & 83.85$\pm$1.55 & \underline{87.93$\pm$0.72} & 88.99$\pm$0.52 & 91.80$\pm$0.20 & 92.27$\pm$0.38 & \\ 
Moderate DS & \underline{35.10$\pm$0.78} & \underline{55.24$\pm$2.45} & 68.01$\pm$3.00 & 78.91$\pm$1.42 & 80.93$\pm$2.77 & 85.59$\pm$0.66 & 85.26$\pm$1.32 & 89.96$\pm$0.48 & 91.81$\pm$0.21 & \\ 
Random & 34.34$\pm$3.70 & 48.49$\pm$8.17 & 62.80$\pm$8.18 & \underline{80.71$\pm$0.37} & 84.56$\pm$0.46 & 85.86$\pm$0.47 & 89.10$\pm$0.16 & 91.40$\pm$0.18 & 92.02$\pm$0.24 & \\ 
BWS (Ours) & \textbf{35.86$\pm$3.15} & \textbf{58.06$\pm$4.62} & \textbf{75.88$\pm$1.07} & \textbf{82.46$\pm$0.67} & \textbf{86.16$\pm$0.49} & \textbf{88.23$\pm$0.50} & \textbf{89.84$\pm$0.69} & \underline{92.06$\pm$0.21} & 92.45$\pm$0.29 & \\ 
Oracle window & 42.65$\pm$0.66 & 64.18$\pm$4.56 & 75.88$\pm$1.07 & 83.47$\pm$1.32 & 86.16$\pm$0.49 & 88.23$\pm$0.50 & 89.84$\pm$0.69 & 92.06$\pm$0.21 & 92.45$\pm$0.29 & \\

\bottomrule
\end{tabular}}
\end{table}

\begin{table}[!ht]
\centering
\caption{Test accuracy of CIFAR-10 dataset by fine-tuning ViT pretrained on ImageNet.We highlight the highest values in bold and the second-highest values in underscore.}
\label{tab:vit_acc}
\vspace{0.2em} 
\small{
\begin{tabular}{c|ccccc}
\toprule
Selection ratio & 1\% & 5\% & 10\% & 20\% & 100\%\\
\midrule

Forgetting & 41.77$\pm$8.60 & \underline{96.95$\pm$0.31} & \textbf{98.05$\pm$0.12} & \textbf{98.54$\pm$0.04} & \multirow{9}{*}{98.60$\pm$0.03}\\ 
EL2N & 36.13$\pm$8.15 & 94.90$\pm$0.61 & 97.45$\pm$0.26 & 98.27$\pm$0.01 & \\ 
LCMat & \underline{66.89$\pm$4.74} & 95.96$\pm$0.10 & 97.47$\pm$0.13 & 98.02$\pm$0.10 & \\ 
AdaCore & 64.65$\pm$4.38 & 96.17$\pm$0.30 & 97.24$\pm$0.17 & 97.87$\pm$0.13 & \\ 
CCS & 56.84$\pm$7.70 & 96.77$\pm$0.11 & 97.88$\pm$0.14 & 98.28$\pm$0.10 & \\ 
Moderate DS & 66.84$\pm$3.38 & 95.87$\pm$0.06 & 97.17$\pm$0.14 & 97.73$\pm$0.05 & \\ 
Random & 66.34$\pm$4.61 & 96.37$\pm$0.21 & 97.42$\pm$0.12 & 98.01$\pm$0.11 & \\ 
BWS (Ours) & \textbf{71.42$\pm$3.54} & \textbf{97.05$\pm$0.23} & \underline{98.03$\pm$0.07} & \underline{98.45$\pm$0.04} & \\ 
Oracle window & 73.81$\pm$2.00 & 97.16$\pm$0.25 & 98.06$\pm$0.09 & 98.45$\pm$0.04 & \\

\bottomrule
\end{tabular}}
\end{table}

\begin{table}[!ht]
\centering
\caption{Test accuracy of 20\% label-noise CIFAR-10 dataset. We highlight the highest values in bold and the second-highest values in underscore.}
\label{tab:noise_acc}
\vspace{0.2em} 
\small{
\begin{tabular}{c|ccccc}
\toprule
Selection ratio & 10\% & 20\% & 30\% & 40\% & 100\% \\
\midrule

Forgetting & 49.82$\pm$0.87 & 64.52$\pm$1.60 & 69.89$\pm$1.10 & 72.21$\pm$0.10 & \multirow{9}{*}{82.66$\pm$0.00}\\ 
EL2N & 7.02$\pm$0.22 & 9.48$\pm$0.15 & 21.11$\pm$0.19 & 38.91$\pm$0.87 & \\ 
LCMat & 59.20$\pm$1.21 & 71.54$\pm$1.28 & 77.77$\pm$0.48 & 81.23$\pm$0.42 & \\ 
AdaCore & 10.96$\pm$0.10 & 10.85$\pm$0.26 & 39.14$\pm$0.65 & 58.51$\pm$0.20 & \\ 
CCS & 56.87$\pm$0.52 & 72.20$\pm$0.60 & 77.16$\pm$0.61 & 80.22$\pm$0.73 & \\ 
Moderate DS & \textbf{78.75$\pm$0.32} & \textbf{86.53$\pm$0.24} & \textbf{89.61$\pm$0.32} & \textbf{91.35$\pm$0.21} & \\ 
Random & 64.24$\pm$1.35 & 75.45$\pm$0.67 & 79.58$\pm$0.70 & 81.99$\pm$0.27 & \\ 
BWS (Ours) & \underline{78.74$\pm$0.56} & \underline{84.77$\pm$0.28} & \underline{88.06$\pm$0.03} & \underline{89.79$\pm$0.08} & \\ 
Oracle window & 81.01$\pm$0.21 & 87.32$\pm$0.12 & 90.06$\pm$0.06 & 91.48$\pm$0.09 & \\

\bottomrule
\end{tabular}}
\end{table}

\begin{table}[!ht]
\centering
\caption{Test accuracy of 40\% label-noise CIFAR-10 dataset. We highlight the highest values in bold and the second-highest values in underscore.}
\label{tab:noise40_acc}
\vspace{0.2em} 
\small{
\begin{tabular}{c|ccccc}
\toprule
Selection ratio & 10\% & 20\% & 30\% & 40\% & 100\% \\
\midrule

Forgetting & 53.90$\pm$1.26 & 66.95$\pm$1.09 & 70.78$\pm$0.18 & 69.65$\pm$0.39 & \multirow{9}{*}{72.71$\pm$0.40}\\ 
EL2N & 5.75$\pm$0.19 & 5.81$\pm$0.17 & 5.68$\pm$0.27 & 8.63$\pm$0.14 & \\ 
LCMat & 43.32$\pm$0.44 & 57.73$\pm$1.35 & 65.00$\pm$0.39 & 69.11$\pm$0.31 & \\ 
AdaCore & 9.76$\pm$0.48 & 10.03$\pm$0.09 & 10.42$\pm$0.32 & 10.73$\pm$0.21 & \\ 
CCS & 52.08$\pm$2.15 & 64.65$\pm$1.34 & 70.08$\pm$0.19 & 72.09$\pm$0.18 & \\ 
Moderate DS & \underline{77.69$\pm$0.97} & \textbf{86.21$\pm$0.19} & \textbf{88.63$\pm$0.13} & \underline{83.42$\pm$0.16} & \\ 
Random & 50.33$\pm$0.40 & 60.21$\pm$1.28 & 65.01$\pm$0.60 & 67.68$\pm$0.65 & \\ 
BWS (Ours) & \textbf{78.40$\pm$0.37} & \underline{85.42$\pm$0.08} & \underline{87.76$\pm$0.17} & \textbf{88.97$\pm$0.16} & \\ 
Oracle window & 80.58$\pm$0.41 & 86.06$\pm$0.17 & 88.12$\pm$0.28 & 88.97$\pm$0.16 & \\ 

\bottomrule
\end{tabular}}
\end{table}

\begin{table}[ht]
\caption{Comparison of window subsets of CIFAR-10 in terms of their 1) test accuracy, measured on models  trained with the window subsets (top rows) and 2) accuracy of kernel ridge regression on the training dataset (bottom rows). The best performing windows align well between the two measures. 
}
\label{tab:window_CIFAR10}
\vspace{0.2em} 
  \centering
  \resizebox{\textwidth}{!}{  
\begin{tabular}{c|c|ccccccccccccccccccc}
\toprule
\multicolumn{1}{c|}{Ratio} & Starting point & 0\% & 5\% & 10\% & 15\% & 20\% & 25\% & 30\% & 35\% & 40\% & 45\% & 50\% & 55\% & 60\% & 65\% & 70\% & 75\% & 80\% & 85\% & 90\% \\
\midrule
        
\multirow{2}{*}{10\%} & Test Acc & 56.34 & 58.34 & 68.54 & 70.24 & 74.77 & 78.54 & 81.32 & 81.71 & 82.24 & 82.46 & \textbf{82.67} & 82.29 & 80.95 & 80.53 & 79.75 & 77.99 & 77.41 & 74.88 & 71.09\\ 
 & Regression Acc  & 56.93 & 59.32 & 61.41 & 63.94 & 65.01 & 65.93 & 67.03 & 67.25 & 67.55 & 67.81 & 67.74 & \textbf{67.81} & 67.73 & 67.55 & 67.27 & 66.91 & 66.65 & 65.98 & 65.10\\ 
 \midrule 
 \multirow{2}{*}{20\%} & Test Acc & 79.08 & 83.39 & 86.03 & 87.79 & 88.33 & \textbf{89.06} & 88.74 & 88.42 & 88.06 & 87.23 & 86.86 & 86.12 & 85.37 & 84.29 & 83.10 & 82.09 & 80.42 & - & -\\ 
 & Regression Acc  & 76.27 & 77.07 & 77.85 & 78.98 & 79.41 & 79.69 & \textbf{79.83} & 79.79 & 79.61 & 79.49 & 79.29 & 79.06 & 78.86 & 78.63 & 78.35 & 78.16 & 77.81 & - & -\\ 
 \midrule 
 \multirow{2}{*}{30\%} & Test Acc & 89.45 & 91.14 & 91.77 & \textbf{91.80} & 91.67 & 90.94 & 90.68 & 89.97 & 89.47 & 88.92 & 88.13 & 87.47 & 86.69 & 85.78 & 84.47 & - & - & - & -\\ 
 & Regression Acc  & 83.81 & 84.23 & 84.42 & \textbf{84.49} & 84.47 & 84.34 & 84.20 & 84.00 & 83.85 & 83.70 & 83.50 & 83.33 & 83.17 & 83.05 & 82.86 & - & - & - & -\\ 
 \midrule 
 \multirow{2}{*}{40\%} & Test Acc & 93.08 & \textbf{93.59} & 93.39 & 93.00 & 92.46 & 91.63 & 91.11 & 90.54 & 89.88 & 89.02 & 88.46 & 87.76 & 86.96 & - & - & - & - & - & -\\ 
 & Regression Acc  & 87.42 & \textbf{87.48} & 87.39 & 87.30 & 87.19 & 87.04 & 86.88 & 86.71 & 86.58 & 86.42 & 86.29 & 86.18 & 86.02 & - & - & - & - & - & -\\ 
\bottomrule
\end{tabular}}
\end{table} 

\begin{table}[ht]
\caption{Comparison of window subsets of CIFAR-100 in terms of their 1) test accuracy, measured on models  trained with the window subsets (top rows) and 2) accuracy of kernel ridge regression on the training dataset (bottom rows). The best performing windows align well between the two measures. 
}
\label{tab:window_CIFAR100}
\vspace{0.2em} 
  \centering
  \resizebox{\textwidth}{!}{  
\begin{tabular}{c|c|ccccccccccccccccccc}
\toprule
\multicolumn{1}{c|}{Ratio} & Starting point & 0\% & 5\% & 10\% & 15\% & 20\% & 25\% & 30\% & 35\% & 40\% & 45\% & 50\% & 55\% & 60\% & 65\% & 70\% & 75\% & 80\% & 85\% & 90\% \\
\midrule
        
\multirow{2}{*}{10\%} & Test Acc & 35.64 & 34.81 & 28.31 & 30.53 & 33.22 & 33.07 & 32.12 & 38.03 & 34.69 & 37.98 & 38.36 & 43.06 & 42.24 & 41.73 & 43.60 & 44.11 & \textbf{45.16} & 42.77 & 42.70\\ 
& Regression Acc  & 11.12 & 11.23 & 11.02 & 10.89 & 10.96 & 11.28 & 12.11 & 12.34 & 12.50 & 12.84 & 12.88 & 13.12 & 13.48 & 13.60 & 13.99 & \textbf{14.24} & 14.16 & 14.02 & 14.06\\ 
\midrule 
\multirow{2}{*}{20\%} & Test Acc & 48.78 & 52.02 & 49.43 & 53.03 & 54.39 & 54.99 & 56.60 & 56.51 & 57.08 & 56.83 & 58.70 & \textbf{58.91} & 58.30 & 56.26 & 56.34 & 56.71 & 52.96 & - & -\\ 
& Regression Acc  & 25.62 & 25.58 & 25.78 & 25.76 & 26.48 & 27.00 & 27.36 & 27.59 & 27.67 & 27.85 & 28.17 & 28.29 & \textbf{28.51} & 28.44 & 28.23 & 27.93 & 27.30 & - & -\\ 
\midrule 
\multirow{2}{*}{30\%} & Test Acc & 62.25 & 61.17 & 62.66 & 63.78 & 66.62 & 67.20 & 66.56 & 65.33 & \textbf{67.51} & 66.61 & 63.93 & 64.88 & 63.09 & 60.47 & 59.06 & - & - & - & -\\ 
& Regression Acc  & 43.27 & 43.46 & 43.83 & 43.85 & 44.11 & \textbf{44.37} & 44.22 & 44.23 & 44.01 & 43.78 & 43.56 & 43.21 & 42.56 & 41.95 & 40.92 & - & - & - & -\\ 
\midrule 
\multirow{2}{*}{40\%} & Test Acc & 70.50 & 69.91 & 72.17 & 70.78 & \textbf{72.70} & 72.11 & 69.79 & 71.38 & 69.06 & 68.87 & 67.83 & 66.25 & 63.98 & - & - & - & - & - & -\\ 
& Regression Acc  & 54.05 & 54.36 & \textbf{54.44} & 54.30 & 53.96 & 53.67 & 53.25 & 52.74 & 52.23 & 51.56 & 50.72 & 49.73 & 48.64 & - & - & - & - & - & -\\ 

\bottomrule
\end{tabular}}
\end{table}

\begin{table}[ht]
\caption{Comparison of window subsets of ImageNet in terms of their 1) test accuracy, measured on models  trained with the window subsets (top rows) and 2) accuracy of kernel ridge regression on the training dataset (bottom rows). 
}
\label{tab:window_ImageNet}
\vspace{0.2em} 
  \centering
  \resizebox{\textwidth}{!}{  
\begin{tabular}{c|c|ccccccccccccccccccc}
\toprule
\multicolumn{1}{c|}{Ratio} & Starting point & 0\% & 5\% & 10\% & 15\% & 20\% & 25\% & 30\% & 35\% & 40\% & 45\% & 50\% & 55\% & 60\% & 65\% & 70\% & 75\% & 80\% & 85\% & 90\% \\
\midrule
\multirow{2}{*}{10\%} & Test Acc & 46.78 & 46.34 & 46.49 & 46.89 & 45.50 & 47.07 & 45.82 & 48.07 & \textbf{49.21} & 48.00 & 48.62 & 47.76 & 46.73 & 46.67 & 46.64 & 45.03 & 43.43 & 40.08 & 32.68\\ 
 & Regression Acc  & 35.05 & 35.45 & 35.71 & 35.84 & 35.74 & 35.72 & 35.89 & 35.99 & 36.03 & 36.06 & 36.05 & 36.04 & 36.19 & 36.23 & \textbf{36.24} & 36.17 & 35.90 & 35.01 & 32.43\\ 
 \midrule 
 \multirow{2}{*}{20\%} & Test Acc & 61.02 & 62.08 & 60.60 & \textbf{62.62} & 61.49 & 62.38 & 62.29 & 61.34 & 61.46 & 61.02 & 59.77 & 59.46 & 58.53 & 56.91 & 54.43 & 52.37 & 46.93 & - & -\\ 
 & Regression Acc  & 44.72 & \textbf{44.91} & 44.83 & 44.72 & 44.59 & 44.54 & 44.35 & 44.21 & 44.12 & 43.94 & 43.89 & 43.84 & 43.75 & 43.65 & 43.39 & 42.81 & 41.30 & - & -\\ 
 \midrule 
 \multirow{2}{*}{30\%} & Test Acc & 67.28 & \textbf{68.27} & 67.89 & 67.90 & 68.00 & 67.81 & 67.85 & 66.87 & 66.91 & 65.68 & 64.53 & 62.77 & 61.66 & 58.88 & 55.25 & - & - & - & -\\ 
 & Regression Acc  & \textbf{52.19} & 52.14 & 51.93 & 51.70 & 51.52 & 51.26 & 51.04 & 50.83 & 50.65 & 50.46 & 50.30 & 50.14 & 49.86 & 49.39 & 48.36 & - & - & - & -\\ 
 \midrule 
 \multirow{2}{*}{40\%} & Test Acc & 70.53 & 70.65 & \textbf{71.35} & 71.27 & 70.87 & 70.31 & 69.87 & 68.79 & 68.42 & 67.05 & 66.11 & 63.92 & 61.54 & - & - & - & - & - & -\\ 
 & Regression Acc  & \textbf{53.48} & 53.41 & 53.23 & 52.98 & 52.74 & 52.54 & 52.27 & 52.05 & 51.87 & 51.67 & 51.42 & 50.99 & 50.14 & - & - & - & - & - & -\\

\bottomrule
\end{tabular}}
\end{table} 

\begin{table}[ht]
\caption{Comparison of window subsets of CIFAR-10 dataset with 20\% label noise in terms of their 1) test accuracy, measured on models  trained with the window subsets (top rows) and 2) accuracy of kernel ridge regression on the training dataset (middle rows). We also report the noise portion with each window subset (bottom rows). The best window alignment between the two measures gets less accurate, compared to the case without label noise, since our method (regression) tends to choose more easier samples. However, such tendency also makes the choice of window subset mostly composed of clean-label samples. 
}
\label{tab:noise_window}
\vspace{0.2em} 
  \centering
  \resizebox{\textwidth}{!}{  
\begin{tabular}{c|c|ccccccccccccccccccc}
\toprule
\multicolumn{1}{c|}{Ratio} & Starting point & 0\% & 5\% & 10\% & 15\% & 20\% & 25\% & 30\% & 35\% & 40\% & 45\% & 50\% & 55\% & 60\% & 65\% & 70\% & 75\% & 80\% & 85\% & 90\% \\
\midrule

\multirow{3}{*}{10\%} & Test Acc & 6.98 & 8.93 & 13.80 & 31.90 & 58.00 & 68.77 & 75.21 & 78.29 & 79.18 & 79.04 & 79.85 & 79.99 & 81.01 & 80.86 & 79.72 & 79.53 & 78.74 & 77.81 & 75.71\\ 
& Regression Acc  & 51.54 & 55.95 & 58.10 & 59.52 & 60.42 & 61.68 & 63.86 & 64.94 & 66.11 & 67.58 & 68.74 & 69.64 & 69.73 & 70.09 & 70.20 & 70.43 & 70.54 & 70.13 & 69.73\\ 
& Noise Portion  & 92\% & 85\% & 69\% & 39\% & 15\% & 8\% & 5\% & 4\% & 4\% & 4\% & 3\% & 3\% & 3\% & 3\% & 3\% & 3\% & 3\% & 3\% & 3\%\\ 
\midrule 
 \multirow{3}{*}{20\%} & Test Acc & 9.56 & 17.79 & 35.29 & 63.55 & 78.99 & 84.46 & 86.52 & 87.07 & 87.32 & 87.17 & 86.85 & 86.47 & 86.32 & 85.36 & 84.77 & 84.08 & 82.82 & - & -\\ 
& Regression Acc  & 73.28 & 75.17 & 76.06 & 77.03 & 78.38 & 78.66 & 79.05 & 79.63 & 80.01 & 80.47 & 80.67 & 80.67 & 80.66 & 80.63 & 80.70 & 80.60 & 80.35 & - & -\\ 
& Noise Portion  & 80\% & 62\% & 42\% & 24\% & 10\% & 6\% & 5\% & 4\% & 4\% & 3\% & 3\% & 3\% & 3\% & 3\% & 3\% & 3\% & 3\% & - & -\\ 
\midrule 
 \multirow{3}{*}{30\%} & Test Acc & 20.90 & 38.99 & 62.51 & 79.74 & 87.82 & 89.80 & 90.06 & 89.83 & 89.34 & 88.94 & 88.46 & 88.06 & 87.66 & 86.93 & 86.04 & - & - & - & -\\ 
& Regression Acc  & 80.43 & 81.39 & 82.33 & 83.20 & 83.45 & 83.84 & 84.10 & 84.30 & 84.36 & 84.55 & 84.47 & 84.60 & 84.39 & 84.41 & 84.16 & - & - & - & -\\ 
& Noise Portion  & 59\% & 44\% & 30\% & 17\% & 8\% & 5\% & 4\% & 4\% & 3\% & 3\% & 3\% & 3\% & 3\% & 3\% & 3\% & - & - & - & -\\ 
\midrule 
 \multirow{3}{*}{40\%} & Test Acc & 38.21 & 59.90 & 77.70 & 87.11 & 90.70 & 91.48 & 91.32 & 90.75 & 90.10 & 89.79 & 89.04 & 88.84 & 88.16 & - & - & - & - & - & -\\ 
& Regression Acc  & 86.19 & 86.50 & 86.63 & 86.87 & 86.96 & 87.11 & 87.21 & 87.16 & 87.29 & 87.39 & 87.25 & 87.25 & 87.16 & - & - & - & - & - & -\\ 
& Noise Portion  & 45\% & 34\% & 23\% & 14\% & 7\% & 5\% & 4\% & 4\% & 3\% & 3\% & 3\% & 3\% & 3\% & - & - & - & - & - & -\\ 
 
\bottomrule
\end{tabular}}
\end{table} 

\begin{table}[ht]
\caption{Comparison of window subsets of CIFAR-10 dataset with 40\% label noise in terms of their 1) test accuracy, measured on models  trained with the window subsets (top rows) and 2) accuracy of kernel ridge regression on the training dataset (middle rows). We also report the noise portion in each window subset (bottom rows). The best window alignment between the two measures gets less accurate, compared to the case without label noise, since our method (regression) tends to choose more easier samples. However, such tendency also makes the choice of window subset mostly composed of clean-label samples. 
}
\label{tab:noise40_window}
\vspace{0.2em} 
  \centering
  \resizebox{\textwidth}{!}{  
\begin{tabular}{c|c|ccccccccccccccccccc}
\toprule
\multicolumn{1}{c|}{Ratio} & Starting point & 0\% & 5\% & 10\% & 15\% & 20\% & 25\% & 30\% & 35\% & 40\% & 45\% & 50\% & 55\% & 60\% & 65\% & 70\% & 75\% & 80\% & 85\% & 90\% \\
\midrule

\multirow{3}{*}{10\%} & Test Acc & 6.28 & 6.12 & 6.82 & 7.82 & 7.60 & 9.81 & 21.05 & 45.22 & 62.11 & 73.62 & 77.57 & 78.94 & 80.58 & 79.96 & 79.59 & 78.40 & 77.92 & 77.07 & 75.51\\ 
& Regression Acc  & 1.87 & 1.88 & 1.47 & 2.16 & 3.13 & 12.03 & 25.36 & 46.69 & 56.92 & 61.44 & 63.84 & 65.21 & 65.17 & 65.18 & 65.45 & 65.63 & 65.11 & 64.73 & 64.03\\ 
& Noise Portion  & 92\% & 93\% & 92\% & 90\% & 88\% & 82\% & 61\% & 35\% & 21\% & 15\% & 13\% & 11\% & 10\% & 9\% & 9\% & 8\% & 7\% & 8\% & 9\%\\ 
\midrule 
 \multirow{3}{*}{20\%} & Test Acc & 5.61 & 6.62 & 6.84 & 8.17 & 13.36 & 28.17 & 52.85 & 72.56 & 82.01 & 85.09 & 86.06 & 85.59 & 85.42 & 84.86 & 83.92 & 83.01 & 82.05 & - & -\\ 
& Regression Acc  & 0.89 & 0.98 & 1.35 & 3.32 & 14.45 & 39.13 & 60.53 & 69.82 & 73.12 & 74.90 & 75.85 & 75.86 & 75.93 & 75.56 & 75.27 & 74.93 & 74.35 & - & -\\ 
& Noise Portion  & 92\% & 91\% & 90\% & 86\% & 74\% & 58\% & 41\% & 25\% & 17\% & 13\% & 11\% & 10\% & 9\% & 8\% & 8\% & 8\% & 8\% & - & -\\ 
\midrule 
 \multirow{3}{*}{30\%} & Test Acc & 5.46 & 6.71 & 10.28 & 17.60 & 35.26 & 56.16 & 72.82 & 83.20 & 87.32 & 88.12 & 87.76 & 87.61 & 86.86 & 86.17 & 85.23 & - & - & - & -\\ 
& Regression Acc  & 0.48 & 0.98 & 4.05 & 18.44 & 51.29 & 71.18 & 79.48 & 82.93 & 84.46 & 85.28 & 85.45 & 85.38 & 85.03 & 84.82 & 84.45 & - & - & - & -\\ 
& Noise Portion  & 90\% & 88\% & 80\% & 69\% & 56\% & 44\% & 32\% & 21\% & 14\% & 12\% & 11\% & 9\% & 9\% & 8\% & 8\% & - & - & - & -\\ 
\midrule 
 \multirow{3}{*}{40\%} & Test Acc & 8.50 & 12.79 & 22.46 & 39.42 & 58.61 & 72.52 & 81.96 & 87.17 & 88.90 & 88.97 & 88.63 & 88.11 & 87.28 & - & - & - & - & - & -\\ 
& Regression Acc  & 2.85 & 11.02 & 30.74 & 60.30 & 79.54 & 86.17 & 88.11 & 88.82 & 89.34 & 89.46 & 89.36 & 89.16 & 88.91 & - & - & - & - & - & -\\ 
& Noise Portion  & 83\% & 75\% & 65\% & 55\% & 46\% & 36\% & 26\% & 18\% & 13\% & 11\% & 10\% & 9\% & 9\% & - & - & - & - & - & -\\ 
 
\bottomrule
\end{tabular}}
\end{table} 


\end{document}